\documentclass[runningheads]{llncs}
\usepackage[T1]{fontenc}
\usepackage{graphicx}
\usepackage{hyperref}
\usepackage{color}

\usepackage{etoolbox}
\usepackage{amssymb,amsmath}
\usepackage{mathtools,latexsym}
\usepackage{ntheorem}
\usepackage{bm,amsbsy}
\usepackage[nameinlink,noabbrev,capitalize]{cleveref}
\usepackage{booktabs}
\usepackage{multirow}
\usepackage{enumitem}
\usepackage{relsize}
\usepackage{adjustbox}
\usepackage[compatibility=false]{caption}
\usepackage{subcaption}
\usepackage{comment}
\usepackage{xcolor}
\usepackage{stmaryrd}
\usepackage{bbding}
\usepackage{titletoc}
\usepackage[noadjust]{cite}
\usepackage{algorithm}
\usepackage{algorithmicx}
\usepackage[noend]{algpseudocode}
\usepackage{xspace}
\usepackage{setspace}
\usepackage{empheq}
\usepackage{ifthen}
\usepackage{array}
\usepackage{adjustbox}         % better centering
\usepackage{booktabs,multirow} % nice tables
\usepackage{nicefrac}       % compact symbols for 1/2, etc.
\dottedcontents{section}[0em]{\bfseries\vspace{0.5em}}{2.9em}{1pc}
\dottedcontents{subsection}[0.75em]{}{3.3em}{1pc}
\renewcommand{\contentsname}{\centering Table of Contents}
\usepackage{pgf}
\usepackage{tikz}
\usepackage{forest}

\usetikzlibrary{arrows,decorations.pathmorphing,decorations.pathreplacing,decorations.footprints,fadings,calc,trees,mindmap,shadows,decorations.text,patterns,positioning,shapes}
\usetikzlibrary{arrows,shadows,backgrounds}
\usetikzlibrary{arrows.meta}
\usetikzlibrary{positioning,fit,automata,matrix}
\usetikzlibrary{shapes.symbols,shapes.misc,shapes.arrows}
\usetikzlibrary{matrix,chains,scopes}
\usetikzlibrary{timeline}

\definecolor{gray}{rgb}{.4,.4,.4}
\definecolor{midgrey}{rgb}{0.5,0.5,0.5}
\definecolor{middarkgrey}{rgb}{0.35,0.35,0.35}
\definecolor{darkgrey}{rgb}{0.3,0.3,0.3}
\definecolor{darkred}{rgb}{0.7,0.1,0.1}
\definecolor{midblue}{rgb}{0.2,0.2,0.7}
\definecolor{darkblue}{rgb}{0.1,0.1,0.5}
\definecolor{defseagreen}{cmyk}{0.69,0,0.50,0}

\definecolor{tyellow1}{HTML}{FCE94F}
\definecolor{tyellow2}{HTML}{EDD400}
\definecolor{tyellow3}{HTML}{C4A000}
\definecolor{torange1}{HTML}{FCAF3E}
\definecolor{torange2}{HTML}{F57900}
\definecolor{torange3}{HTML}{C35C00}
\definecolor{tbrown1}{HTML}{E9B96E}
\definecolor{tbrown2}{HTML}{C17D11}
\definecolor{tbrown3}{HTML}{8F5902}
\definecolor{tgreen1}{HTML}{8AE234}
\definecolor{tgreen2}{HTML}{73D216}
\definecolor{tgreen3}{HTML}{4E9A06}
\definecolor{tblue1}{HTML}{729FCF}
\definecolor{tblue2}{HTML}{3465A4}
\definecolor{tblue3}{HTML}{204A87}
\definecolor{tpurple1}{HTML}{AD7FA8}
\definecolor{tpurple2}{HTML}{75507B}
\definecolor{tpurple3}{HTML}{5C3566}
\definecolor{tred1}{HTML}{EF2929}
\definecolor{tred2}{HTML}{CC0000}
\definecolor{tred3}{HTML}{A40000}
\definecolor{tlgray1}{HTML}{EEEEEC}
\definecolor{tlgray2}{HTML}{D3D7CF}
\definecolor{tlgray3}{HTML}{BABDB6}
\definecolor{tdgray1}{HTML}{888A85}
\definecolor{tdgray2}{HTML}{555753}
\definecolor{tdgray3}{HTML}{2E3436}

\newcommand{\hlight}[1]{{\color{darkred}#1}}
\newcommand{\rhlight}[1]{\hlight{#1}}
\newcommand{\dghlight}[1]{{\color[RGB]{0,120,0}#1}}

\hypersetup{
  colorlinks = true,
  linkcolor = midblue,
  urlcolor  = midblue,
  citecolor = midblue,
  anchorcolor = midblue
}

\makeatletter
\renewcommand\paragraph{\@startsection{paragraph}{4}{\z@}%
  {3.25ex \@plus1ex \@minus.2ex}%
  {-1em}%
  {\normalfont\normalsize\bfseries}}
\makeatother

\newenvironment{Proof}{\noindent{\em Proof.~}}{\hfill$\Box$\\[-0.25cm]}

\newcommand{\xendex}{\hfill\ensuremath\triangleleft}
\AtEndEnvironment{example}{$\xendex$}

\newcommand{\fml}[1]{{\mathcal{#1}}}

\newcommand{\tn}[1]{\textnormal{#1}}
\newcommand{\tbf}[1]{\textnormal{\bfseries #1}}
\newcommand{\tsf}[1]{\textsf{#1}}
\newcommand{\msf}[1]{\ensuremath\mathsf{#1}}
\newcommand{\mbf}[1]{\ensuremath\mathbf{#1}}
\newcommand{\mbb}[1]{\ensuremath\mathbb{#1}}
\newcommand{\mrm}[1]{\ensuremath\mathrm{#1}}
\newcommand{\mfrak}[1]{\ensuremath\mathfrak{#1}}
\newcommand{\ite}{\tn{ite}}
\newcommand{\oper}[1]{\ensuremath\textnormal{\smaller{\textsf{#1}}}}

\newcommand{\hamd}{\ensuremath\mathfrak{D}_h}
\newcommand{\encls}[1]{\ensuremath\mathfrak{E}_{#1}}

\newcommand{\outc}{\oper{outc}}
\newcommand{\TRUE}{\textbf{true}\xspace}
\newcommand{\FALSE}{\textbf{false}\xspace}
\newcommand{\SAT}{\ensuremath\mathsf{SAT}}
\newcommand{\minhs}{\ensuremath\mathsf{MinimumHS}}
\newcommand{\decide}{\ensuremath\mathsf{Decide}}

\newcommand{\prtaxp}{\ensuremath\mathsf{reportAXp}}
\newcommand{\prtcxp}{\ensuremath\mathsf{reportCXp}}
\newcommand{\falselits}{\ensuremath\mathsf{PickFalseLits}}
\newcommand{\mkfix}{\ensuremath\mathsf{FixAttr}}
\newcommand{\mkuniv}{\ensuremath\mathsf{FreeAttr}}

\DeclareMathOperator*{\textco}{\textnormal{\bfseries{\textsf{CO}}}}
\DeclareMathOperator*{\chkco}{\textnormal{\bfseries{\textsf{CO}}}\!}
\newcommand{\consistent}[1]{\chkco\left(#1\right)}
\newcommand{\lencode}[1]{\left\llbracket{#1}\right\rrbracket_{\fml{T}}}
\newcommand{\llencode}[2]{\left\llbracket{#1}\right\rrbracket_{#2}}

\newcommand{\oneaxp}{\ensuremath\mathsf{oneAXp}}
\newcommand{\onecxp}{\ensuremath\mathsf{oneCXp}}
\newcommand{\onexp}{\ensuremath\mathsf{oneXP}}
\newcommand{\minxp}{\ensuremath\mathsf{minXP}}

\newcommand{\predicate}{\mbb{P}}
\newcommand{\predaxp}{\predicate_{\tn{axp}}}
\newcommand{\predcxp}{\predicate_{\tn{cxp}}}
\newcommand{\cl}{\ensuremath\varsigma}

\newcommand{\waxp}{\ensuremath\mathsf{WeakAXp}}
\newcommand{\wcxp}{\ensuremath\mathsf{WeakCXp}}
\newcommand{\axp}{\ensuremath\mathsf{AXp}}
\newcommand{\cxp}{\ensuremath\mathsf{CXp}}

\newcommand{\wlaxp}{\ensuremath\mathsf{WeakLAXp}}
\newcommand{\wlcxp}{\ensuremath\mathsf{WeakLCXp}}

\newcommand{\Prob}{\ensuremath\tn{Pr}}
\newcommand{\prob}{\Prob}

\newcommand{\wpaxp}{\ensuremath\mathsf{WeakPAXp}}

\DeclareMathOperator*{\nentails}{\,\nvDash}
\DeclareMathOperator*{\entails}{\,\vDash}
\DeclareMathOperator*{\lequiv}{\leftrightarrow}
\DeclareMathOperator*{\limply}{\rightarrow}

\DeclareMathOperator*{\ddp}{\tn{D}^{\tn{p}}}
\DeclareMathOperator*{\stwop}{\Sigma_2^{\tn{p}}}
\newcommand{\paptupl}{\ensuremath(X,\fml{H},\fml{M},\fml{T},\varsigma)}
\newcommand{\papdef}{$P=\paptupl$\xspace}

\newcommand{\jnote}[1]{\medskip\noindent$\llbracket$\textcolor{darkred}{joao}: \emph{\textcolor{middarkgrey}{#1}}$\rrbracket$\medskip}

\newcommand{\jnoteF}[1]{}

\newcommand{\keepmark}{{\small\Checkmark}}
\newcommand{\dropmark}{{\small\XSolidBrush}}

\newcounter{tableeqn}[table]

\newcolumntype{L}[1]{>{\raggedright\let\newline\\\arraybackslash\hspace{0pt}}m{#1}}
\newcolumntype{C}[1]{>{\centering\let\newline\\\arraybackslash\hspace{0pt}}m{#1}}
\newcolumntype{R}[1]{>{\raggedleft\let\newline\\\arraybackslash\hspace{0pt}}m{#1}}

\makeatletter
\def\xscriptsize{\@setfontsize\scriptsize{8pt}{9pt}}
\def\xtiny{\@setfontsize\tiny{7pt}{8pt}}
\def\xlarge{\@setfontsize\large{11pt}{12pt}}
\def\xLarge{\@setfontsize\Large{12pt}{14pt}}
\def\xLARGE{\@setfontsize\LARGE{13pt}{15pt}}
\def\xhuge{\@setfontsize\huge{20pt}{20pt}}
\def\xHuge{\@setfontsize\Huge{28pt}{28pt}}
\makeatother

\setlist[enumerate]{topsep=1pt,itemsep=1.0pt}
\setlist[itemize]{topsep=1pt,itemsep=1.0pt}

\tikzset{
  0 my edge/.style={densely dashed, my edge},
  my edge/.style={-{Stealth[]}},
}

\newcommand{\vboxedeq}[3]{\begin{empheq}[box={\fboxsep=#1\fbox}]{equation*}\label{#2}#3\end{empheq}}

\crefname{enumi}{}{}

\newboolean{fullversion}              % Include Q&A slides
\setboolean{fullversion}{true}       %

\newboolean{shownotes}                % Include personal notes
\setboolean{shownotes}{true}         %

\begin{document}
\title{%
  %Logic-Based Formal Explainability in AI%
  Logic-Based Explainability in Machine Learning%
  %Logic-Based Explanations \\of Machine Learning Models%
  %\thanks{Supported by organization x.}
}
%
%\titlerunning{Abbreviated paper title}
% If the paper title is too long for the running head, you can set
% an abbreviated paper title here
%
\author{Joao Marques-Silva%\inst{1}
  \orcidID{0000-0002-6632-3086}
  %\and
  %Second Author\inst{2,3}\orcidID{1111-2222-3333-4444} \and
  %Third Author\inst{3}\orcidID{2222--3333-4444-5555}
}
\authorrunning{J. Marques-Silva}
% First names are abbreviated in the running head.
% If there are more than two authors, 'et al.' is used.
%
\institute{IRIT, CNRS, Toulouse, France\\
\email{\href{mailTo:joao.marques-silva@irit.fr}{joao.marques-silva@irit.fr}}%, %\\
%\url{http://jpmarquessilva.github.io}
%\and
%ABC Institute, Rupert-Karls-University Heidelberg, Heidelberg, Germany\\
%\email{\{abc,lncs\}@uni-heidelberg.de}
}

%
%\maketitle              % typeset the header of the contribution
%
{\def\addcontentsline#1#2#3{}\maketitle}
\begin{abstract}
  The last decade witnessed an ever-increasing stream of successes in
  Machine Learning (ML).
  These successes offer clear evidence that ML is bound to become
  pervasive in a wide range of practical uses, including many that
  directly affect humans.
  Unfortunately, the operation of the most successful ML models is 
  incomprehensible for human decision makers.
  As a result, the use of ML models, especially in high-risk and
  safety-critical settings is not without concern.
  In recent years, there have been efforts on devising approaches for
  explaining ML models.
  Most of these efforts have focused on so-called model-agnostic
  approaches. However, all model-agnostic and related approaches offer no
  %%the vast majority of the work on explaining ML models offers
  guarantees of rigor, hence being referred to as non-formal. For
  example, such non-formal explanations can be consistent with
  different predictions, which renders them useless in practice.
  This paper overviews the ongoing research efforts on computing
  rigorous model-based explanations of ML models; these being referred
  to as formal explanations.
  These efforts encompass a variety of topics, that include the actual
  definitions of explanations, the characterization of the complexity
  of computing explanations, the currently best logical encodings for
  reasoning about different ML models, and also how to make
  explanations interpretable for human decision makers, among others. 

  % \and Model-accurate explanations.
\end{abstract}

\tableofcontents\clearpage

\section{Introduction} \label{sec:intro}

% 1. Progress in ML and concerns about trust

%%(...Progress in ML/AI \& concerns about trust...)\\

% 2. Explainability in AI, and recent work on explainability in ML?

%%(... Explainability in AI and more recently in ML...)\\

Recent years witnessed remarkable advancements in artificial
intelligence (AI), concretely in machine learning
(ML)~\cite{bengio-nature15,hinton-cacm17,bengio-cacm20,bengio-cacm21}.
These advancements have triggered an ever-increasing range of
practical applications~\cite{bengio-bk16}.
Some of these applications often impact humans, with credit worthiness
representing one such application, among many others~\cite{eu-aiact21}.
Unfortunately, the most promising ML models are inscrutable in their
operation, with the term \emph{black-box} being often ascribed to such
ML models. 
Black-box ML models cause distrust, especially when their operation is
difficult to understand or it is even incorrect, not to mention
situations where the operation of ML models is the likely cause for
events with disastrous
consequences~\cite{dearman-alr19,cnn-tesla21,it-tesla21}, but also the
case of unfairness and bias~\cite{propublica16,theverge20}. (The
issues caused by AI systems are illustrated by an ever-increasing list
of incidents~\cite{aiidb23,mcgregor-aaai21}.)
Moreover, recent work argues~\cite{bianchi-jair23} that Perrow's
framework of normal accidents~\cite{perrow-bk84} (which has been used
to explain the occurrence of catastrophic accidents in the past) also
applies to AI systems, thereby conjecturing that such (catastrophic)
accident(s) in AI systems should be expected in the near future.
Motivated by this state of affairs, but also by recent
regulations and
recommendations~\cite{eu-gdpr16,eu-hlegai19,oecd-rcai21}, and by
existing proposals of regulation of AI/ML
systems~\cite{eu-aiact21,unesco-reai21,algwatch20}, there is a
pressing need for building trust into the operation of ML models.
The demand for clarifying the operation of black-box decision making
has motivated the rapid growth of research in the general theme of
\emph{explainable AI} (XAI). XAI can be viewed as the process of
aiding human decision makers to understand the decisions made by AI/ML
systems, with the purpose of delivering trustworthy AI.
The importance of both trustworthy AI and XAI is illustrated by recent 
guidelines, recommendations and regulations put forward by the
European Union (EU), the United States government, the Australian 
government, the OECD and
UNESCO~\cite{eu-gdpr16,darpa-xai16,eu-hlegai19,eu-hlegai20,algwatch20,eu-coplan21,eu-aiact21,us-nraird19,au-ai-ethics21,au-ai-plan21,oecd-rcai21,unesco-reai21},
among others.
Motivated by the above, there have been calls for the use of formal
methods in the verification of systems of AI and
ML~\cite{seshia-cacm22}, with explainability representing a key
component~\cite{msi-aaai22}. There have also been efforts towards
developing an understanding of past incidents in AI
systems~\cite{mcgregor-corr20,mcgregor-aaai21,yampolskiy-corr21,zhou-corr22,mcgregor-corr22a,mcgregor-corr22b}.

Well-known explainability approaches include so-called model-agnostic
methods~\cite{guestrin-kdd16,lundberg-nips17,guestrin-aaai18} and, for
neural networks, approaches based on variants of saliency
maps~\cite{vedaldi-iclr14,muller-dsp18,muller-ieee-proc21}.
Unfortunately, the most popular XAI approaches proposed in recent
years are marred by lack of rigor, and provide explanations that are
often logically
unsound~\cite{inms-corr19,nsmims-sat19,ignatiev-ijcai20}. (One
illustrative example is that of an explanation consistent both with a 
declined bank loan application and with an approved bank loan
application~\cite{msi-aaai22}.)
The drawbacks of these (non-formal) XAI approaches raise important
concerns in settings that impact humans. Example settings include
those referred to as \emph{high-risk} and
\emph{safety-critical}\footnote{%
The definition of \emph{high-risk} in this proposal is aligned with
recent EU documentation~\cite{eu-aiact21}. Concrete examples include
the management and operation of critical infrastructure, credit
worthiness, law enforcement, among many others~\cite{eu-aiact21}.
Some authors refer to \emph{high-stakes} when addressing related
topics~\cite{rudin-naturemi19}.
By safety-critical, we take the meaning that is common in formal
methods~\cite{knight-icse02}, namely settings where errors are
unacceptable, e.g.\ where human lives are at risk. There is growing
interest in deploying ML-enabled systems in safety-critical
applications (e.g.~\cite{trivedi-ieee-comp17}).
}.
The use of unsound explainability methods in either high-risk or
safety-critical could evidently have catastrophic consequences. (And
there are unfortunately too many examples of bugs having massive 
economic cost, or that resulted in the loss of
lives~\cite{mcquaid-jsep12,wired05,zdnet07,newscientist13,arstech18,medium19}.)

A more recent alternative XAI approach offers formal guarantees of
rigor, it is logic-based, and it is in most cases based on efficient
automated reasoning tools.
This document offers an overview of the advancements made in the
general field of \emph{formal explainability} in AI (FXAI).

% 3. What this paper is about:

\paragraph{A brief history of FXAI.}
The recent work on formal explainability in machine learning
finds its roots in the independent efforts of two research
teams~\cite{darwiche-ijcai18,inms-aaai19}\footnote{%
  It should be noted that efforts towards explaining the operation of
  systems of artificial intelligence (AI) can be traced back to at
  least the late 70s~\cite{swartout-ijcai77}, with follow up work
  since
  then~\cite{swartout-aij83,shanahan-ijcai89,selman-aaai90,gottlob-ese90,gottlob-jacm95,simari-aij02,uzcategui-aij03,amgoud-aaai06,amgoud-aamasj08,amgoud-aij09,toni-ecai14,jarvisalo-kr16,imms-ecai16}. Nevertheless,
  the interest in formalizing explanations is documented in much
  earlier work~\cite{oppenheim-ps48}. A distinctive aspect of recent
  work is not only the focus on ML models, namely classifiers, but
  also the research on novel topics, e.g.~contrastive and
  probabilistic explanations, which we will define later.}.
The initial goals of this earlier work seemed clear at the outset: to
propose a formal alternative to the mostly informal approaches to
explainability that were being investigated at the time. Nevertheless,
the experimental results in these initial works also raised concerns
about the practical applicability of formal explanations.
However, it soon became clear that there was much more promise to
formal explainability than what the original works anticipated.
Indeed, we claim that it is now apparent that formal explainability
represents an emerging field of research, and one of crucial
importance. A stream of results in recent years amply support this
claim~\cite{darwiche-ijcai18,inms-aaai19,darwiche-aaai19,inms-nips19,inms-corr19,nsmims-sat19,hazan-aies19,darwiche-pods20,ignatiev-ijcai20,darwiche-ecai20,icshms-cp20,inams-aiia20,marquis-kr20,darwiche-kr20,toni-kr20,mazure-sum20,iims-corr20,msgcin-nips20,barcelo-nips20,iims-corr20,msgcin-icml21,ims-ijcai21,kwiatkowska-ijcai21,ims-sat21,asher-cdmake21,cms-cp21,mazure-cikm21,hiims-kr21,marquis-kr21,barcelo-nips21,tan-nips21,lorini-clar21,amgoud-ecsqaru21,hiicams-corr21,toni-aij21,kutyniok-jair21,darwiche-jair21,iincms-corr21,msi-aaai22,hiicams-aaai22,iisms-aaai22,rubin-aaai22,darwiche-aaai22,marquis-aaai22,tan-stoc22,tan-icml22a,amgoud-ijcai22,marquis-ijcai22a,leite-kr22,lorini-wollic22,labreuche-sum22,barcelo-nips22,iims-jair22,marquis-dke22,darwiche-jlli22-web,waldchen-phd22,iincms-corr22,yisnms-corr22,hms-corr22,ims-corr22,barcelo-corr22,ihincms-corr22,cms-aij23,hims-aaai23,yisnms-aaai23,hcmpms-tacas23,katz-tacas23,amgoud-ijar23}. %
%~\cite{darwiche-ijcai18,inms-aaai19,darwiche-aaai19,inms-nips19,inms-corr19,nsmims-sat19,hazan-aies19,darwiche-pods20,ignatiev-ijcai20,darwiche-ecai20,icshms-cp20,inams-aiia20,marquis-kr20,iims-corr20,msgcin-nips20,barcelo-nips20,msgcin-icml21,ims-ijcai21,kwiatkowska-ijcai21,ims-sat21,cms-cp21,hiims-kr21,marquis-kr21,barcelo-nips21,tan-nips21,lorini-clar21,amgoud-ecsqaru21,kutyniok-jair21,darwiche-jair21,iincms-corr21,msi-aaai22,hiicams-aaai22,iisms-aaai22,rubin-aaai22,darwiche-aaai22,marquis-aaai22,tan-stoc22,tan-icml22a,amgoud-ijcai22,marquis-ijcai22a,leite-kr22,lorini-wollic22,labreuche-sum22,barcelo-nips22,iims-jair22,marquis-dke22,darwiche-jlli22-web,waldchen-phd22,iincms-corr22,yisnms-corr22,hms-corr22,ims-corr22,barcelo-corr22,ihincms-corr22,cms-aij23,hims-aaai23,yisnms-aaai23,hcmpms-tacas23,katz-tacas23,amgoud-ijar23}.
%
Among these, several results are significant.
It has been shown that computing one explanation is tractable for a
number of
classifiers~\cite{msgcin-nips20,iims-corr20,msgcin-icml21,hiims-kr21,cms-cp21,hiicams-aaai22,darwiche-aaai22,iims-jair22}.
Different duality results have been
obtained~\cite{inms-nips19,inams-aiia20}, which relate different
kinds of explanations.
Practically efficient logic encodings have been devised for computing
explanations for a number of families of
classifiers~\cite{inms-corr19,ignatiev-ijcai20,ims-ijcai21,ims-sat21,iisms-aaai22,marquis-aaai22}.
Compilation approaches for explainability have been studied in a
number of recent works~\cite{darwiche-ijcai18,darwiche-aaai19,darwiche-pods20,darwiche-ecai20,darwiche-jair21,darwiche-aaai22}.
A number of computational complexity results, that cover the computation
of one explanation but also other queries, have been
proved~\cite{ims-ijcai21,ims-sat21,msgcin-icml21,hiims-kr21,marquis-kr21}.
Different explainability queries have been
studied~\cite{marquis-kr20,hiims-kr21,marquis-kr21}.
The size of formal explanations have been addressed by considering
probabilistic
explanations~\cite{kutyniok-jair21,iincms-corr21,waldchen-phd22,iincms-corr22,barcelo-corr22}.
The effect of input constraints on explainability, that restrict the
points in feature space to consider, has been studied in recent works~\cite{rubin-aaai22,yisnms-corr22}.
The use of surrogate models for computing formal explanations of
complex models has been proposed~\cite{mazure-cikm21}.
Formal explanations have been applied in different application
domains~\cite{kwiatkowska-ijcai21}.
Furthermore, initial links between explainability and fairness,
robustness and model learning have been
uncovered~\cite{inms-nips19,icshms-cp20}.
Given the above, the purpose of this document is to offer an account
of what we feel have been the most important results in this novel
field of research, up until the time of writing. %\\
%

%\ifthenelse{\boolean{}}{
%  \jnote{}
%}{}

\jnoteF{Add recent papers by Tan.}

\jnoteF{The paper covers logic-based formal explainability, and highlights
  the applications of automated reasoners, e.g.\ reasoners for SAT,
  MILP, SMT, etc.}

\paragraph{Main goals.}
The paper aims to offer a high-level comprehensive overview of the
emerging field of formal explainability. The paper starts by covering
existing formal definitions of explanations. One class of explanations
answers a \textbf{Why?} question; these are referred to as abductive 
explanations. Another class of explanations answers a \textbf{Why not?}
question; these are referred to as contrastive explanations. Then, the
paper builds on the rigorous definitions of explanations to study how
formal explanations can be computed in practice, highlighting some of
the algorithms used.
The paper uses running examples to illustrate how such explanations
are computed in practice.
Moreover, the paper also overviews  families of classifiers for
which computing one explanation is tractable. These include decision
trees and naive bayes classifiers, among several others. Furthermore,
the paper summarizes recent progress along a number of lines of
research, which aim at making formal explainability a practical option
of choice.
Concrete examples of lines of research include: i) answering a growing 
number of explainability queries, e.g.\  enumeration of explanations;
ii) computing probabilistic explanations, which trade-off explanation
size for rigor; iii) taking into account input constraints and
distributions, since not all inputs may be possible for an ML model;
and also iv) approximating complex ML models with simpler models,
which are easier to explain. The paper also overviews a number of
additional topics of research in the general area of formal
explainability.
Since the paper aims at offering a broad overview of the field, some
more technical aspects are omitted, and left to the existing
references.

\paragraph{Additional goals.}
This document also takes the opportunity to deconstruct a number of
misconceptions that pervade Machine Learning research.
One unfortunately common misconception is that logic is inadequate for
reasoning about ML models. This paper, but also a growing list of
references (see above), offer ample evidence that this is certainly
not the case. For example, formal explanations for random
forests~\cite{ims-ijcai21} were shown to be more efficient to compute
than those obtained with heuristic methods.
Another common misconception is that computationally hard
(e.g.\ NP-hard, $\ddp$-hard, etc.) problems are
\emph{intractable},
%\footnote{Some bibliography on computational complexity does indeed
%  associate problems that are computationally hard (from a
%  computational complexity perspective) with practical intractability.
%  However, evidence gathered over more than two decades attests to the
%  contrary.}
and so large-scale problems cannot be solved in practice. By now,
there are more than two decades of comprehensive experimental evidence
that attests to the contrary~\cite{sat-handbook21,vardi-cup20}. In
many practical settings, automated reasoners are often (and sometimes
most often) remarkably efficient at solving computationally hard
problems of very large
scale~\cite{vardi-cacm10a,vardi-cacm14a,vardi-cup20}.
Boolean satisfiability (SAT) solvers (but also mixed integer linear
programming (MILP) solvers) are prime examples of the progress
that has been observed in improving, sometimes dramatically, the
practical efficiency of automated
reasoners~\cite{sat-handbook21,mc-handbook18,bixby-or02,bixby-aor07,berthold-eurojco22}\footnote{%
It should be clarified that SAT, MILP and SMT automated reasoners and
their variants solve their target problems \emph{exactly}, provided
such reasoners are given enough
time~\cite{sat-handbook21,mc-handbook18,bixby-or02,bixby-aor07,berthold-eurojco22}.
Nevertheless, these automated reasoners
should not be confused with the burgeoning field of exact exponential
and parameterized algorithms~\cite{fomin-bk10,fomin-bk15}.
The former, i.e.\ automated reasoners, are without exception
extensively validated and evaluated in practical settings, being
applied to large-scale
problem-solving~\cite{sat-handbook21,mc-handbook18}; regarding the
latter, decades of research have resulted in significant theoretical 
advances, but these have not been matched by practical impact.}. The
bottom line is that some of these computational problems may be
intractable in theory, but in practice that is hardly ever the
case~\cite{vardi-cacm14a,vardi-cup20}.
Another quite common misconception is the existence of so-called
interpretable models, e.g.\ decision trees, lists and sets, but even
linear classification models. For example, some of the best known
model agnostic (and so non-formal) explainability
approaches~\cite{guestrin-kdd16,lundberg-nips17} learn a simple
\emph{interpretable} model as the explanation for a more complex
model.
Some other authors propose the use of interpretable models as the
explanation itself~\cite{rudin-naturemi19,molnar-bk20,rudin-ss22},
specially in high-risk and safety-critical settings.
There is by now comprehensive
evidence~\cite{msgcin-nips20,iims-corr20,hiims-kr21,ims-sat21,iims-jair22} 
that even these so-called interpretable models can provide
explanations that are \emph{arbitrarily} redundant. Therefore, even
so-called interpretable models ought to be explained with the methods
described in this paper%\\
\footnote{More importantly, the growing evidence on the need to
explain ``interpretable''
models~\cite{msgcin-nips20,iims-jair22,hiims-kr21,ims-sat21,iims-jair22}
justifies wondering, in hindsight, about the practical relevancy of
learning (quasi-)optimal interpretable 
models~\cite{nijssen-kdd07,hebrard-cp09,nijssen-dmkd10,rudin-aistats15,leskovec-kdd16,bertsimas-ml17,rudin-jmlr17a,rudin-jmlr17b,rudin-kdd17,verwer-cpaior17,rudin-mpc18,rudin-aistats18,nipms-ijcai18,ipnms-ijcar18,meel-cp18,meel-aies19,verwer-aaai19,rudin-nips19,avellaneda-corr19,avellaneda-aaai20,schaus-cj20,schaus-aaai20,rudin-icml20,janota-sat20,hebrard-ijcai20,schaus-ijcai20a,schaus-ijcai20b,meel-ecai20,ignatiev-cp20,ignatiev-jair21,ilsms-aaai21,demirovic-aaai21,szeider-aaai21a,szeider-aaai21b,verwer-icml21,mcilraith-cp21,ansotegui-corr21,demirovic-jmlr22,meel-jair22,verwer-aaai22,rudin-aaai22,rudin-facct22,rudin-cikmw22}.}.

\paragraph{Organization.}
The paper is organized in two main parts.
The first part introduces a number of well-established topics.
\cref{sec:prelim} introduces the definitions and notation used
throughout the paper.
\cref{sec:dxps} introduces the formal definitions of explanations that
have been studied in recent years.
Based on the proposed definitions of explanations,~\cref{sec:cxps}
studies algorithms for the computation of explanations.
In addition, as shown in~\cref{sec:txps}, for some families of
classifiers, there are polynomial-time algorithms for computing one
explanation.
The rest of the document covers the second part of the paper, and
targets topics related with ongoing research. Thus, this second part
is presented with less detail.
\cref{sec:qxps} goes beyond computing one explanation, and delves into
explainability queries, that include enumeration of explanations and
deciding feature inclusion in explanations.
\cref{sec:pxps} addresses the size of explanations, and proposes
\emph{probabilistic explanations} as a mechanism to reduce explanation
size.
\cref{sec:ixps} overviews approaches for accounting for input
constraints and distributions.
\cref{sec:axps} overviews work on approximating complex ML models with
surrogate (or distilled) simpler models, which are easier to explain.
\cref{sec:exts} summarizes a number of additional topics of research.
Finally,~\cref{sec:conc} identifies some directions of research, and
concludes the paper.

%Over the last few years, a large number of works surveyed the ongoing
%efforts on explaining ML models~\cite{}. Without exception, these
%works do not cover the uses of logic in explainability.
%%
%This paper complements earlier works, and offers
%%In contrast with past works, this paper 
%%This paper changes this state of affairs, and offers
%an overview of the progress that has been observed in
%exploiting logic-based (and so rigorous) approaches for
%explainability.
%%
%For completeness, the paper also glosses over non-logic (and so most
%often non-rigorous) approaches for explainability.

% 4. Brief overview of FXAI

%(...Overview of XAI...)\\

% 5. Why FXAI?

%(...Why FXAI?...)\\

\begin{comment}
%
%
\paragraph{Emergence of formal explainability.}
%
\cref{fig:topics} depicts the breadth of research challenges in the
emerging field of formal explainability.
%

\begin{figure}[t]
  \begin{center}
    \includegraphics[scale=0.425]{./figs/FormalXAI.png}
  \end{center}
  \caption{Topics of research in formal explainability} \label{fig:topics}
\end{figure}

\jnote{%
  Papers to cite:
  \begin{enumerate}
  \item Our team
  \item CRIL
  \item UCLA
  \item Rubin
  \item Kwiatkowska
  \item Leite(?)
  \item Amgoud
  \item Arenas \& Barcelo
  \item Tan
  \item Kutyniok
  \end{enumerate}
}
%
%
\end{comment}

% 5. Organization of the paper

%The paper is organized as follows.
%

\section{Preliminaries} \label{sec:prelim}

\paragraph{Computational Complexity}
The paper addresses a number of well-known classes of decision and
function (or search) problems. For decision problems, these include P,
NP, $\ddp$, $\stwop$, %$\tn{FP}^{\tn{NP}}$, $\tn{NP}^{\tn{PP}}$,
among others.
For function problems, we will also consider standard classes,
including FP, FNP, among others.
(The interested reader is referred to a standard
reference on computational complexity~\cite{arora-bk09}.)

%%
%% Logic foundations
%%

\subsection{Logic Foundations} \label{ssec:logic}

\jnoteF{%
  \begin{enumerate}
  \item Formulas
  \item Assignments and models
  \item Inconsistent formulas: MUSes, MCSes, MSSes
  \item Optimization problems
  \end{enumerate}
}

Throughout this section, we adopt notation and definitions from
standard
references~\cite{papadimitriou-bk94,bradley-cc07,crama-bk11,mlcs-bk12,mc-handbook18,sat-handbook21}.

\subsubsection{Propositional Logic \& Boolean Satisfiability.}
\label{sssec:sat}

This section studies the decision problem for propositional logic, 
also referred to as the Boolean Satisfiability (SAT)
problem~\cite{sat-handbook21}. (The presentation follows standard
references, e.g.~\cite{buning-bk99,crama-bk11,sat-handbook21}.)
SAT is well-known to be an NP-complete~\cite{cook-stoc71} decision
problem, with algorithm implementations that can be traced to the
early 1960s~\cite{davis-jacm60,davis-cacm62}.

\paragraph{Syntax -- well-formed propositional formulas.}
We consider a set of propositional atoms $X=\{x_1,\ldots,x_n\}$
(these are also most often referred to as boolean variables), and
associate an index with each atom, i.e. $i$ with $x_i$,
$i=1,\ldots,n$, represented as the set $\mfrak{X}$.
A (well-formed) propositional formula, or simply a formula, is defined
inductively as follows:
\begin{enumerate}
\item An atom $x_i$ is a formula.
\item If $\varphi$ is a formula, then $\neg\varphi$ is a
  formula.
  (The logic operator $\neg$ is referred to as negation.) 
  %\item If $\varphi_1$ and $\varphi_2$ are formulas, then
  %  $\varphi_1\land\varphi_2$ is a formula.
\item If $\varphi_1$ and $\varphi_2$ are formulas, then
  $\varphi_1\lor\varphi_2$ is a formula.
  (The logic operator $\lor$ is referred to as disjunction.)
\item If $\varphi$ is a formula, then $(\varphi)$ is a formula.
\end{enumerate}
A literal is an atom $x_i$ or its negation $\neg{x_i}$.
We can use additional logic operators, defined in terms of the basic
operators above. Well-known examples include:
\begin{enumerate}
\item $\varphi_1\land\varphi_2$ represents the formula
  $\neg(\neg\varphi_1\lor\neg\varphi_2)$.
  (The logic operator $\land$ is referred to as conjunction.)
\item $\varphi_1\limply\varphi_2$ represents the formula
  $\neg\varphi_1\lor\varphi_2$.
  (The logic operator $\limply$ is referred to as implication.) 
\item $\varphi_1\lequiv\varphi_2$ represents the formula
  $(\varphi_1\limply\varphi_2)\land(\varphi_2\limply\varphi_1)$.
  (The logic operator $\lequiv$ is referred to as equivalence. Also,
  we use $\limply$ in the definition for simplicity, and could just
  use the initial logic operators.)
\end{enumerate}
Parentheses are used to enforce precedence between operators.
Otherwise, the precedence between operators is:
$\neg,\land,\lor,\limply,\lequiv$.
A clause is a disjunction of literals.
A term is a conjunction of literals. For both clauses and terms, we
will disallow having either clauses or terms defined using an
atom and its negation, or repetition of literals. This is formalized
as follows.
Given $A,B\subseteq\mfrak{X}$, with $A\cap{B}=\emptyset$, a clause
is defined by,
\[\omega_{A,B}\triangleq\left(\lor_{i\in{A}}x_i\lor\lor_{i\in{B}}\neg{x_i}\right)\]
and a term is defined by,
\[\tau_{A,B}\triangleq\left(\land_{i\in{A}}x_i\land\land_{i\in{B}}\neg{x_i}\right)\]
when clear from the context, we will drop the subscript from
$\omega_{A,B}$ and $\tau_{A,B}$.
$\mbb{W}$ denotes the set of all possible clauses defined
on the set of atoms $X$.
Similarly, $\mbb{T}$ denotes the set of all possible terms
defined on the set of atoms $X$.
A conjunctive normal form (CNF) formula is a conjunction of clauses.
A disjunctive normal form (DNF) formula is a disjunction of terms.

To simplify some of the subsequent definitions in this section,
clauses and terms will also be viewed as sets of literals, each CNF
formula as a set of clauses, and each DNF formula as a set of
terms.

\paragraph{Semantics -- assignments \& valuations.}
An assignment is any point in $\mbf{u}\in\{0,1\}^n$. (Throughout, we
associate 0 with both \FALSE and $\bot$, and 1 with both \TRUE and
$\top$.)
The actual value ascribed to a propositional formula is derived from
the assignment of propositional values to the formula's atoms. Each
such complete assignment is referred to as an \emph{interpretation}.
Given a formula $\varphi$, and an assignment $\mbf{u}\in\{0,1\}^n$,
the valuation of $\varphi$ given $\mbf{u}$ is represented by
$\varphi^{\mbf{u}}$,
%%$\nu(\varphi,\mbf{u})$,
and it is defined inductively as follows:
\begin{enumerate}
\item
  $\varphi^{\mbf{u}}=1$
  %%$\nu(\varphi,\mbf{u})=1$
  if $\varphi=x_i$ and $u_i=1$ or $\varphi=\neg{x_i}$ and $u_i=0$.
\item
  $\varphi^{\mbf{u}}=0$
  %%$\nu(\varphi,\mbf{u})=0$
  if $\varphi=x_i$ and $u_i=0$ or $\varphi=\neg{x_i}$ and $u_i=1$.
\item
  $\varphi^{\mbf{u}}=\neg(\psi^{\mbf{u}})$,
  if $\varphi=\neg\psi$.
%\item
%  $\varphi^{\mbf{u}}=\psi_1^{\mbf{u}}\land\psi_2^{\mbf{u}}$,
%  %%$\nu(\varphi,\mbf{u})=\nu(\psi_1,\mbf{u})\land\nu(\_psi_2,\mbf{u})$,
%  if $\varphi=\psi_1\land\psi_2$.
\item
  $\varphi^{\mbf{u}}=\psi_1^{\mbf{u}}\lor\psi_2^{\mbf{u}}$,
  %%$\nu(\varphi,\mbf{u})=\nu(\psi_1,\mbf{u})\lor\nu(\_psi_2,\mbf{u})$,
  if $\varphi=\psi_1\lor\psi_2$.
\item
  $\varphi^{\mbf{u}}=(\psi^{\mbf{u}})$,
  %%$\nu(\varphi,\mbf{u})=\nu(\psi_1,\mbf{u})\lor\nu(\_psi_2,\mbf{u})$,
  if $\varphi=(\psi)$.
\end{enumerate}
Clearly, for any $\mbf{u}\in\{0,1\}^n$, $\varphi^{\mbf{u}}\in\{0,1\}$.
Also, it is plain to extend the semantics to the other logic
operators: $\land,\limply,\lequiv$.
A propositional formula $\varphi$ can also be viewed as representing a
boolean function that maps $\{0,1\}^n$ to $\{0,1\}$. The same symbol 
will be used to refer to both formula and function,
i.e.\ $\varphi:\,\,\{0,1\}^n\to\{0,1\}$. Given some assignment
$\mbf{u}\in\{0,1\}^n$, it is the case that
$\varphi(\mbf{u})=\varphi^{\mbf{u}}$.

Given a formula $\varphi$,  $\mbf{u}\in\{0,1\}$ is a \emph{model} of
$\varphi$ if it makes $\varphi$ \TRUE,
i.e.\ $\varphi^{\mbf{u}}=1$.
A formula $\varphi$ is \emph{satisfiable} or \emph{consistent}
(represented by $\varphi\nentails\perp$) if it admits a model;
otherwise, it is \emph{unsatisfiable} or \emph{inconsistent}
(represented by $\varphi\entails\perp$).

Practical reasoners for the SAT problem represent a success story of 
Computer Science~\cite{vardi-cacm10a,vardi-cup20}. Modern
conflict-driven clause learning (CDCL) SAT reasoners routinely decide
formulas with millions of variables and tens of millions of
clauses~\cite[Chapter~4]{sat-handbook21}. (This success hinges on the
paradigm of learning clauses from search
conflicts~\cite{mss-iccad96,mss-tcomp99}.)
Furthermore, SAT reasoners are the underlying engine used to achieve
significant performance gains in different areas of automated
reasoning, including
different boolean optimization
problems~\cite[Chapters~23,~24,~28]{sat-handbook21}, 
answer-set programming~\cite{schaub-bk12},
constraint programming~\cite{stuckey-cj09},
quantified boolean formulas~\cite[Chapters~30,~31]{sat-handbook21},
but also for reasoners for fragments of first order
logic~\cite[Chapter~33]{sat-handbook21}
and theorem
proving~\cite{korovin-lics03,korovin-ijcar08,voronkov-cav13,voronkov-cav14}.
SAT reasoners are often publicly available and their performance
improvements are regularly
assessed\footnote{\url{http://www.satcompetition.org/}.}.
There exist publicly available toolkits that streamline prototyping
with SAT reasoners. At the time of writing, the reference example is
PySAT~\cite{imms-sat18}.

It should be noted that the high performance reasoners mentioned above
most often require logic formulas represented in CNF (or in clausal
form). There are well-known, efficient, procedures for converting
arbitrary (non-clausal) logic formulas into clausal
form~\cite{tseitin-scmml68,plaisted-jsc86}. Given a propositional
formula $\varphi$, defined on a set of propositional atoms $X$,
$\psi=\llencode{\varphi}{}$ is the \emph{clausification} of $\varphi$,
such that $\psi$ is defined on a set of atoms $X\cup{A}$, where $A$
denotes additional auxiliary atoms. One important result is that
$\varphi$ and $\psi$ are \emph{equisatisfiable}, i.e.\ $\varphi$ is 
satisfiable iff $\psi$ is satisfiable.
Encodings are detailed further in~\cref{ssec:encs}.

\paragraph{Entailment and equivalence.}
Given two formulas $\varphi$ and $\tau$, we say that $\tau$
\emph{entails} $\varphi$, denoted by $\tau\entails\varphi$, if,
\[
\forall(\mbf{u}\in\{0,1\}^n).\left[\tau^{\mbf{u}}\limply\varphi^{\mbf{u}}\right]
\]
which serves to indicate that any model of $\tau$ is also a model of
$\varphi$.
%
%%\[
%%\forall(\mbf{u}\in\{0,1\}^n.\left[\left(\nu(\tau,\mbf{u})=1\right)\limply\left(\nu(\tau,\mbf{u})=1\right)\right]
%%\]
We say that $\varphi_1\equiv\varphi_2$ iff
$\varphi_1\entails\varphi_2$ and $\varphi_2\entails\varphi_1$.

Let $\tau_{A,B}$ and $\tau_{C,D}$ be terms, with both $A,B$ and
$C,D$ representing disjoint pairs of subsets of $\mfrak{X}$.
Then, we have that,
\begin{proposition}
  $\tau_{A,B}\equiv\tau_{C,D}$ iff $A=C$ and $B=D$.
\end{proposition}

\begin{proposition}
  If $\tau_{A,B}\entails\tau_{C,D}$ and
  $\tau_{A,B}\not\equiv\tau_{C,D}$, then
  $C\subseteq{A}$,
  $D\subseteq{B}$, and
  $C\subsetneq{A}$ or
  $D\subsetneq{B}$.
\end{proposition}
Clearly, similar results can be stated for clauses.

\begin{example}
  The terms $x_1\land\neg{x_2}\land{x_3}$  and $x_1\land\neg{x_2}$ are
  represented, respectively, by $\tau_{A,B}$ and $\tau_{C,D}$, with
  $A=\{1,3\}$, $B=\{2\}$, $C=\{1\}$ and $D=\{2\}$.
  %%$\tau_{\{1\},\{2\}}$, $\tau_{\{1,3\},\{2\}}$.
  It is the case that
  $x_1\land\neg{x_2}\land{x_3}\entails{x_1}\land\neg{x_2}$.
  As can be concluded, $C\subsetneq{A}$ and $D\subseteq{B}$.
\end{example}  

\paragraph{Prime implicants and implicates.}
%~\\
Let $\varphi$ represent a propositional formula, and let
$\tau_{A,B}$ represent a term. $\tau_{A,B}$ is a \emph{prime
implicant} of $\varphi$ if,
\begin{enumerate}
\item $\tau_{A,B}\entails\varphi$.
\item For any other term $\tau_{C,D}$, such that
  $\tau_{A,B}\entails\tau_{C,D}$ and
  $\tau_{A,B}\not\equiv\tau_{C,D}$, then $\tau_{C,D}\nentails\varphi$.
\end{enumerate}
Whenever the first condition holds
(i.e.\ $\tau_{A,B}\entails\varphi$), then we say that $\tau_{A,B}$ is
an \emph{implicant} of $\varphi$.

\begin{example}
  Consider the terms $\tau_{A,B}=x_1\land{x_2}$ and $\tau_{C,D}=x_1$.
  (Note that $A=\{1,2\},B=\emptyset,C=\{1\},D=\emptyset$.)
  Let
  $\varphi=(x_1\land{x_2}\land{x_3})\lor({x_1}\land{x_2}\land\neg{x_3})$.
  Clearly, for any assignment $\mbf{u}\in\{0,1\}^3$ to $x_1,x_2,x_3$,
  if $x_1=x_2=1$, then $\tau_{A,B}=1$ and also $\varphi=1$,
  independently of the value of $x_3$. Hence,
  $\tau_{A,B}\entails\varphi$.
  However,
  $\tau_{C,D}\nentails\varphi$, in that there are assignments
  $\mbf{u}$ to $x_1,x_2,x_3$ such that $\tau_{C,D}^{\mbf{u}}=1$ but
  $\varphi^{\mbf{u}}=0$. For example, whenever $x_2=0$, then $\varphi$
  takes value 0, and that is not necessarily the case with
  $\tau_{C,D}$. Proving that $\tau_{A,B}$ is a prime implicant of
  $\varphi$ would apparently require proving that any term with
  literals that are a proper subset of $\tau_{A,B}$ are not implicants
  of $\varphi$. As discussed below, in practice one can devise more
  efficient algorithms.
\end{example}

For completeness, we also mention prime implicates. Let $\varphi$ be a
propositional formula, and let $\omega_{A,B}$ be a clause.
$\omega_{A,B}$ is a \emph{prime implicate} of $\varphi$ if,
\begin{enumerate}
\item $\varphi\entails\omega_{A,B}$.
\item For any other clause $\omega_{C,D}$, such that
  $\omega_{C,D}\entails\omega_{A,B}$ and
  $\omega_{A,B}\not\equiv\omega_{C,D}$, then $\varphi\nentails\tau_{C,D}$.
\end{enumerate}
Whenever the first condition holds
(i.e.\ $\varphi\entails\omega_{A,B}$), then we say that $\omega_{A,B}$
is an \emph{implicate} of $\varphi$.

%\begin{example}
%
%\end{example}

A well-known result (which can be traced to~\cite{rymon-amai94}) is
that, for a propositional formula $\varphi$, prime implicants are
minimal hitting sets (MHSes) of the prime implicates and
vice-versa\footnote{%
Recall that a set $\fml{H}$ is a \emph{hitting set} of a set of
sets $\fml{S}=\{S_1,\ldots,S_k\}$ if $\fml{H}\cap{S_i}\not=\emptyset$
for $i=1,\ldots,k$. $\fml{H}$ is a minimal hitting set of $\fml{S}$,
if $\fml{H}$ is a hitting set of $\fml{S}$, and there is no proper
subset of $\fml{H}$ that is also a hitting set of $\fml{S}$.}. This
result is at the core of recent algorithms for enumerating prime
implicants and implicates of a propositional formula
$\varphi$~\cite{pimms-ijcai15}.
%\begin{proposition}
%\end{proposition}
  
Given a propositional formula $\varphi$ and a term $\tau$, with
$\tau\entails\varphi$, a prime implicant $\pi$, with
$\tau\entails\pi$, can be computed with at most a linear number of
calls to an NP
oracle~\cite{umans-tcad06,bradley-fmcad07,pimms-ijcai15}.
(The same observations apply to the case of prime implicates.)
In~\cref{sec:cxps}, we will revisit some of these results when 
computing formal explanations.

\begin{comment}
%
(...)\\
A term $\tau$ is a conjunction of literals\footnote{%
To simplify the notation, $\tau$ is also viewed as a set of
literals.}.
%
A term is consistent if there exists at least one
$\mbf{u}\in\{0,1\}^n$ such that $\nu(\tau,\mbf{u})=1$, and where
$\nu(\tau,\mbf{u})$ is the valuation of $\tau$ given $\mbf{u}$, and
is defined inductively as follows:
\begin{enumerate}
\item $\nu(\tau,\mbf{u})=1$ if $\tau=x_i$ and $u_i=1$ or
  $\tau=\neg{x_i}$ and $u_i=0$.
\item $\nu(\tau,\mbf{u})=0$ if $\tau=x_i$ and $u_i=0$ or
  $\tau=\neg{x_i}$ and $u_i=1$.
\item
  $\nu(\tau,\mbf{u})=\nu(\psi_1,\mbf{u})\land\nu(\_psi_2,\mbf{u})$,
  if $\tau=\psi_1\land\psi_2$.
\end{enumerate}  
%
Let $\mbb{T}$ denote the set of all possible consistent terms
defined on the set of atoms $X$.

A term $\tau\in\mbb{T}$ is a \emph{prime implicant} of some formula
$\varphi$ if,
\[
\tau\entails\varphi\land\forall_{\tau'\in\mbb{T}\land\tau'\not\equiv\tau}
\left(\tau'\entails\tau\right)\limply\left(\tau'\nentails\varphi\right)
\]
%Given a function $\kappa:\{0,1\}^n\to\{0,1\}$, a term
%$\tau\in\mbb{T}$ is a prime implicant of $\kappa$ if,
%~\\
%%\jnoteF{Define PIs and IPs.}
%
\end{comment}

\paragraph{Reasoning about inconsistency.}
In many situations, there is the need to reason about inconsistent
formulas~\cite{msm-ijcai20}. For example, we may be interested in
explaining the reasons of inconsistency, but we may also be interested
in identifying which clauses to ignore (or equivalently to remove from
the formula) so as to restore consistency\footnote{%
  This paragraph aims at brevity. However, there are recent
  up-to-date treatments of these topics~\cite{msm-ijcai20}.}.
The general setting is to consider a set of clauses
$\fml{B}$. $\fml{B}$ represents some background knowledge base, and so
it is assumed to be consistent. We say that $\fml{B}$ contains the
\emph{hard} clauses (or in general the hard constraints). A clause is
hard when it cannot be removed (from the set of clauses) to recover
consistency. Furthermore, we also consider a set clauses
$\fml{S}$, such that
$\varphi_{\fml{B},\fml{S}}=\fml{B}\cup\fml{S}\entails\perp$, and
$\varphi_{\fml{B},\fml{S}}$ is the formula (or set of constraints) we
want to reason about. The clauses in $\fml{S}$ represent the
\emph{soft} clauses (or constraints), and these can be removed to
restore consistency. (In the rest of this document, we will just refer
to $\varphi$, being implicit that $\varphi$ is characterized by the 
background knowledge $\fml{B}$ and by the soft clauses $\fml{S}$.)

\begin{definition}[MUS] \label{def:mus}
  Let $\varphi=\fml{B}\cup\fml{S}$ be an inconsistent set of
  clauses (or constraints), i.e.\ $\varphi\entails\bot$.
  $\fml{M}\subseteq\fml{S}$ is a {\em Minimal Unsatisfiable Subset}
  (MUS) if $\fml{B}\cup\fml{M}\entails\bot$ and
  $\forall_{\fml{M}'\subsetneq\fml{M}},\,\fml{B}\cup\fml{M}'\nentails\bot$. 
\end{definition}
Informally, an MUS provides some irreducible information that suffices
to be added to the background knowledge $\fml{B}$ to attain an
inconsistent formula; thus, an MUS represents an explanation for the
causes of inconsistency. Alternatively, one might be interested in
correcting the formula, removing some clauses to achieve consistency. 
\begin{definition}[MCS] \label{def:mcs}  
  Let $\varphi=\fml{B}\cup\fml{S}$ be an inconsistent set of
  clauses ($\varphi\entails\bot$).
  $\fml{C}\subseteq\fml{S}$ is a {\em Minimal Correction Subset} (MCS)
  if $\fml{B}\cup\fml{S}\setminus\fml{C}\nentails\bot$ and
  $\forall_{\fml{C}'\subsetneq\fml{C}}$, $\fml{B}\cup\fml{S}\setminus\fml{C}'\entails\bot$.
\end{definition}
With each MCS $\fml{C}$, one associates a Maximal Satisfiable Subset
(MSS), given by $\fml{S}\setminus\fml{C}$.

\begin{example} \label{ex:musmcs}
  Let
  $c_1=(x_1)$, $c_2=(x_2)$, $c_3=(x_3)$,
  $c_4=(\neg{x_1}\lor\neg{x_2})$, and $c_5=(\neg{x_1}\lor\neg{x_3})$.
  Moreover, let $\varphi=\fml{B}\cup\fml{S}$, with
  $\fml{B}=\{c_4,c_5\}$ and $\fml{S}=\{c_1,c_2,c_3\}$.
  Hence, it is simple to conclude that an example of an MUS is
  $\{c_1,c_2\}$, an example of an MCS is $\{c_1\}$, and an example of
  an MSS is $\{c_2,c_3\}$.
\end{example}

Let $\msf{MUS}$, be a predicate, $\msf{MUS}:2^{\fml{S}}\to\{0,1\}$,
such that $\msf{MUS}(\fml{M})=1$ iff $\fml{M}\subseteq\fml{S}$ is an
MUS of $\varphi=\fml{B}\cup\fml{S}$.
Moreover, let $\msf{MCS}$, be a predicate, $\msf{MCS}:2^{\fml{S}}\to\{0,1\}$,
such that $\msf{MCS}(\fml{C})=1$ iff $\fml{C}\subseteq\fml{S}$ is an
MCS of $\varphi=\fml{B}\cup\fml{S}$.
Furthermore, we define,
\begin{align}
   \mbb{MU}(\varphi) & = \{\fml{M}\subseteq\fml{S}\,|\,\msf{MUS}(\fml{M})\} \\
   \mbb{MC}(\varphi) & = \{\fml{C}\subseteq\fml{S}\,|\,\msf{MCS}(\fml{M})\} %%\\
\end{align}

Moreover, there exists a well-known (subset-)minimal hitting set
relationship between MUSes and MCSes:
%~\cite{reiter-aij87,lozinskii-jetai03}:
%
\begin{proposition}
  Given $\varphi=\fml{B}\cup\fml{S}$, and $\fml{M}\subseteq\fml{S}$,
  $\fml{C}\subseteq\fml{S}$, then
  \begin{enumerate}
  \item $\fml{M}$ is an MUS iff it is a minimal hitting set of the 
    MCSes in $\mbb{MC}$;
  \item $\fml{C}$ is an MCS iff it is a minimal hitting set of the 
    MUSes in $\mbb{MU}$.
  \end{enumerate}
\end{proposition}
The MHS relationship between MUSes and MCSes was first demonstrated in
the context of model-based diagnosis~\cite{reiter-aij87} and later
investigated for %in the context of 
propositional formulas in clausal
form~\cite{lozinskii-jetai03}. As immediate from the original
work~\cite{reiter-aij87}, the MHS relationship applies in general to
systems of constraints, where each is represented as a first-order
logic statement.

\begin{example}
  For the formula from~\cref{ex:musmcs}, it is immediate to conclude
  that,
  \[
  \begin{array}{l}
    \mbb{MU}(\varphi)=\{\{c_1,c_2\},\{c_1,c_3\}\}\\
    \mbb{MC}(\varphi)=\{\{c_1\},\{c_2,c_3\}\}\\
  \end{array}
  \]
  Moreover, one can observe that each MUS is an MHS of the MCSes, and
  that each MCS is an MHS of the MUSes.
\end{example}

Complexity-wise, deciding whether a set of clauses is an MUS is
known to be
$\ddp$-complete~\cite{papadimitriou-jcss88,papadimitriou-bk94}.
It is also well-known that an MUS can be computed with at most a
number of calls to an NP-oracle that grows linearly with the number of
clauses in the worst-case~\cite{chinneck-jc91}.
MUSes and MCSes are tightly related with optimization
problems~\cite{msm-ijcai20}. For example, if each soft constraint is
assigned a unit cost, then solving the maximum satisfiability problem
(MaxSAT) corresponds to finding a maximum cost
MSS~\cite{mhlpms-cj13,sat-handbook21}.
Under the assumption of a unit cost assigned to each clause, then an
MCS can be computed with a logarithmic number of calls to an
NP-oracle, e.g.\ by solving (unweighted) MaxSAT. (If non-unit costs
are assumed, then a worst-case linear number of calls to an NP-oracle
is required.)
However, in practice there are more efficient algorithms that may
require in the worst-case a number of calls to an NP-oracle larger
than logarithmic~\cite{mshjpb-ijcai13,mpms-ijcai15,mipms-sat16}.
There have been very significant improvements in the practical
performance of MaxSAT solvers in recent
years~\cite[Chapters~23,~24,~28]{sat-handbook21}.
A number of solvers for MaxSAT and related problems are publicly
available\footnote{%
\url{https://maxsat-evaluations.github.io/} and
\url{http://www.cril.univ-artois.fr/PB16/}
\url{http://www.satcompetition.org/2011/} (MUS track).}.

\jnoteF{Basics of formulas, models, MUSes, MCSes, etc.}

\paragraph{Quantification problems.}
It is often of interest to quantify some of the variables in a
propositional formula (by default it is assumed that all variables are
existentially quantified).
This is achieved by using two more logic operators, namely $\forall$
for universal quantification and $\exists$ for existential
quantification.
QBF (Quantified Boolean Formulas) is the problem of deciding whether a
quantified propositional formula is true or false. The problem of
deciding QBF is PSPACE-complete~\cite{arora-bk09}.
In this paper, we will briefly study quantified problems with two
levels of quantifier alternation, concretely $\exists\forall$, which
is a well-known $\stwop$-complete decision problem.
There have been very significant improvements in the practical
performance of QBF solvers in recent
years~\cite[Chapters~30,~31]{sat-handbook21}.
A number of solvers for QBF are publicly available\footnote{%
\url{http://www.qbflib.org/}.}.

\paragraph{Logic-based abduction.}
Abductive reasoning can be traced to the work of
C.~Peirce~\cite{peirce-works31}, with its first uses in artificial 
intelligence in the early 1970s~\cite{morgan-aij71,pople-ijcai73}.
Logic-based abduction can be viewed as the problem of finding a
(minimimum or minimal) subset of hypotheses, which is consistent with
some background theory, and such that those hypotheses are sufficient
for some manifestation.
A propositional abduction problem (PAP) is represented by a 5-tuple
\papdef~\cite{jarvisalo-kr16,imms-ecai16}. $X$ is a finite set of
atoms. $\fml{H}$, $\fml{M}$ and $\fml{T}$ denote propositional
formulas representing, respectively, the set of hypotheses, the set of
manifestations, and the background theory. ($\fml{H}$ is further
constrained to be a set of clauses.) $\varsigma$ is a function that
associates a cost with each clause of $\fml{H}$,
$\varsigma\::\:\fml{H}\to\mathbb{R}^{+}_0$.
Given the background theory $\fml{T}$, a set $\fml{E}\subseteq\fml{H}$
of hypotheses is an explanation (for the manifestations) if: (i)
$\fml{E}$ entails the manifestations $\fml{M}$ (given $\fml{T}$),
i.e.\ $\fml{T}\land\fml{E}\entails\fml{M}$; and (ii)
$\fml{E}$ is consistent (given $\fml{T}$), i.e.\
$\fml{T}\land\fml{E}\nentails\bot$.
%%$T\land S\nentails\bot$; $T\land S\entails M$.
%
The propositional abduction problem is usually defined as computing a
minimum cost (or cardinality minimal) explanation for the
manifestations subject to the background theory. Moreover, one can
consider explanations of (subset-)minimal cost.
The complexity of logic-based abduction has been studied in the
past~\cite{bylander-aij91,gottlob-jacm95}. There are also recent
practical algorithms for propositional
abduction~\cite{jarvisalo-kr16,imms-ecai16}. 

%\begin{definition}[Explanations for $P$~\cite{jarvisalo-kr16}]
%Let \papdef be a PAP. The set of explanations of $P$ is given by the
%set $\tn{Expl}(P) = \{S\subseteq H\:|\: T\land S\nentails\bot ,
%T\land S\entails M\}$.
%%
%The minimum-cost solutions of $P$ are given by $\tn{Expl}_c(P) =
%\tn{argmin}_{E\in\tn{Expl}(P)}(c(E))$.
%\end{definition}

The computation of a prime implicant of some propositional formula
$\varphi$, defined on atoms $X$, can be formulated as a problem of
abduction.
Let $\mbf{u}$ be an assignment to the atoms of $\varphi$, such that
$\varphi^{\mbf{u}}=1$. Given $\mbf{u}$, construct the set $\fml{H}$ of
hypotheses as follows: if $u_i=1$, then add $x_i$ to $\fml{H}$,
otherwise add $\neg{x_i}$ to $\fml{H}$. 
We let $\fml{T}=\emptyset$ and $\fml{M}=\varphi$.
Given the definition of $\fml{H}$, then a (subset-)minimal set
$\fml{E}\subseteq\fml{H}$ is a prime implicant if,
(i) $\fml{E}\entails\varphi$; and (ii) $\fml{E}\nentails\bot$.
Observe that, by hypothesis, condition (ii) is
trivially satisfied. Hence, a prime implicant of $\varphi$ (given
$\fml{H}$) is a subset-minimal set of literals
$\fml{E}\subseteq\fml{H}$ such that $\fml{E}\entails\varphi$.

In practice, when $\varphi$ is non-clausal, deciding entailment is
somewhat more complicated. In these cases, and as mentioned earlier,
most often one needs to clausify $\varphi$, so that it can be reasoned
about. A difficulty with efficient clausification procedures is that
these introduce auxiliary variables. As a result, we need to follow a
different approach for computing a prime implicant.

Let $\psi=\llencode{\varphi}{\fml{V}}$ be the propositional clausal
representation of $\varphi$, given some logic theory $\fml{V}$, which
uses additional auxiliary atoms represented as set ${A}$.
(\cref{ssec:encs} details further the use of logic encodings.)
We distinguish an auxiliary propositional atom $t\in{A}$, such that
$t=1$ for the assignments to $X$ and $A$ which satisfy $\psi$ (and so
$\varphi$).
In this new setting, we let $\fml{T}=\psi$ and $\fml{M}=(t)$.
Given a (consistent) set of literals $\fml{H}$, representing
a satisfying assignment to the atoms of $\varphi$, then a
(subset-)minimal set $\fml{E}\subseteq\fml{H}$ is a prime implicant
if,
(i) $\psi\land\fml{E}\entails{t}$; and (ii)
$\fml{E}\nentails\bot$.
As before, condition (ii) is trivially satisfied. Hence, a prime
implicant of $\varphi$ (given $\fml{H}$) is a subset-minimal set of
literals $\fml{E}\subseteq\fml{H}$ such that
$\psi\land\fml{E}\entails{t}$.

\jnoteF{%
  \begin{enumerate}
  \item Definition of abduction: minimum and minimal;
  \item Relationship with prime-implicants
  \end{enumerate}
}

\paragraph{Propositional Languages.}
A propositional language represents a subset of the set of
propositional formulas, and several such subsets have been extensively
studied~\cite{darwiche-jacm01,darwiche-jancl01,darwiche-jair02}\footnote{%
  For brevity, we will not delve into defining propositional languages
  and queries/transformations of interest. The interested reader is
  referred to the bibliography~\cite{darwiche-jair02}.}.
One well-known example is negation normal form (NNF), representing a
directed-acyclic graph of $\land$ and $\lor$ operators, where the
leaves are atomic propositions or their negation. Other well-known
examples are DNF and CNF formulas. By imposing additional constraints
on the $\lor$ and $\land$ nodes of NNF formulas, one can devise
classes of propositional languages that exhibit important tractability
properties. A detailed analysis of propositional languages is
available in~\cite{darwiche-jair02}.
Some results on formal explainability have been derived for
propositional languages in recent
years~\cite{darwiche-ijcai18,darwiche-ecai20,marquis-kr20,marquis-kr21,hiicams-aaai22}.

\jnoteF{Foundations of KC as well}

\subsubsection{First Order Logic.}
\label{sssec:fol}
This section briefly mentions one well-known extension of
propositional logic, namely first order logic (FOL). FOL extends
propositional logic with predicates, functions, constants and
quantifiers, and such that variables are allowed to take values from
arbitrary domains.
In contrast with propositional logic, where an interpretation is an
assignment of values to the formula's atoms, in the case of FOL, an
interpretation must ascribe a meaning to predicates, functions and
constants.
Whereas validity in FOL is undecidable~\cite{mlcs-bk12}, one can
reason in concrete theories, with a well-known example being
satisfiability modulo theories.

%\subsubsection{Satisfiability Modulo Theories (SMT).}
\paragraph{Satisfiability modulo theories (SMT).}
%
%\subsubsection{First Order Logic (FOL) \& Satisfiability Modulo Theories (SMT).}
%\paragraph{First order logic (FOL) \& satisfiability modulo theories (SMT).}
%~\\
By providing first order logic with concrete theories, e.g.\ integer
arithmetic, real arithmetic or mixed integer-real arithmetic, among
many other possibilities, one obtains decidable fragments, for which
practically efficient reasoners have been developed over the last two
decades. (There exist undecidable theories in
SMT~\cite{kroening-bk16}, but that is beyond the goals of this
document.
SMT solvers generalize SAT to reason with fragments of first order
logic~\cite{kroening-bk16,mc-handbook18,sat-handbook21}.
Throughout this paper, we will use SMT reasoners solely as an
alternative for mixed-integer linear programming reasoners (see
below), even though SMT reasoners allow for significantly more general
fragments of FOL.
Similar to the case of SAT, we can solve optimization problems over
SMT formulas (MaxSMT), and we can also reason about inconsistency.
Previous definitions (see~\cref{sssec:sat}) also apply in this
setting.

%\subsubsection{Mixed Integer Linear Programming (MILP).}
\paragraph{Mixed integer linear programming (MILP).}
MILP can be formulated as a first-order logic theory
(e.g.\ \cite{bradley-cc07}, where variables can take values from
boolean, integer and real domains, and where the allowed binary
functions are $+$ and $-$, with their usual meanings, and the allowed
binary predicates are $=$ and $\le$, also with their usual
meanings. We will also allow a countable number of unary constant
functions $b\cdot$, with $b\in\mbb{R}$ (thus accounting for the other 
possible cases of $\mbb{B}$ and $\mbb{Z}$), and where each $b\cdot$
represents a coefficient. 
Starting from a set $V$ of (variable) numbers,
i.e.\ $V=\{1,\ldots,m\}$, we consider a partition of $V$ into $B$, $I$
and $R$. The general MILP formulation is thus:
\[
\begin{array}[t]{lcl}
  \tn{min} & \quad\quad & \sum_{j=1}^{n}c_jx_j \\[5pt]
  \tn{s.t.} & & 
  \begin{array}[t]{l}
    \sum_{j=1}^{n} a_{ij}x_j\le{b_i}, i=1,\ldots,r\\[1.5pt]
    x_j\in\{0,1\}, j\in{B}\\
    x_j\in\mbb{Z}, j\in{I}\\
    x_j\in\mbb{R}, j\in{R}\\
    a_{ij},b_i\in\mbb{R}
  \end{array}
\end{array}
\]
Several proprietary and publicly available MILP solvers are available%
\footnote{%
For example,
\url{https://www.ibm.com/ae-en/analytics/cplex-optimizer},
\url{https://www.gurobi.com/},
\url{https://sourceforge.net/projects/lpsolve/}.}, with significant
performance gains reported over the
years~\cite{bixby-or02,bixby-aor07,berthold-eurojco22}.

%\subsubsection{Additional Definitions.}
\paragraph{Additional definitions.}
The definitions introduced in the propositional logic case can be
generalized to the case of FOL, SMT, MILP, etc.
These generalizations include entailment, prime implicants and
implicates, but also the definitions associated with reasoning about
inconsistency.
A discussion of some of these concepts beyond propositional logic is
available for example in~\cite{msm-ijcai20}.

\subsubsection{Encodings \& Interfacing Reasoners.} \label{ssec:encs}
%\paragraph{Encodings and interfacing reasoners.}
%
Throughout the document, we will extensively refer to SAT, SMT and
MILP reasoners. Consistency checking with a reasoner for theory
$\fml{T}$ on a $\fml{T}$-theory formula $\varphi_{\fml{T}}$ is
represented by $\textco(\varphi_{\fml{T}};\fml{T})$, and denotes
whether $\varphi_{\fml{T}}$ has at least one model (given $\fml{T}$),
i.e.\ an interpretation that satisfies $\varphi_{\fml{T}}$. For
simplicity, the parameterization on $\fml{T}$ is omitted, and so we
use $\textco(\varphi_{\fml{T}})$ instead.
These theory reasoners operate on formulas of a suitable logic
language.
Given some logic formula $\varphi$, 
$\llbracket\varphi\rrbracket_{\fml{T}}$
denotes the encoding of $\varphi$ in a representation suitable for
reasoning by a decision oracle for theory~$\fml{T}$. (For simplicity,
we just use $\llbracket\varphi\rrbracket$.) 
As shown below, the computation of formal explanations assumes the
existence of a reasoner that decides the satisfiability (or
consistency) of a statement expressed in theory~$\fml{T}$.

For the case of propositional theories, SAT reasoners most often work
with clausal representations. As noted earlier, there exist
procedures for converting arbitrary (non-clausal) logic formulas into
clausal form~\cite{tseitin-scmml68,plaisted-jsc86}, which require the
use of additional propositional atoms.
There are also well-known encodings of constraints into clausal
form~\cite[Chapters~02,~28]{sat-handbook21}. Examples include
cardinality constraints, e.g.\ AtMost$K$ (i.e.\ $\sum_{i}x_i\le{K}$,
with boolean $x_i$) or AtLeast$K$ (i.e.\ $\sum_{i}\ge{k}$, with
boolean $x_i$) constraints, and pseudo-boolean constraints
($\sum{i}a_ix_i\le{b}$, also with boolean $x_i$), among many others.

%%
%% ML foundations
%%
\subsection{Classification Problems} \label{ssec:clf}

\jnoteF{%
  \begin{enumerate}
  \item Features, domains
  \item Classification function
  \end{enumerate}
}

Classification problems in ML are defined on a set of features (or
attributes) $\fml{F}=\{1,\ldots,m\}$ and a set of classes
$\fml{K}=\{c_1,c_2,\ldots,c_K\}$.
Each feature $i\in\fml{F}$ takes values from a domain $\fml{D}_i$.
In general, domains can be categorical or ordinal, with values that
can be boolean or integer. (Although real-valued could be considered
for some of the classifiers studied in the paper, we opt not to
specifically address real-valued features.)
The set of domains is represented by
$\mbb{D}=(\fml{D}_1,\ldots,\fml{D}_m)$.
Feature space is defined as
$\mbb{F}=\fml{D}_1\times{\fml{D}_2}\times\ldots\times{\fml{D}_m}$;
$|\mbb{F}|$ represents the total number of points in $\mbb{F}$.
%if none of the features is real-valued. 
%
For boolean domains, $\fml{D}_i=\{0,1\}=\fml{B}$, $i=1,\ldots,m$, and
$\mbb{F}=\fml{B}^{m}$.
The notation $\mbf{x}=(x_1,\ldots,x_m)$ denotes an arbitrary point in
feature space, where each $x_i$ is a variable taking values from
$\fml{D}_i$. The set of variables associated with features is
$X=\{x_1,\ldots,x_m\}$.
Moreover, the notation $\mbf{v}=(v_1,\ldots,v_m)$ represents a
specific point in feature space, where each $v_i$ is a constant
representing one concrete value from $\fml{D}_i$. 
With respect to the set of classes $\fml{K}$, the size of $\fml{K}$
is assumed to be finite; no additional restrictions are imposed on
$\fml{K}$. Nevertheless, with the goal of simplicity, the paper
considers examples where $|\fml{K}|=2$, concretely $\fml{K}=\{0,1\}$,
or alternatively $\fml{K}=\{\ominus,\oplus\}$.
An ML classifier $\fml{M}$ is characterized by a (non-constant)
\emph{classification function} $\kappa$ that maps feature space
$\mbb{F}$ into the set of classes $\fml{K}$,
i.e.\ $\kappa:\mbb{F}\to\fml{K}$.
Each classifier $\fml{M}$ is represented unambiguously by the tuple
$(\fml{F},\mbb{D},\mbb{F},\fml{K},\kappa)$. (With a mild abuse of
notation, we also write
$\fml{M}=(\fml{F},\mbb{D},\mbb{F},\fml{K},\kappa)$.)
An \emph{instance} (or observation)
denotes a pair $(\mbf{v}, c)$, where $\mbf{v}\in\mbb{F}$ and
$c\in\fml{K}$, with $c=\kappa(\mbf{v})$. 
%(We also use the term \emph{instance} to refer to $\mbf{v}$, leaving
%$c$ implicit.)

The \emph{classifier decision problem} (CDP) is to decide whether the
logic statement $\exists(\mbf{x}\in\mbb{F}).(\kappa(\mbf{x})=c)$, for
$c\in\fml{K}$, is true. 
%, i.e.\ does there exist at least one point $\mbf{x}$ in feature
%space for which the prediction is $c$?
Given some target class $c\in\fml{K}$, the goal of CDP is to decide
whether there exists some point $\mbf{x}$ in feature space for which
the prediction is $c$.
For example, for a neural network or a random forest, it is easy to
prove that CDP is NP-complete. In contrast, for univariate decision
trees, CDP is in P. This is further discussed in the next section.
%

%\subsubsection{Classifiers in Machine Learning}
\subsubsection{Examples of Classifiers.} % in Machine Learning
\label{ssec:mlclf}
%
%~\\

This paper studies decision trees (DTs), decision sets (DSs) and
decision lists (DLs) in greater detail.
Nevertheless, formal explainability has been studied in the context of
several other well-known families of classifiers, including naive
bayes classifiers (NBCs)~\cite{msgcin-nips20},
decision diagrams and graphs~\cite{hiims-kr21},
tree
ensembles~\cite{inms-corr19,ignatiev-ijcai20,ims-ijcai21,iisms-aaai22},
monotonic classifiers~\cite{msgcin-icml21,cms-cp21},
neural networks~\cite{inms-aaai19},
and bayesian network
classifiers~\cite{darwiche-ijcai18,darwiche-aaai19}.
Additional information can be found in the cited references.

%\paragraph{Decision trees, diagrams, graphs, lists \& sets.}
%%
%~\\

\paragraph{Decision trees (DTs).}
%~\\
DTs are among the still most widely used family of classifiers, with
applications in both ML and data mining
(DM)~\cite{breiman-bk84,quinlan-ml86,quinlan-bk93,kumar-bk09,flach-bk12}. %kumar-kis08
A decision tree is a directed acyclic graph, with one root node that
has no incoming edges, and the remaining nodes having exactly one
incoming edge. Terminal nodes have no outgoing edges, and non-terminal
nodes have two or more outgoing edges. Each terminal node is
associated with a class, i.e.\ the predicted class for the node. Each
non-terminal node is associated with exactly one feature\footnote{%
Thus, this paper only considers \emph{univariate} DTs, for which where
each non-terminal node test a single feature. In contrast, for
multivariable DTs~\cite{utgoff-ml95}, we assume that each non-terminal
node can test arbitrary functions of the features. For ordinal
features, multivariate DTs are also referred to as
\emph{oblique}~\cite{salzberg-jair94}.}. Each outgoing
edge is associated with a literal defined using the values of the
feature, and such that any value of the feature domain is consistent
with exactly one of the literals of the outgoing edges.
A tree path $\fml{P}$ connects the root node with one of the tree's
terminal nodes.
Common (implicit) assumptions of DTs are that: (i) all paths in a DT 
are consistent; and
(ii) for each point $\mbf{v}$ in feature space, there exists exactly 
one path $\fml{P}$ that is consistent with $\mbf{v}$.
(Observe that (ii) requires that the branches at each node capture all 
values in the domain of the tested feature and that the branches'
conditions be mutually disjoint.)
Given these assumptions of DTs, it is easy to see that CDP is in P.
One simply picks the target class and a terminal node predicting the
class, and reconstructs the path to the root; a procedure that runs in
linear time on the size of the tree.
An example of a DT is shown in~\cref{fig:02a:dt}. (This example will be
analyzed in greater detail below.)

\paragraph{Decision lists (DLs) and sets (DSs).}
%~\\
DLs and DSs also find a wide range of
applications~\cite{clark-ml89,flach-bk12,leskovec-kdd16,rudin-kdd17,rudin-jmlr17a}.
DLs and DSs represent, respectively, ordered and unordered rule sets.
There exist in-depth studies of DLs~\cite{rivest-ml87}, but in
contrast DSs are less well-understood.
Each rule is of the form:
\begin{equation*}
  \begin{array}{ccccccccc}
  \tn{R$_{j}$:} &\quad& \tn{IF~} &\quad& (\tau_j) &\quad& \tn{~THEN~} &\quad& d_j \\
  \end{array}
\end{equation*}
where $\tau_j$ represents a boolean expression defined on the features
and their domains, and $d_j\in\fml{K}$. We say that the rule
\emph{fires} if its literal ($\tau_j$) is consistent (or holds true).
For DLs, and with the exception of the first rule, all other rules are
of the form:
\begin{equation*}
  \begin{array}{ccccccccc}
  \tn{R$_{l}$:} &\quad& \tn{ELSE~IF~} &\quad& (\tau_l) &\quad& \tn{~THEN~} &\quad& d_l \\
  \end{array}
\end{equation*}
For the last (default) rule, it is required that $\tau_l$ is a
tautology, i.e.\ the rule always fires if all others do not. (This
basically corresponds to solely having ELSE as the rule's condition.)
The default rule is marked as $\tn{R}_{\tn{\sc{def}}}$.
An example of a DL is shown in~\cref{fig:01a:dl}. An example of a DS
would be the same DL, but without the ELSE's, i.e.\ there would be no
order among the listed rules. (The DL example will be analyzed in
greater detail below.)
In contrast with DLs, the lack of order of rules in DSs raises a
number of issues~\cite{ipnms-ijcar18}. One issue is \emph{overlap},
i.e.\ two or more rules predicting different classes may fire on the
same point of feature space. A second issue is \emph{coverage},
i.e.\ without a default rule, it may happen that no rule fires on some
points of feature space. It is conjectured that is is $\stwop$-hard to
learn DSs that ensure both no overlap and ensuring coverage of all
points in feature space~\cite{ipnms-ijcar18}.

\paragraph{Neural networks (NNs).}
%~\\
We consider a simple architecture for an NN, concretely feed-forward
NNs, which we refer to as NNs. (A comprehensive treatment of NNs
can be found elsewhere~\cite{bengio-bk16}.)
An NN is composed of a number of layers of neurons. The output values
of the neurons in a given layer $\mbf{x}^l$ are computed given the
output values of the neurons in the previous layer
$\mbf{x}^{l-1}$, up to a number $L$ of layers, and such that the
inputs represent layer 0. Furthermore, each neuron computes an
intermediate value given the output values of the
neurons in the previous layer, and the weights of the connections
between layers. For each layer, the intermediate computed values are
represented by $\mbf{y}^{l}$. The output value of each neuron is the
result of applying a non-linear activation function on the values of
$\mbf{y}^{l}$, thus obtaining $\mbf{x}^{l}$. Assuming a
ReLU~\cite{hinton-icml10} activation function, one obtained the
following:
\begin{equation} \label{eq:nndef}
  \begin{array}{l}
    \mbf{y}^{l} = \mbf{A}^{l}\cdot(\mbf{x}^{l-1})^{\tn{T}}\\[2.5pt]
    \mbf{x}^{l} = \max(\mbf{y}^{l},\mbf{0})\\
  \end{array}
\end{equation}
where $\mbf{x}^0$ denote the input values, $\mbf{x}^{L}$ denotes
the output values, and $\mbf{A}^{l}$ denotes the weights matrix (that
also accounts for a possible bias vector, by including a variable
$x_0^{l-1}=1$).
For classification problems, there are different mechanisms to predict
the actual class associated with the computed output values. One
option is to have each output represent a class. Another option is to
pick the class depending on the range of values taken by the output
variable. This alternative is illustrated with the running example
presented later in this section.

A recent alternative to NNs, namely binarized neural
networks (BNNs)~\cite{bengio-nips16} has also been investigated from
the perspective of explainability~\cite{nsmims-sat19}.

\paragraph{Monotonic classifiers.}
Monotonic classifiers find a number of important
applications, and have been studied extensively in recent
years~\cite{gupta-nips16,gupta-nips17,lin-nips20,vandenbroeck-nips20}.
Let $\preccurlyeq$ denote a partial order on the set of classes
$\fml{K}$. For example, we assume
$c_1\preccurlyeq{c_2}\preccurlyeq\ldots{c_K}$.
Furthermore, we assume that each domain $D_i$ is ordered such that the
value taken by feature $i$ is between a lower bound $\lambda(i)$ and
an upper bound $\mu(i)$. Given
$\mbf{v}_1=(v_{11},\ldots,v_{1i},\ldots,v_{1m})$ and
$\mbf{v}_2=(v_{21},\ldots,v_{2i},\ldots,v_{2m})$, we say that
$\mbf{v}_1\le\mbf{v}_2$ if,
$\forall(i\in\fml{F}).v_{1i}\le{v_{2i}}$.
Finally, a classifier is monotonic if whenever
$\mbf{v}_1\le\mbf{v}_2$, then
$\kappa(\mbf{v}_1)\preccurlyeq\kappa(\mbf{v}_2)$.

\paragraph{Additional families of classifiers.}
Formal explainability has been studied in the context of other
families of classifiers, including
random forests (RFs)~\cite{ims-ijcai21,mazure-cikm21},
boosted trees (BTs)~\cite{inms-corr19,ignatiev-ijcai20,iisms-aaai22},
tree ensembles (TEs, which include both RFs and BTs)~\cite{iisms-aaai22},
decision graphs (DGs)\& diagrams~\cite{hiims-kr21},
naive bayes classifiers (NBCs)~\cite{msgcin-nips20},
monotonic classifiers~\cite{msgcin-icml21},
propositional language classifiers~\cite{hiicams-aaai22}, and
bayesian network classifiers~\cite{darwiche-ijcai18,darwiche-aaai19}.
Most of these classifiers are covered in standard references on
ML~\cite{flach-bk12,shalev-shwartz-bk14,bengio-bk16}.

A random forest is represented by a set of decision trees, each tree
trained from a random sample of the original dataset. In the
originally proposed formulation of RFs~\cite{breiman-ml01}, the
selected class is picked by majority voting among all trees. As an
example, we show that CDP for RFs is NP-complete.

\begin{proposition}
  For RFs, CDP is NP-complete.
\end{proposition}

\begin{Proof}
  The decision problem is clearly in NP. Simply pick a point in
  feature space, and then compute in polynomial time the prediction of
  the RF, the decision problem answers \tbf{true} if the prediction is
  $c$, and it answers \tbf{false} if the prediction is other than
  $c$.\\
  To prove NP-hardness, we reduce the decision problem for
  propositional formulas represented in CNF, a problem well-known to
  be complete for NP. Let $\psi$ be a CNF formula with $m$
  propositional variables and $n$ clauses, $\cl_1,\ldots,\cl_n$:
  %Proof for RFs, with reduction of CNF with $m$ variables and $n$
  %clauses $c_1,\ldots,c_n$:
  \begin{itemize}[nosep]
  \item Create $n$ decision trees, one for each clause $\cl_j$, and
    such that DT $j$ predicts 1 if at least one literal of $\cl_j$ is
    satisfied, and 0 if all literals of $\cl_j$ are falsified.
  \item Also, create $n-1$ decision trees, each with a single terminal
    node with prediction 0.
  \end{itemize}
  Clearly, the reduction runs in polynomial time.
  Moreover, it is immediate that the RF picks class 1 if and only if
  the formula is satisfied. Let the assignment to $\psi$ be such that
  $\psi(x_1,\ldots,x_m)$ is satisfied. In this case, each DT
  associated with a clause will predict 1 and the other $n-1$ DTs will
  predict 0; hence the prediction will be 1.
  Furthermore, for the prediction to be 1, it must be the case that
  the $n$ DTs associated with the clauses must predict 1, since one
  must offset the $n-1$ trees that guaranteedly predict 0.
\end{Proof}

A simpler argument could be used to prove that CDP for multivariate
decision trees is also NP-complete. As an example, the CNF formula
could be tested on a single tree node.

\subsubsection{Running Examples.}
%
%%\paragraph{Running examples.}
%
Throughout the paper, the following running examples will be used
to illustrate some of the main results.

\begin{example}[DL]  \label{ex:runex01a}
  The first running example is a simple DL, that is
  adapted from~\cite[Section~6.1]{flach-bk12}. (The original
  classification problem is to decide whether some animal is a
  dolphin. The features have been numbered, respectively 1 is Gills, 2
  is Teeth, 3 is Beak, and 4 is Length. Moreover, the feature values
  have been replaced by numbers. These changes are meant to facilitate
  the logical analysis of the classifier, and do not affect in any
  way the computed explanations.)
  %
  % 2  1: Gills \in { No(=0), Yes(=1) }
  % 4  2: Teeth \in { Few(=0), Many(=1) }
  % 3  3: Beak \in { No(=0), Yes(=1) }
  % 1  4: Length \in { 3(=0), 4(=1), 5(=2) }
  As a result, $\fml{F}_1=\{1,2,3,4\}$,
  $\fml{D}_{1i}=\{0,1\},i=1,2,3$, $\fml{D}_{14}=\{0,1,2\}$, 
  $\fml{K}_1=\{0,1\}$, and $\kappa_1$ is defined
  by~\cref{fig:01a:dl}.
  Clearly,
  $\mbb{F}=\{0,1\}\times\{0,1\}\times\{0,1\}\times\{0,1,2\}$.
  Moreover, the target instance is $(\mbf{v},c)=((0,0,1,2),1)$.
  Each of the rules $\tn{R}_1$, $\tn{R}_2$, and $\tn{R}_3$ tests a
  single literal, and a final default rule $\tn{R}_{\tn{\sc{def}}}$
  fires on the points in feature space inconsistent with the other
  rules.
  Finally,~\cref{tab:tt01a} lists the class predicted by the DL for
  every point in feature space.
\end{example}

\begin{figure}[t]
  \begin{subfigure}[b]{0.6\textwidth}
    \begin{center}
      \vboxedeq{8pt}{eq:fig1}{
        \renewcommand{\arraystretch}{1.2}
        \begin{array}{lllcl}
          \tn{R$_{1}$:} & \tn{IF~}      & (x_1=1) & \tn{~THEN~} & 0 \\
          \tn{R$_{2}$:} & \tn{ELSE IF~} & (x_2=1) & \tn{~THEN~} & 1 \\
          \tn{R$_{3}$:} & \tn{ELSE IF~} & (x_4=1) & \tn{~THEN~} & 0 \\
          \tn{R$_{\tn{\sc{def}}}$:} & \tn{ELSE~} & \quad\quad & \tn{~THEN~} & 1 %\\
        \end{array}
      }
      %  \\ \hline
      %\end{tabular}
    \end{center}
    \caption{Decision list} \label{fig:01a:dl}
  \end{subfigure}
  \begin{subfigure}[b]{0.3975\textwidth}
    \begin{center}
      \vboxedeq{3pt}{eq:fig2}{
        \renewcommand{\arraycolsep}{0.5em}
        \renewcommand{\arraystretch}{1.15}
        \begin{array}{c|c}
        \tn{Item} & \tn{Definition}
        \\ \hline
        \fml{F}_1 & \{1,2,3,4\} \\ \hline
        \fml{D}_{1i},i=1,2,3 & \{0,1\} \\ \hline
        \fml{D}_{14} & \{0,1,2\} \\ \hline
        \fml{K}_1 & \{0,1\} %\\
        %\hline
        \end{array}
      }
    \end{center}
    \caption{Classification problem}
  \end{subfigure}
  %
  %
  %\medskip

  %
  \begin{subfigure}[b]{\textwidth} %0.47125
    \begin{center}
      \vboxedeq{3pt}{eq:fig3}{
        \renewcommand{\arraycolsep}{0.5em}
        \renewcommand{\arraystretch}{1.15}
        \begin{array}{c|c|c}
        \tn{Feature \#} & \tn{Feature Name} & \tn{Original \& Mapped Domain}
        \\ \hline
        1 & \tn{Gills} & \{\tn{No}{\to}0,\tn{Yes}{\to}1\} \\ \hline
        2 & \tn{Teeth} & \{\tn{Few}{\to}0,\tn{Many}{\to}1\} \\ \hline
        3 & \tn{Beak}  & \{\tn{No}{\to}0,\tn{Yes}{\to}1\} \\ \hline
        4 & \tn{Length} & \{\tn{3}{\to}0,\tn{4}{\to}1,\tn{5}{\to}2\} \\
        \hline
        \multicolumn{2}{c|}{\tn{Classes}} &
        \{\tn{No}{\to}0,\tn{Yes}{\to}{1}\}
        %\\ \hline
        \end{array}
      }
      %\renewcommand{\arraystretch}{1.15}
      %\begin{tabular}{|c|c|c|} \hline
      %  Feature \# & Feature Name & Original \& Mapped Domain \\ \hline
      %  1     & Gills      & $\{\tn{No}{\to}0,\tn{Yes}{\to}1\}$ \\ \hline
      %  2     & Teeth      & $\{\tn{Few}{\to}0,\tn{Many}{\to}1\}$ \\ \hline
      %  3     & Beak       & $\{\tn{No}{\to}0,\tn{Yes}{\to}1\}$ \\ \hline
      %  4     & Length     & $\{\tn{3}{\to}0,\tn{4}{\to}1,\tn{5}{\to}2\}$ \\
      %  \hline
      %\end{tabular}
      \caption{Mapping of features} \label{fig:01a:tab}
    \end{center}
  \end{subfigure}
  \caption{Decision list, adapted from~\cite{flach-bk12}}
  \label{fig:runex01a}
\end{figure}

\begin{table}[t]
  \begin{center}
    \renewcommand{\tabcolsep}{0.5em}
    \renewcommand{\arraystretch}{1.05}
    \begin{tabular}{C{1.25cm}|C{0.75cm}C{0.75cm}C{0.75cm}C{0.75cm}C{1.25cm}|C{2.75cm}}
      \toprule
      Entry &
      $x_1$ & $x_2$ & $x_3$ & $x_4$ & Rule & $\kappa_1(x_1,x_2,x_3,x_4)$
      \\ \toprule
      00 &
      0    &  0    &  0    & 0     & $\tn{R}_{\tn{\sc{def}}}$ & 1 \\
      01 &
      0    &  0    &  0    & 1     & $\tn{R}_{3}$ & 0 \\
      02 &
      0    &  0    &  0    & 2     & $\tn{R}_{\tn{\sc{def}}}$ & 1 \\
      \midrule
      03 &
      0    &  0    &  1    & 0     & $\tn{R}_{\tn{\sc{def}}}$ & 1 \\
      04 &
      0    &  0    &  1    & 1     & $\tn{R}_{3}$ & 0 \\
      05 &
      0    &  0    &  1    & 2     & $\tn{R}_{\tn{\sc{def}}}$ & 1 \\
      \midrule
      06 &
      0    &  1    &  0    & 0     & $\tn{R}_{2}$ & 1 \\
      07 &
      0    &  1    &  0    & 1     & $\tn{R}_{2}$ & 1 \\
      08 &
      0    &  1    &  0    & 2     & $\tn{R}_{2}$ & 1 \\
      \midrule
      09 &
      0    &  1    &  1    & 0     & $\tn{R}_{2}$ & 1 \\
      10 &
      0    &  1    &  1    & 1     & $\tn{R}_{2}$ & 1 \\
      11 &
      0    &  1    &  1    & 2     & $\tn{R}_{2}$ & 1 \\
      \midrule
      12 &
      1    &  0    &  0    & 0     & $\tn{R}_{1}$ & 0 \\
      13 &
      1    &  0    &  0    & 1     & $\tn{R}_{1}$ & 0 \\
      14 &
      1    &  0    &  0    & 2     & $\tn{R}_{1}$ & 0 \\
      \midrule
      15 &
      1    &  0    &  1    & 0     & $\tn{R}_{1}$ & 0 \\
      16 &
      1    &  0    &  1    & 1     & $\tn{R}_{1}$ & 0 \\
      17 &
      1    &  0    &  1    & 2     & $\tn{R}_{1}$ & 0 \\
      \midrule
      18 &
      1    &  1    &  0    & 0     & $\tn{R}_{1}$ & 0 \\
      19 &
      1    &  1    &  0    & 1     & $\tn{R}_{1}$ & 0 \\
      20 &
      1    &  1    &  0    & 2     & $\tn{R}_{1}$ & 0 \\
      \midrule
      21 &
      1    &  1    &  1    & 0     & $\tn{R}_{1}$ & 0 \\
      22 &
      1    &  1    &  1    & 1     & $\tn{R}_{1}$ & 0 \\
      23 &
      1    &  1    &  1    & 2     & $\tn{R}_{1}$ & 0 \\
      \bottomrule
    \end{tabular}
  \end{center}
  \caption{Truth table for~\cref{ex:runex01a}, where the target
    instance $(\mbf{v},c)=((0,0,1,2),1)$ corresponds to entry 05.}
  \label{tab:tt01a}
\end{table}

\begin{example}[DT] \label{ex:runex02a}
  The second running example is the decision tree shown
  in~\cref{fig:runex02a}. (This DT is adapted from~\cite{rudin-nips19}
  by replacing the names of the features and renaming the binary
  domains to boolean. The original DT was produced with the tool OSDT
  (optimal sparse decision trees)~\cite{rudin-nips19}.)
  For this DT classifier (see~\cref{fig:02a:def}),
  $\fml{F}_2=\{1,2,3,4,5\}$, $\fml{D}_{2i}=\{0,1\},i=1,\ldots,5$, 
  $\fml{K}_2=\{0,1\}$, and $\kappa_2$ is captured by the DT shown in 
  the~\cref{fig:02a:dt}.
  As can be observed, the DT has 15 nodes, with the non-terminal nodes
  being $N=\{1,2,4,5,7,8,10\}$ and the terminal nodes being
  $T=\{3,6,9,11,12,13,14,15\}$. Each non-terminal node is associated
  with a feature from $\fml{F}$ (we assume univariate DTs), and each
  outgoing edge tests one or more values from the feature's domain.
  %(Following recent work~\cite{iims-jair22}, we use set notation to
  %represent the literal associated with each edge, e.g.\
  For example, the edge $(2,4)$ is associated with the literal
  $x_2=0$, being consistent with points in feature space where $x_2$
  takes value 0.
  Each terminal node is associated with a prediction from $\fml{K}$.
  The set of paths is $\fml{R}$. Throughout the paper, $\fml{R}$ is
  partitioned into two sets, namely $\fml{P}$ associated with
  prediction 1, and $\fml{Q}$ associated with prediction 0. (The split
  of $\fml{R}$ serves to aggregate paths according to their
  prediction, but the naming is arbitrary, and we could consider other
  splits, e.g.\ $\fml{P}$ for prediction 0, and $\fml{Q}$ for
  prediction 1.)
  Moreover, $\fml{P}=\{P_1,P_2,P_3,P_4,P_5\}$, with
  $P_1=\langle1,2,4,7,10,15\rangle$, $P_2=\langle1,2,4,7,11\rangle$,
  $P_3=\langle1,2,5,8,13\rangle$, $P_4=\langle1,2,5,9\rangle$, and
  $P_5=\langle1,3\rangle$.
  Similarly, $\fml{Q}=\{Q_1,Q_2,Q_3\}$, with
  $Q_1=\langle1,2,4,6\rangle$, $Q_2=\langle1,2,4,7,10,14\rangle$, and
  $Q_3=\langle1,2,5,8,12\rangle$.
  Furthermore, the target instance we will study is
  $(\mbf{v},c)=((0,0,1,0,1),1)$, being consistent with path $P_1$ and
  so with prediction 1.
  %
  %Given the instance $(\mbf{v},c)=((0,0,1,0,1),1)$, we set
  %$P_1=\langle1,2,4,7,10,15\rangle$, $P_2=\langle1,2,4,7,11\rangle$,
  %$P_3=\langle1,2,5,8,13\rangle$, 
  %$P_4=\langle1,2,5,9\rangle$, $P_5=\langle1,3\rangle$, and then
  %$Q_1=\langle1,2,4,6\rangle$, $Q_2=\langle1,2,4,7,10,14\rangle$,
  %$Q_3=\langle1,2,5,8,12\rangle$.
  %%
  %Moreover, $P_1$ is the path consistent with the instance.
  %As can be observed,
  %$\mrm{\Lambda}(P_1)=\{(x_1\in\{0\}),(x_2\in\{0\}),(x_3\in\{1\}),(x_4\in\{0\}),(x_5\in\{1\})\}$,
  %and the literals associated with $\mbf{v}=(0,0,1,0,1)$ are
  %$\{x_1=0,x_2=0,x_3=1,x_4=0,x_5=1\}$, hence being pairwise consistent.
  %
  Finally, \cref{tab:tt02a} shows parts of the truth table of the
  example DT, that will be used later when analyzing the instance
  $(\mbf{v},c)=((0,0,1,0,1),1)$.
\end{example}

\begin{figure}[t]
  \begin{subfigure}[b]{0.6\textwidth}
    \scalebox{0.9}{% Example from Rudin et al. NeurIPS'19 paper
%%
%\tikzset{every label/.style={xshift=-0.35ex,
%  yshift=-5.25ex,
%  text width=1ex,
%  align=right, inner sep=1pt, font=\tiny, text=midblue}}
%%
%\tikzset{tlabel/.style={xshift=0.25ex, yshift=1.75ex, text width=1ex,
%    align=right, inner sep=1pt, font=\tiny, text=midblue}}
%%%\tikzset{every node/.style={---rectangle---}}
%
\forestset{
  BDT/.style={
    for tree={
      l=1.5cm,s sep=1.15cm,
      if n children=0{}{circle}, %rectangle
      %if n children=0{}{draw},
      draw=black,%draw=midblue,
      text=black,%text=midblue,
      edge={
        my edge
      },
      if n=1{
        edge+={0 my edge},
      }{},
      edge=thick,
    }
  },
}
%
% middle-middle=x : x_1
% top-left=x : x_2
% bottom-right=x : x_3
% bottom-left=x : x_4
% top-right=x : x_5
%
\begin{forest}
  BDT
  [{$x_1$}, label={[yshift=-6.875ex]{{\tiny1}}} %middle-middle=x
    [{$x_2$}, label={[yshift=-6.875ex]{{\tiny2}}}, %top-left=x
      edge label={node[midway,left,xshift=-0.5pt] {{\scriptsize$=0$}}}
      [{$x_3$}, label={[yshift=-6.875ex]{{\tiny4}}}, %bottom-right=x
        edge label={node[midway,left,xshift=-2.5pt] {{\scriptsize$=0$}}}
        [\rhlight{\textbf{0}}, label={[yshift=-5.0ex]{{\tiny6}}},
          edge label={node[midway,left,xshift=-0.5pt] {{\scriptsize$=0$}}},
          rectangle, fill={tred3!20} ]
        [{$x_4$}, label={[yshift=-6.875ex]{{\tiny7}}}, %bottom-left=x
          edge label={node[midway,right,xshift=0.5pt] {{\scriptsize$=1$}}}
          [{$x_5$}, label={[yshift=-6.875ex]{{\tiny10}}}, %top-right=x
            edge label={node[midway,left,xshift=-1.5pt] {{\scriptsize$=0$}}}
            [\rhlight{\textbf{0}}, label={[yshift=-5.0ex]{{\tiny14}}},
              edge label={node[midway,left,xshift=-1.5pt] {{\scriptsize$=0$}}},
              rectangle, fill={tred3!20} ]
            [\dghlight{\textbf{1}}, label={[yshift=-5.0ex]{{\tiny15}}},
              edge label={node[midway,right,xshift=0.5pt] {{\scriptsize$=1$}}},
              rectangle, fill={tgreen3!25} ]
          ]
          [\dghlight{\textbf{1}}, label={[yshift=-5.0ex]{{\tiny11}}},
            edge label={node[midway,right,xshift=0.5pt] {{\scriptsize$=1$}}},
            rectangle, fill={tgreen3!25} ]
        ]
      ]
      [{$x_4$}, label={[yshift=-6.875ex]{{\tiny5}}}, %bottom-left=x
        edge label={node[midway,right,xshift=1.5pt] {{\scriptsize$=1$}}}
        [{$x_5$}, label={[yshift=-6.875ex]{{\tiny8}}}, %top-right=x
          edge label={node[midway,left,xshift=-0.5pt] {{\scriptsize$=0$}}}
          [\rhlight{\textbf{0}}, label={[yshift=-5.0ex]{{\tiny12}}},
            edge label={node[midway,left,xshift=-0.5pt] {{\scriptsize$=0$}}},
            rectangle, fill={tred3!20} ]
          [\dghlight{\textbf{1}}, label={[yshift=-5.0ex]{{\tiny13}}},
            edge label={node[midway,right,xshift=0.5pt] {{\scriptsize$=1$}}},
            rectangle, fill={tgreen3!25} ]
        ]
        [\dghlight{\textbf{1}}, label={[yshift=-5.0ex]{{\tiny9}}},
          edge label={node[midway,right,xshift=0.5pt] {{\scriptsize$=1$}}},
          rectangle, fill={tgreen3!25} ]
      ]
    ]
    [\dghlight{\textbf{1}}, label={[yshift=-5.0ex]{{\tiny3}}},
      edge label={node[midway,right,xshift=0.5pt] {{\scriptsize$=1$}}},
      rectangle, fill={tgreen3!25} ]
  ]
\end{forest}}
    \caption{Decision tree} \label{fig:02a:dt}
  \end{subfigure}
  \begin{subfigure}[b]{0.4\textwidth}
    \begin{center}
      \scalebox{0.9}{
        \begin{tabular}{lC{1.5cm}} \toprule
          Feature in~\cite{rudin-nips19} & Boolean feature \\ \toprule
          middle-middle=x & $x_1$ \\ \midrule
          top-left=x & $x_2$ \\ \midrule
          bottom-right=x & $x_3$ \\ \midrule
          bottom-left=x & $x_4$ \\ \midrule
          top-right=x & $x_5$ \\ %\midrule
          \bottomrule
        \end{tabular}
      }
    \end{center}

    \bigskip%\medskip%\smallskip

    \begin{center}
      \scalebox{0.9}{
        \begin{tabular}{cc} \toprule
          \multicolumn{2}{c}{Definitions} \\ \toprule
          $\fml{F}_2$ & $\{1,2,3,4,5\}$ \\ \midrule
          $\fml{D}_{21},\ldots,\fml{D}_{25}$ & $\{0,1\}$ \\ \midrule
          $\fml{K}_2$ & $\{0,1\}$ \\
          \bottomrule
        \end{tabular}
      }
    \end{center}
    %\vspace*{2cm}
    \smallskip
    \caption{Mapping of features} \label{fig:02a:def}
  \end{subfigure}
  \caption{Decision tree, adapted from~\cite[Figure~5b]{rudin-nips19}, for
    the \tsf{tic-tac-toe} dataset, and also studied in~\cite{iims-jair22}}
  \label{fig:runex02a}
\end{figure}

\begin{table}[t]
  \begin{subtable}[b]{0.485\textwidth}
    \begin{center}
      \renewcommand{\tabcolsep}{0.5em}
      \renewcommand{\arraystretch}{1.05}
      \begin{tabular}{C{0.45cm}C{0.45cm}C{0.45cm}C{0.45cm}C{0.45cm}C{1.0cm}} \toprule
        $x_3$ & $x_5$ & $x_1$ & $x_2$ & $x_4$ & $\kappa_2(\mbf{x})$ \\ \toprule
        1   &   1   &   0   &  0   &   0   &  1 \\
        1   &   1   &   0   &  0   &   1   &  1 \\
        1   &   1   &   0   &  1   &   0   &  1 \\
        1   &   1   &   0   &  1   &   1   &  1 \\
        1   &   1   &   1   &  0   &   0   &  1 \\
        1   &   1   &   1   &  0   &   1   &  1 \\
        1   &   1   &   1   &  1   &   0   &  1 \\
        1   &   1   &   1   &  1   &   1   &  1 \\ \bottomrule
      \end{tabular}
  \end{center}
    \caption{With $x_3=x_5=1$, it is that case that
      $\kappa_2(x_1,x_2,x_3,x_4,x_5)=1$, independently of the values of
      the other features} \label{tab:tt02aa}
  \end{subtable}
  \quad
  \begin{subtable}[b]{0.485\textwidth}
    \begin{center}
      \renewcommand{\tabcolsep}{0.5em}
      \renewcommand{\arraystretch}{1.05}
      \begin{tabular}{C{0.45cm}C{0.45cm}C{0.45cm}C{0.45cm}C{0.45cm}C{1.0cm}} \toprule
        $x_3$ & $x_5$ & $x_1$ & $x_2$ & $x_4$ & $\kappa_2(\mbf{x})$ \\ \toprule
        0   &   0   &   0   &  0   &   0   &  0 \\
        0   &   1   &   0   &  0   &   0   &  0 \\
        1   &   0   &   0   &  0   &   0   &  0 \\
        1   &   1   &   0   &  0   &   0   &  1 \\
        \bottomrule
      \end{tabular}
  \end{center}
    \caption{With $x_1=x_2=x_4=0$, it is that case that
      $\kappa_2(x_1,x_2,x_3,x_4,x_5)$ can take a value other than 1,
      depending on the values assigned to $x_3$ and $x_5$}
    \label{tab:tt02ab}
  \end{subtable}
  \caption{Partial truth tables of $\kappa_2$. These serve to analyze
    the values taken by $\kappa_2(x_1,x_2,x_3,x_4,x_5)$ for different 
    combinations of feature values, starting from
    $\mbf{x}=(0,0,1,0,1)$.} \label{tab:tt02a}
\end{table}

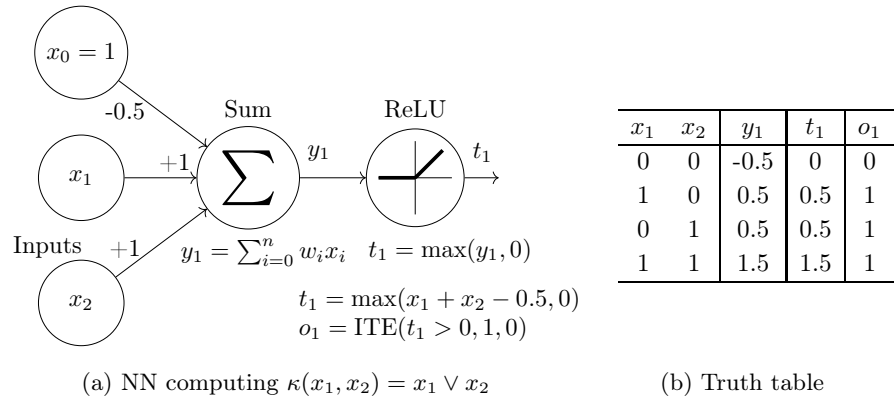
\begin{figure}[t]
  %\begin{adjustbox}{padding=0ex 0ex 0ex 1ex,minipage=\linewidth,frame}
    \begin{subfigure}[b]{0.65\textwidth}
      \begin{tabular}{r@{}}
        \hspace*{-0.575cm}\scalebox{0.925}{\tikzset{basic/.style={draw,fill=none,
  text badly centered,minimum width=3em}}
\tikzset{input/.style={basic,circle,minimum width=3.5em}}
\tikzset{weights/.style={basic,rectangle,minimum width=2em}}
\tikzset{functions/.style={basic,circle, minimum width=4em}}
\newcommand{\addaxes}{\draw (0em,1em) -- (0em,-1em)
  (-1em,0em) -- (1em,0em);}
\newcommand{\relu}{\draw[line width=1.5pt] (-1em,0) -- (0,0)
  (0,0) -- (0.75em,0.75em);}
\newcommand{\stepfunc}{\draw[line width=1.5pt] (0.65em,0.65em) -- (0,0.65em) 
  -- (0,-0.65em) --
  (-0.65em,-0.65em);}

\begin{tikzpicture}[scale=1.2]
  % Draw input nodes
  \node[input] (f0) at (0,1.5cm) {$x_0=1$};
  \node[input] (f1) at (0,0cm) {$x_1$};
  \node[input] (f2) at (0,-1.5 cm) {$x_2$};
  % Draw summation node
  \node[functions] (sum) at (2,0) {\scalebox{0.9}{\Huge$\sum$}};
  \node[above=0.75cm] at (sum) {Sum};
  \node[below=0.75cm,xshift=0.225cm] at (sum) {$y_1=\sum_{i=0}^n w_ix_i$};
  % Draw node for activation function
  \node[functions] (activation) at (4,0) {};
  % Place activation function in its node
  \begin{scope}[xshift=4cm,scale=1.25]
    \addaxes
    % flexible selection of activation function
    \relu
    % \stepfunc
  \end{scope}
  % Connect sum to relu
  \draw[->] (sum) -- (activation);
  \draw[->] (activation) -- ++(1,0);
  % Edges
  \draw[->] (f0) -- (sum);
  \draw[->] (f1) -- (sum);
  \draw[->] (f2) -- (sum);
  \node at (0.525cm,0.8cm) {-0.5};
  \node at (1.12cm,0.175cm) {+1};
  \node at (0.525cm,-0.8575cm) {+1};
  % Labels
  %\node[above=0.625cm] at (f0) {Inputs};
  \node[below left=0.275cm and -0.575cm of f1] {Inputs};
  \node[above=0.75cm] at (activation) {ReLU};
  \node[below=0.75cm,xshift=0.5cm] at (activation) {$t_1=\max(y_1,0)$};
  \node[above right=-0.4cm and 0.2cm of sum] {$y_1$};
  \node[above right=-0.4cm and 0.2cm of activation] {$t_1$};
  %
  % Expressions
  \node[below right=0.925cm and -2.35cm of activation] {$\begin{array}{l}t_1=\max(x_1+x_2-0.5,0)\\o_1=\tn{ITE}(t_1>0,1,0)\\\end{array}$};
\end{tikzpicture}}
      \end{tabular}
      \caption{NN computing $\kappa(x_1,x_2)=x_1\lor{x_2}$}
      \label{fig:exnn:nn}
    \end{subfigure}
    \begin{subfigure}[b]{0.325\textwidth}
      \centering
      \begin{minipage}{\textwidth}
        \scalebox{0.95}{
          \renewcommand{\tabcolsep}{0.5em}
          \renewcommand{\arraystretch}{1.125}
          \begin{tabular}{cc|c|c|c} \hline
            $x_1$ & $x_2$ & $y_1$ & $t_1$ & $o_1$ \\ \hline
            0     &  0    & -0.5 &  0    & 0   \\
            1     &  0    & 0.5  &  0.5  & 1   \\
            0     &  1    & 0.5  &  0.5  & 1   \\
            1     &  1    & 1.5  &  1.5  & 1   \\ \hline
          \end{tabular}
        }
        \vspace*{1.025cm}
      \end{minipage}
      \caption{Truth table}
      \label{fig:exnn:tt}
    \end{subfigure}
    \caption{Simple neural network} %%\label{fig:exnn}
    \label{fig:runex03a}
  %\end{adjustbox}
  %
\end{figure}

\begin{example}[NN] \label{ex:runex03a}
  The third running example is an NN, as shown
  in~\cref{fig:runex03a}.
  For this example, $\fml{F}_3=\{1,2\}$,
  $\mbb{D}=\{\fml{D}_{31},\fml{D}_{32}\}$, with
  $\fml{D}_{31}=\fml{D}_{32}=\{0,1\}$,
  and so $\mbb{F}_3=\{0,1\}^2$, $\fml{K}=\{0,1\}$, and
  $\kappa_3(x_1,x_2)=(\max(x_1+x_2-0.5,0)>0)$.
  We also have, from~\eqref{eq:nndef}:
  \[
  \begin{array}{l}
    \mbf{A}^1 = \left[ \begin{matrix}
        -0.5 & {+}1 & {+}1 \end{matrix} \right] \\[4pt]
    \mbf{x}^0 = \left[ \begin{matrix} 1 & x^0_1 & x^0_2 \end{matrix}
      \right] \\[4.5pt]
    \mbf{y}^{1} = \left[\begin{matrix}
        y^1_1\end{matrix}\right] = \mbf{A}^{1} \cdot
    (\mbf{x}^{0})^{\tn{T}}\\
  \end{array}
  \]
  
  It is easy to conclude that the classifier corresponds to the Boolean
  function $f(x_1,x_2)=x_1\lor{x_2}$.
  This is confirmed by the truth table shown in~\cref{fig:exnn:tt}.
  %confirms that the classifier correctly predicts the OR between $x_1$
  %and $x_2$.
  %
  For this example, the instance considered is $(\mbf{v},c)=((1,0),1)$.
\end{example}

\begin{example}[Monotonic classifier]
  The fourth and final running example is a monotonic classifier,
  adapted from~\cite{msgcin-icml21}.
  The goal is to predict a student's grade given the grades on the
  different components of assessment.
  The different grading components have domain $\{0,\ldots,10\}$.
  It is also the case that
  ${F}\preccurlyeq{E}\preccurlyeq{D}\preccurlyeq{C}\preccurlyeq{B}\preccurlyeq{A}$,
  where the operator $\preccurlyeq$ is used to represent the order
  between different grades.
  The details of the classifier are summarized in~\cref{fig:runex04a}.
  The classifier ``predicts'' a student's grade given the grades in
  different grading components, using the following formulas:
  \[
  \begin{array}{rcl}
    M & = & \ite(S\ge9,A,\ite(S\ge7,B,\ite(S\ge5,C,%\\[1.5pt]
    %& &
    \ite(S\ge4,D,\ite(S\ge2,E,F)))))\\[3pt]
    S & = & \max\left[0.3
      \times{Q}+0.6\times{X}+0.1\times{H},R\right]%\\%[3pt]
  \end{array}
  \]
  Also, it is clearly the case that,
  $\kappa_4(\mbf{x_1})\preccurlyeq\kappa_4(\mbf{x}_2)\tn{~if~$\mbf{x}_1\le\mbf{x}_2$}$,
  and so the classifier is monotonic.
\end{example}

%% Monotonic classifier
%\ifthenelse{\boolean{fullversion}}{
%  \jnote{Add example of monotonic classifier.}
%}{}

%
\begin{figure}[t]
  \begin{subfigure}[b]{1.0\textwidth}
    \begin{center}
      \renewcommand{\tabcolsep}{0.5em}
      \renewcommand{\arraystretch}{1.05}
      \begin{tabular}{C{1.75cm}C{3.0cm}C{2.25cm}C{1.75cm}} \toprule
        Feature~id & Feature~variable & Feature~name & Domain \\
        \midrule
        1        & $Q$        & Quiz       & $\{0,\ldots,10\}$ \\
        2        & $X$        & Exam       & $\{0,\ldots,10\}$ \\
        3        & $H$        & Homework   & $\{0,\ldots,10\}$ \\
        4        & $R$        & Project    & $\{0,\ldots,10\}$ \\
        \bottomrule
      \end{tabular}
    \end{center}
    \caption{Features and domains}
  \end{subfigure}

  \medskip

  \begin{subfigure}[b]{1.0\textwidth}
    \begin{center}
      \renewcommand{\tabcolsep}{0.5em}
      \renewcommand{\arraystretch}{1.05}
      \begin{tabular}{cccc} \toprule
        Variable & \multicolumn{2}{c}{Meaning} & Range \\ \toprule
        $S$ & \multicolumn{2}{c}{Final score} & $\in\{0,\ldots,10\}$
        \\ \midrule
        $\kappa(\cdot)\triangleq{M}$ & \multicolumn{2}{c}{Student grade} &
        $\in\{A,B,C,D,E,F\}$
        \\ \bottomrule
      \end{tabular}
    \end{center}
    \caption{Definition of $\kappa_4$}
  \end{subfigure}
  \caption{Example of a monotonic classifier~\cite{msgcin-icml21}}
  \label{fig:runex04a}
\end{figure}
%

%\paragraph{Naive bayes classifiers (NBCs).}
%%
%~\\
%
%\paragraph{Tree ensembles: random forests \& boosted trees.}
%%
%~\\
%
%\paragraph{Neural networks.}
%%
%~\\

\subsection{Non-Formal Explanations} \label{ssec:nfxp}

As observed in~\cref{sec:intro}, most of past work on XAI has studied
non-formal explainability approaches. We will briefly summarize the
main ideas. The interested reader is referred to the many surveys on
the
topic~\cite{berrada-ieee-access18,muller-dsp18,muller-bk19,pedreschi-acmcs19,muller-ieee-proc21,guan-ieee-tnnls21,doran-jair22,holzinger-bk22,holzinger-xxai22b}.
\jnoteF{%
  Topics to cover:
  \begin{enumerate}
  \item LIME, SHAP
  \item Anchor
  \item Saliency-based methods
  \item Intrinsic interpretability~\cite{rudin-naturemi19,molnar-bk20}
  \end{enumerate}
}
There is a burgeoning and fast growing body of work on non-formal
approaches for computing explanations. The best known approaches offer
no guarantees of rigor, and include model-agnostic approaches or
solutions based on saliency maps for neural networks.

%LIME~\cite{guestrin-kdd16}, SHAP~\cite{lundberg-nips17} and
%Anchor~\cite{guestrin-aaai18} exemplify model agnostic
%explainers\footnote{%
%The impact of these tools can only be viewed as impressive
%(e.g.\ \url{https://bit.ly/3eXIiNU}, \url{https://bit.ly/3BJL4z7}, and 
%\url{https://bit.ly/3djA1Do}), especially given the documented evidence
%about the unsoundness of their
%results~\cite{inms-corr19,nsmims-sat19,ignatiev-ijcai20}.}

\paragraph{Model-agnostic methods.}
%~\\
The most visible approaches for explaining ML models are
model-agnostic
methods~\cite{guestrin-kdd16,lundberg-nips17,guestrin-aaai18}.
These can be organized into methods that learn a simpler interpretable
model, e.g.\ a linear model or a decision tree. This is the case with
LIME~\cite{guestrin-kdd16} and SHAP~\cite{lundberg-nips17}. The
difference between LIME and SHAP is the approach used to learn the
model, with LIME being based on iterative sampling, and SHAP based on
the \emph{approximate} computation of Shapley values. (It should be 
underscored that the Shapley values in SHAP are not computed exactly,
but only approximated. Indeed, the complexity of exactly computing
Shapley values is unwieldy~\cite{vandenbroeck-aaai21,barcelo-aaai21},
with one exception being a fairly restricted form of boolean
circuits~\cite{barcelo-aaai21}, referred to as deterministic,
decomposable boolean circuits~\cite{darwiche-jair02}, and which
capture binary decision trees\footnote{%
Earlier work~\cite{lundberg-naturemi20} also investigated the exact
computation of Shapley values for decision trees. However, issues
about the proposed algorithm have been raised by more recent
work~\cite{barcelo-aaai21,vandenbroeck-aaai21}.}.
These model agnostic methods can also be viewed as associating a
measure of relative importance to each features, being often referred
to as \emph{feature attribution} methods.
One alternative model-agnostic approach is to identify which features
are the most relevant for the prediction. We refer to such approaches
as \emph{feature selection} methods. One concrete example is
Anchor~\cite{guestrin-aaai18}. Similar to other model agnostic
approaches, Anchor is based on sampling.
It should be noted that model-agnostic approaches exhibit a number of
important drawbacks, the most critical of which is
unsoundness~\cite{inms-corr19,nsmims-sat19,ignatiev-ijcai20}.
Despite critical limitations, including the risk of unsound
explanations, the impact of these tools can only be viewed as
impressive\footnote{See for example
\url{https://bit.ly/3eXIiNU}, \url{https://bit.ly/3BJL4z7}, and
\url{https://bit.ly/3djA1Do}.}.

\paragraph{Neural networks \& saliency maps.}
%~\\
In the concrete case of neural networks, past work proposed the use
of variants of saliency
maps~\cite{vedaldi-iclr14,muller-bk19,muller-dsp18,muller-ieee-proc21},
that give a graphical interpretation of a prediction. One popular
approach is based on so-called layerwise relevancy
propagation~\cite{muller-plosone15}.
However, recent work has revealed important drawbacks of these  
approaches~\cite{adebayo-nips18,adebayo-xai19,eisemann-corr19,landgraf-icml20}.

\paragraph{Intrinsic interpretability.}
%~\\
Some authors have advocated the use of so-called \emph{interpretable}
ML models, for which the explanation is the model
itself~\cite{rudin-naturemi19,molnar-bk20,rudin-ss22,rudin-nips22,rudin-facct22,rudin-dss22,rudin-cikmw22,gupta-tplp22}.
For example, it is widely accepted that decision trees are
interpretable. Claims about the interpretability of decision trees go
back at least until the early
2000s~\cite[Sec.~9,~Page~206]{breiman-ss01}. Motivated by their
interpretability, decision trees have been applied to a wide range of
domains, including the medical
domain~\cite{tanner-plosntd08,valdes-naturesr16,belmonte-ieee-access20}.
Unfortunately, recent results~\cite{iims-jair22} demonstrate that
decision trees can hardly be deemed interpretable, at least as long as
interpretability correlates with the succinctness of explanations.
Interpretability of decision lists and sets is at least as problematic
as it is for decision trees. If that were not the case, then one would
be able to just represent DTs as DLs or DSs~\cite{rivest-ml87}. 
The bottom line is that even interpretable ML models should be
explained, as the comprehensive results in earlier
work~\cite{iims-jair22} demonstrate.

%\subsubsection{Known Limitations}
%
%\subsubsection{Addressing Limitations}

\paragraph{Assessment.}
A number of additional limitations of non-formal explanations have
been reported in recent 
years~\cite{lukasiewicz-corr19,lakkaraju-aies20a,lakkaraju-aies20b,weller-ecai20,iincms-corr21,ghassemi-ldh21,lakkaraju-corr22a}. Furthermore,
a number of authors have raised concerns about the current uses of
XAI and its
technology~\cite{nagendran-bmj20,newman-brookings21,kahn-fortune22},
with examples of misuse also reported~\cite{israni-jorlt17}.

\section{Formal Explainability}  \label{sec:dxps}

Formal explanation
%\footnote{%
% There is an extensive body of work on non-formal XAI approaches to
%  XAI
%~\cite{berrada-ieee-access18,muller-dsp18,muller-bk19,pedreschi-acmcs19,muller-ieee-proc21,guan-ieee-tnnls21,holzinger-bk22,holzinger-xxai22b,doran-jair22}.}
approaches have been studied in a growing body of research in recent  
years~\cite{darwiche-ijcai18,inms-aaai19,darwiche-aaai19,inms-nips19,inms-corr19,nsmims-sat19,hazan-aies19,darwiche-pods20,ignatiev-ijcai20,darwiche-ecai20,icshms-cp20,inams-aiia20,marquis-kr20,darwiche-kr20,toni-kr20,mazure-sum20,iims-corr20,msgcin-nips20,barcelo-nips20,iims-corr20,msgcin-icml21,ims-ijcai21,kwiatkowska-ijcai21,ims-sat21,asher-cdmake21,cms-cp21,mazure-cikm21,hiims-kr21,marquis-kr21,barcelo-nips21,tan-nips21,lorini-clar21,amgoud-ecsqaru21,hiicams-corr21,toni-aij21,kutyniok-jair21,darwiche-jair21,iincms-corr21,msi-aaai22,hiicams-aaai22,iisms-aaai22,rubin-aaai22,darwiche-aaai22,marquis-aaai22,tan-stoc22,tan-icml22a,amgoud-ijcai22,marquis-ijcai22a,leite-kr22,lorini-wollic22,labreuche-sum22,barcelo-nips22,iims-jair22,marquis-dke22,darwiche-jlli22-web,waldchen-phd22,iincms-corr22,yisnms-corr22,hms-corr22,ims-corr22,barcelo-corr22,ihincms-corr22,cms-aij23,hims-aaai23,yisnms-aaai23,hcmpms-tacas23,katz-tacas23,amgoud-ijar23}. %
%~\cite{darwiche-ijcai18,inms-aaai19,darwiche-aaai19,inms-nips19,inms-corr19,nsmims-sat19,hazan-aies19,darwiche-pods20,darwiche-ecai20,marquis-kr20,darwiche-kr20,toni-kr20,mazure-sum20,inams-aiia20,msgcin-nips20,iims-corr20,msgcin-icml21,ims-ijcai21,kwiatkowska-ijcai21,hiims-kr21,marquis-kr21,ims-sat21,asher-cdmake21,cms-cp21,mazure-cikm21,hiicams-corr21,toni-aij21,lorini-clar21,kutyniok-jair21,darwiche-jair21,amgoud-ecsqaru21,tan-nips21,barcelo-nips21,iincms-corr21,hiicams-aaai22,iisms-aaai22,msi-aaai22,rubin-aaai22,marquis-aaai22,marquis-ijcai22a,amgoud-ijcai22,tan-stoc22,tan-icml22a,leite-kr22,lorini-wollic22,waldchen-phd22,iincms-corr22,hms-corr22,ims-corr22,yisnms-corr22,iims-jair22,darwiche-jlli22-web}. %
%
This section introduces formal explanations and describes some of
their properties.

\paragraph{Explanation problems.}
Given a classifier $\fml{M}=(\fml{F},\mbb{D},\mbb{F},\fml{K},\kappa)$,
we consider two explanation problems.
First, we mostly study a given \emph{local} explanation problem
$\fml{E}_L=(\fml{M},\mbf{v},c)$, with $\mbf{v}\in\mbb{F}$,
$c\in\fml{K}$ and $c=\kappa(\mbf{v})$, which respects a concrete point
in feature space, i.e.\ a concrete prediction.
Second, we will also consider a \emph{global} explanation problem
$\fml{E}_G=(\fml{M},c)$, with $c\in\fml{K}$, which respects solely a
concrete prediction $c$, that can be predicted in many points of
feature space.

%%Given the above, the universe of \emph{explanation problems} is
%%defined by
%%$\mbb{E}_I=\{\fml{E}\,|\,\fml{E}=(\fml{M}, (\mbf{v},c)),
%%\fml{M}\in\mbb{M}, \mbf{v}\in\mbb{F}, c\in\fml{K}, c=\kappa(\mbf{v})\}$.
%
As a result, a tuple $(\fml{M},(\mbf{v},c))$ will allow us to
unambiguously represent the classification problem $\fml{M}$ for which
we want to compute the local AXp's and CXp's given the instance
$(\mbf{v},c)$.
Similarly, $(\fml{M},c)$ also unambiguously represents the
classification problem for which we want to compute the global AXp's
given the prediction $c$.

\subsection{Abductive Explanations}

\jnoteF{Std definition}

This paper uses the definition of \emph{abductive
  explanation}~\cite{inms-aaai19} (AXp), which corresponds to a
PI-explanation~\cite{darwiche-ijcai18} in the case of boolean  
classifiers. AXp's represent prime implicants of the discrete-valued
classifier function (which computes the predicted class)\footnote{%
  There exist also standard references with detailed overviews of the
  uses of prime implicants in the context of boolean
  functions~\cite{somenzi-bk06,crama-bk11}.
  Generalizations of prime implicants beyond boolean domains have been 
  considered before~\cite{marquis-fair91}.
  Prime implicants have also been referred to as minimum satisfying
  assignments in first-order logic~(FOL)~\cite{mcmillan-cav12}, and
  have been studied in modal and description
  logics~\cite{bienvenu-jair09}.
}. AXp's can also be viewed as an instantiation of logic-based
abduction~\cite{gottlob-ese90,selman-aaai90,bylander-aij91,gottlob-jacm95}.
Throughout this paper we will opt to use the acronym AXp to refer to
abductive explanations.

Let us consider a given classifier, computing a classification function
$\kappa$ on feature space $\mbb{F}$, a point $\mbf{v}\in\mbb{F}$, with
prediction $c=\kappa(\mbf{v})$, and let $\fml{X}$ denote a subset of
the set of features $\fml{F}$, 
$\fml{X}\subseteq\fml{F}$. $\fml{X}$ is a weak AXp for the instance
$(\mbf{v},c)$ if,
\begin{equation} \label{eq:axp1}
  \begin{array}{rcl}
    \waxp(\fml{X}) & ~:=~~ &
    \forall(\mbf{x}\in\mbb{F}).%
    \left[\bigwedge_{i\in\fml{X}}(x_i=v_i)\right]\limply(\kappa(\mbf{x})=c)\\ %\nolimits
  \end{array}
\end{equation}
(We could highlight that $\waxp$ is parameterized on $\kappa$,
$\mbf{v}$ and $c$, but opt not to clutter the notation, and so these
dependencies will be left implicit.)
Thus, given an instance $(\mbf{v},c)$, a (weak) AXp is a subset of
features which, if fixed to the values dictated by $\mbf{v}$, then the
prediction is guaranteed to be $c$, independently of the values
assigned to the other features.

\begin{example} \label{ex:waxp01a}
  With respect to the DL of~\cref{fig:runex01a}, it is apparent that
  $\fml{X}=\{1,4\}$ is a (weak) abductive explanation for the instance 
  $(\mbf{v},c)=((0,0,1,2),1)$.
  Indeed, if $(x_1=0)\land(x_4=2)$, then the prediction
  will be 1, independently of the values taken by the other features.
  This can easily be concluded from~\cref{tab:tt01a}; if $x_1$ and $x_4$ 
  are fixed, then the possible entries are 02, 05, 08 and 11, all with
  prediction 1.
  Hence, we can write that,
  \[
  \forall(\mbf{x}\in\mbb{F}_1).
  \left[(x_1=0)\land(x_4=2)\right]\limply(\kappa_1(\mbf{x})=1)
  \]
  Observe that any set $\fml{Z}$, with
  $\fml{X}\subseteq\fml{Z}\subseteq\fml{F}$, is also a weak AXp.
\end{example}

Moreover, $\fml{X}\subseteq\fml{F}$ is an AXp if, besides being a weak
AXp, it is also subset-minimal, i.e.
\begin{equation} \label{eq:axp2a}
  \begin{array}{rcl}
    \axp(\fml{X}) & ~:=~~ &
    \waxp(\fml{X})\land\forall(\fml{X}'\subsetneq\fml{X}).\neg\waxp(\fml{X}')\\
  \end{array}
\end{equation}  

\begin{example} \label{ex:axp01a}
  From~\cref{tab:tt01a}, and given the weak AXp $\{1,4\}$
  (see~\cref{ex:waxp01a}) it is also possible to conclude that if
  either $x_1$ or $x_4$ are allowed to change their value, then the
  prediction can be changed. Hence, $\fml{X}=\{1,4\}$ is effectively
  an AXp.
\end{example}

Observe that an AXp can be viewed as a possible irreducible answer to
a ``\tbf{Why?}'' question, i.e.\ why is the classifier's prediction
$c$?
It should be plain in this work, but also in earlier work, that the
representation of AXp's using subsets of features aims at simplicity.
The sufficient condition for the prediction is evidently the
conjunction of literals associated with the features contained in the
AXp.

The following example demonstrates the importance of explaining
decision trees, even if these are most often deemed interpretable.
\begin{example} \label{ex:xp02a}
  We consider the DT from~\cref{fig:runex02a}, and the instance
  $((0,0,1,0,1),1)$.
  Intrinsic interpretability~\cite{rudin-naturemi19,molnar-bk20} would
  argue that the explanation for this instance is the path consistent
  with the instance. Hence, we would claim that,
  \[
  \begin{array}{cccc}
    \tn{IF~} &
       [(x_1=0)\land(x_2=0)\land(x_3=1)\land(x_4=0)\land(x_5=1)] &
       \tn{~THEN~} & 1 \\
  \end{array}
  \]
  However,~\cref{tab:tt02a} clarifies that, as long as features 3 and
  5 are assigned the same value, then the prediction remains
  unchanged. Hence, a far more \emph{intuitive} explanation would
  be,
  \[
  \begin{array}{cccc}
    \tn{IF~} &
       [(x_3=1)\land(x_5=1)] & \tn{~THEN~} & 1 \\
  \end{array}
  \]
  Clearly, the (only) AXp for the given instance is exactly that one,
  i.e.\ $\fml{X}=\{3,5\}$, and we can state,
  \[
  \forall(\mbf{x}\in\mbb{F}_2).
  \left[(x_3=1)\land(x_5=1)\right]\limply(\kappa_2(\mbf{x})=1)
  \]
\end{example}
\cref{ex:xp02a} illustrates important limitations of DTs in terms of
interpretability, and justifies recent work on explaining
DTs~\cite{iims-corr20,iims-corr22,iims-jair22}. More importantly, it
has been shown that the redundancy in tree paths (i.e.\ features
unnecessary for the prediction) can be arbitrarily large on the number
of features~\cite{iims-jair22}.
Given the recent efforts on learning optimal (and quasi-optimal)
``interpretable'' models~\cite{nijssen-kdd07,hebrard-cp09,nijssen-dmkd10,rudin-aistats15,leskovec-kdd16,bertsimas-ml17,rudin-jmlr17a,rudin-jmlr17b,rudin-kdd17,verwer-cpaior17,rudin-mpc18,rudin-aistats18,nipms-ijcai18,ipnms-ijcar18,meel-cp18,meel-aies19,verwer-aaai19,rudin-nips19,avellaneda-corr19,avellaneda-aaai20,schaus-cj20,schaus-aaai20,rudin-icml20,janota-sat20,hebrard-ijcai20,schaus-ijcai20a,schaus-ijcai20b,meel-ecai20,ignatiev-cp20,ignatiev-jair21,ilsms-aaai21,demirovic-aaai21,szeider-aaai21a,szeider-aaai21b,verwer-icml21,mcilraith-cp21,ansotegui-corr21,demirovic-jmlr22,meel-jair22,verwer-aaai22,rudin-aaai22},
that include learning optimal decision trees and sets, recent results
demonstrate~\cite{ims-sat21,iims-jair22} that even such optimal and
interpretable models should be explained.

%% SEE BELOW
%
%Moreover, and by observing that $\waxp$ is monotone, then the
%definition of predicate $\axp$ can be changed as follows:
%%
%\begin{equation} \label{eq:axp2b}
%  \begin{array}{rcl}
%    \axp(\fml{X}) & ~:=~~ &
%    \waxp(\fml{X})\land\forall(t\in\fml{X}).\neg\waxp(\fml{X}\setminus\{t\})\\
%  \end{array}
%\end{equation}  
%%
%Clearly,~\eqref{eq:axp2b} simplifies the implementation of algorithms
%for the computation of AXp's.
%\jnoteF{Monotonicity to simplify the problem!!!}

\subsection{Contrastive Explanations}

\jnoteF{Std definition}

Similarly to the case of AXp's, one can define (weak) contrastive
explanations (CXp's)~\cite{miller-aij19,inams-aiia20}.
$\fml{Y}\subseteq\fml{F}$ is a weak CXp for the instance $(\mbf{v},c)$
if,
\begin{equation} \label{eq:cxp1}
  \begin{array}{rcl}
    \wcxp(\fml{Y}) & ~:=~~ & \exists(\mbf{x}\in\mbb{F}).%
    \left[\bigwedge_{i\not\in\fml{Y}}(x_i=v_i)\right]\land(\kappa(\mbf{x})\not=c)\\ %\nolimits
  \end{array}
\end{equation}
(As before, for simplicity we keep the parameterization of $\wcxp$ on
$\kappa$, $\mbf{v}$ and $c$ implicit.)
Thus, given an instance $(\mbf{v},c)$, a (weak) CXp is a subset of
features which, if allowed to take any value from their domain, then
there is an assignment to the features that changes the prediction to
a class other than $c$, this while the features not in the explanation
are kept to their values (\emph{ceteris paribus}).

\begin{example}
  For the DT of~\cref{fig:runex02a}, and the instance
  $(\mbf{v},c)=((0,0,1,0,1),1)$, it is the case that $\fml{Y}=\{3\}$
  is a (weak) contrastive explanation. Indeed, if we allow the value
  of feature 3 to change, then there exists some point in feature space,
  e.g.\ $(0,0,0,0,1)$, for which the remaining features take the
  values in $\mbf{v}$, and such that the prediction changes to 0.
  Hence, we can write,
  \[
  \exists(\mbf{x}\in\mbb{F}_2).%
  \left[(x_1=0)\land(x_2=0)\land(x_4=0)\land(x_5=1)\right]\land(\kappa_2(\mbf{x})\not=1)
  \]
  Intuitively, we are saying that it suffices to change the value of
  feature 3 to get a different prediction.
\end{example}

Furthermore, a set $\fml{Y}\subseteq\fml{F}$ is a CXp if, besides
being a weak CXp, it is also subset-minimal, i.e.
\begin{equation} \label{eq:cxp2a}
  \begin{array}{rcl}
    \cxp(\fml{Y}) & ~:=~~ &
    \wcxp(\fml{Y})\land\forall(\fml{Y}'\subsetneq\fml{Y}).\neg\wcxp(\fml{Y}')\\
  \end{array}
\end{equation}  

\begin{example} \label{ex:cxp01a}
  For the DL of~\cref{fig:runex01a}, it is plain that if $x_1$ is
  allowed to change value (i.e.\ entry 14 of~\cref{tab:tt01a}) or if
  $x_4$ is allowed to change value (i.e.\ entry 01
  of~\cref{tab:tt01a}), then the prediction will change. Hence, either
  $\{1\}$ or $\{4\}$ are weak contrastive explanations for the given
  instance.
  Furthermore, both weak CXp's are irreducible, and so both are
  effectively CXp's.
\end{example}

A CXp can be viewed as a possible irreducible answer to a ``\tbf{Why
  Not?}'' question, i.e.\ why isn't the classifier's prediction a
class other than $c$?
A different perspective for a contrastive explanation is as the answer
to a \emph{How?} question, i.e.\ how to change the features so as to
change the prediction. In recent literature this alternative view has
been investigated under the name \emph{actionable
  recourse}~\cite{liu-fat19,alfano-fat20,valera-facct21,valera-corr20}.
It should be underlined that whereas AXp's correspond to prime
implicants of the boolean function $(\kappa(\mbf{x})=c)$ that are
consistent with some point $\mbf{v}\in\mbb{F}$, CXp's are \emph{not}
prime implicates of function $(\kappa(\mbf{x})=c)$. Nevertheless, the
concept of \emph{counterexample} studied in formal
explainability~\cite{inms-aaai19} corresponds to prime implicates of
the function $(\kappa(\mbf{x})=c)$ (which are not restricted to be
consistent with some specific point $\mbf{v}\in\mbb{F}$).
\label{def:cex}

%%\ifthenelse{\boolean{shownotes}}{
%%  \jnote{Define CEx's}
%%}

One important observation is that, independently of what $\kappa$
represents, the $\waxp$ and $\wcxp$ predicates (respectively defined
using~\eqref{eq:axp1} and~\eqref{eq:cxp1}) are \emph{monotone}\footnote{%
  Clearly, from the definition of $\waxp$ (resp.~$\wcxp$), if
  $\waxp(\fml{Z})$ (resp.\ $\wcxp(\fml{Z})$) holds, then 
  $\waxp(\fml{Z}')$ (resp.~$\wcxp(\fml{Z}')$) also holds for any
  superset $\fml{Z}'$ of $\fml{Z}$. If $\waxp(\fml{Z})$
  (resp.~$\wcxp(\fml{Z})$) does not hold,
  then $\waxp(\fml{Z}')$ (resp.~$\wcxp(\fml{Z}')$) also does not hold
  for any subset $\fml{Z}'$ of $\fml{Z}$.}.
This means that the tests for minimality (i.e.,
respectively~\eqref{eq:axp2a} and~\eqref{eq:cxp2a}) can be simplified
to:
\begin{equation} \label{eq:axp2b}
  \begin{array}{rcl}
    \axp(\fml{X}) & ~:=~~ &
    \waxp(\fml{X})\land\forall(t\in\fml{X}).\neg\waxp(\fml{X}\setminus\{t\})\\
  \end{array}
\end{equation}  
and,
\begin{equation} \label{eq:cxp2b}
  \begin{array}{rcl}
    \cxp(\fml{Y}) & ~:=~~ &
    \wcxp(\fml{Y})\land\forall(t\in\fml{Y}).\neg\wcxp(\fml{Y}\setminus\{t\})\\
  \end{array}
\end{equation}  
Observe that, instead of considering all possible subsets of
$\fml{X}$ (resp.~$\fml{Y}$), it suffices to consider the subsets
obtained by removing a single element from $\fml{X}$
(resp.~$\fml{Y}$).
This observation is %also
at the core of the algorithms proposed in recent years for computing
AXp's and CXp's of a growing range of families of
classifiers~\cite{inms-aaai19,inms-nips19,nsmims-sat19,msgcin-nips20,iims-corr20,msgcin-icml21,ims-ijcai21,kwiatkowska-ijcai21,hiims-kr21,ims-sat21,hiicams-corr21}.
%%
%%\jnoteF{Monotonicity to simplify the problem.}
%%
As will be clarified in~\cref{sec:cxps}, the computation of AXp's can
be related with MUS extraction, and the computation of CXp's can be
related with MCS extraction.

%\begin{example}
%  For the example DT of~\cref{fig:runex02a}, there are three paths ...
%\end{{example}

Given a local explanation problem $\fml{E}$, the sets of AXp's and
CXp's are defined as follows,
\begin{align}
  \mbb{A}(\fml{E}) & = \{ \fml{X}\subseteq\fml{F} \,|\,\axp(X)\}
  \label{eq:setaxp}
  \\[2.5pt]
  \mbb{C}(\fml{E}) & = \{ \fml{Y}\subseteq\fml{F} \,|\,\cxp(X)\} %%\\
  \label{eq:setcxp}
\end{align}
%
%%\jnoteF{Define $\mbb{A}(\fml{E})$ and $\mbb{C}(\fml{E})$.}
%

\subsection{Global Abductive Explanations \& Counterexamples}
\label{ssec:gaxp}

The definition of AXp's considered until now is \emph{localized}, in
that it takes a concrete point $\mbf{v}$ into account. However,
abductive explanations can be defined only with respect to the class,
and ignore concrete points in feature space; these are referred to as
global AXp's.
Following~\cite{inms-nips19}, let $\pi:\mbb{F}\to\{0,1\}$, represent a
term that is a prime implicant of the predicate $[\kappa(\mbf{x})=c]$,
i.e.,
\[
\forall(\mbf{x}\in\mbb{F}).\pi(\mbf{x})\limply\left(\kappa(\mbf{x})=c\right)
\]
(Each literal will be of the form $x_i=u_i$, where $u_i$ is taken from
$\fml{D}_i$.)
The set of literals of $\pi$ is a global abductive explanation of the
prediction $c$.

We are also interested in the prime implicates of the predicate
$[\kappa(\mbf{x})=c]$, which will be convenient to represent by
$\neg\eta$, where $\eta$ is a term $\eta:\mbb{F}\to\{0,1\}$,
\[
\forall(\mbf{x}\in\mbb{F}).\left(\kappa(\mbf{v})=c\right)
\limply\left[\neg\eta(\mbf{x})\right]
\]
This statement can be rewritten as follows,
\[
\forall(\mbf{x}\in\mbb{F}).\eta(\mbf{x})\limply\left(\kappa(\mbf{v})\not=c\right)
\]
The set of literals in $\eta$ is referred to as a
\emph{counterexample} (CEx) for the prediction $c$, and represents the
negation of a prime implicate for the predicate $[\kappa(\mbf{x})=c]$.
Clearly, both global AXp's and CEx's are irreducible (and so
subset-minimal).
%

%a CEx \emph{breaks} a global AXp if the CEx has a literal
%inconsistent with the global AXp. Similarly, a global AXp can break
%(resp.~a CEx) 

\begin{example} \label{ex:gaxpcex}
  For the DL of~\cref{fig:01a:dl}, let $c=0$, i.e.\ the predicted
  class is 0. It is plain that the predicted class is 0 whenever
  $x_1=1$. Thus, $\{(x_1=1)\}$ is a global abductive explanation for
  class $c=0$.
  Similarly, if $x_2=0$ and $x_4=1$, then the predicted class is again
  guaranteed to be 0. Thus, the other global abductive explanation is
  $\{(x_2=0),(x_4=1)\}$.
  We could use minimal hitting set duality between prime implicants
  and implicates~\cite{rymon-amai94} to list the counterexamples.
  However, we can also directly reason in terms of the DL to uncover
  the CEx's, as shown in~\cref{tab:gaxpcex}.
\end{example}

\begin{table}[t]
  \begin{center}
    \renewcommand{\tabcolsep}{0.5em}
    \renewcommand{\arraystretch}{1.075}
    \begin{tabular}{C{3.0cm}C{7.5cm}} \toprule
      Global AXp's & $\{\{x_1=1\},\{x_2=0,x_4=1\}\}$ \\ \midrule
      Counterexamples & $\{x_1=0,x_2=1\},\{x_1=0,x_4=0\}$ \\ \bottomrule
    \end{tabular}
  \end{center}
  \caption{Global abductive explanations and counterexamples for the
    DL of~\cref{fig:01a:dl}.}
  \label{tab:gaxpcex}
\end{table}

(Local) AXp's, CXp's and global AXp's and CEx's reveal important
relationships between prime implicants and implicates, as discussed
later in~\cref{ssec:morenotes}.

\subsection{Duality Results} \label{ssec:xpdual}

This section overviews duality results in formal explainability, which
have been established in recent years~\cite{inams-aiia20,inms-nips19}.

\jnoteF{%
  NIPS'19 duality result, i.e.\ PIs vs.\ IPs\\
  AI*IA'20 duality result, i.e.\ MUSes vs. MCSes, and relation with PIs/IPs.
}

\paragraph{Duality between AXp's and CXp's.}
%~\\
Given the definition of sets of AXp's and CXp's (see~\eqref{eq:setaxp}
and~\eqref{eq:setcxp}), and by building on Reiter's seminal
work~\cite{reiter-aij87}, recent work~\cite{inams-aiia20} proved the
following duality between minimal hitting sets:
\begin{proposition}[Minimal hitting-set duality between AXp's and CXp's]
  \label{prop:xpdual}
  Given a local explanation problem $\fml{E}$, we have that,
  \begin{enumerate}
  \item $\fml{X}\subseteq\fml{F}$ is an AXp (and so
    $\fml{X}\in\mbb{A}(\fml{E})$)
    iff $\fml{X}$ is an MHS of the CXp's in $\mbb{C}(\fml{E})$.
  \item $\fml{Y}\subseteq\fml{F}$ is a CXp (and so
    $\fml{X}\in\mbb{C}(\fml{E})$)
    iff $\fml{Y}$ is an MHS of the AXp's in $\mbb{A}(\fml{E})$.
  \end{enumerate}
\end{proposition}
We refer to~\cref{prop:xpdual} as MHS duality between AXp's and CXp's.
The previous result has been used in more recent papers for enabling the
enumeration of
explanations~\cite{msgcin-icml21,ims-sat21,hiims-kr21}.

\begin{example}
  For the DL of~\cref{ex:runex01a}, and the instance
  $(\mbf{v},c)=((0,0,1,2),1)$, we have argued
  (see~\cref{ex:waxp01a,ex:axp01a,ex:cxp01a}) that an AXp is $\{1,4\}$
  and that $\{1\}$ and $\{4\}$ are CXp's. Clearly, the AXp is a MHS of
  the CXp's and vice-versa. Hence, we have listed all the AXp's and
  CXp's for this instance.
\end{example}

Furthermore, a consequence of \cref{prop:xpdual} is the following
result:

\begin{proposition} \label{prop:xpdual2}
  Given a classifier function $\kappa:\mbb{F}\to\fml{K}$, defined on a
  set of features $\fml{F}$, a feature $i\in\fml{F}$ is
  included in some AXp iff $i$ is included in some CXp.
\end{proposition}

\paragraph{Duality between global AXp's and
  counterexamples~\cite{inms-nips19}.}
Another minimal hitting-set duality result, different
from~\cref{prop:xpdual}, was investigated in earlier
work~\cite{inms-nips19}, and relates \emph{global} AXp's (i.e.\ not
restricted to be consistent with a specific point $\mbf{v}\in\mbb{F}$)
and counterexamples (see also~\cpageref{def:cex}).
Given the definition of (global) AXp's and
CEx's (see~\cref{ssec:gaxp}), we say that two sets of literals
\emph{break} each other if these have inconsistent literals.
Furthermore,~\cite{inms-nips19} proves the following result, 

\begin{proposition}
  For a global explanation problem, every CEx breaks every global AXp
  and vice-versa. 
\end{proposition}

\jnoteF{Define global explanation problem}

\begin{example}
  From~\cref{ex:gaxpcex}, it is plain to conclude
  (see~\cref{tab:gaxpcex}) that each global abductive explanation
  breaks each counterexample and vice-versa.
\end{example}

\subsection{Additional Notes} \label{ssec:morenotes}

\paragraph{Relationship with non-formal explainability.}
Past work has shown how formal explanations can serve to assess the
quality or rigor of non-formal
explanations~\cite{inms-corr19,nsmims-sat19,ignatiev-ijcai20}.
For example, a non-formal explanation can be \emph{corrected} and made
subset-minimal, using the non-formal explanation as a starting point
for the computation of some other, formal,
explanations~\cite{inms-corr19,ignatiev-ijcai20}.
Moreover, some authors have recently noticed what is referred to as
the \emph{disagreement problem in XAI}~\cite{lakkaraju-corr22a}. From
the perspective of formal explainability, differences in explanations
just represent different AXp's, which can exist. More important, as
discussed in~\cref{sec:qxps}, it is conceptually feasible, and often
practically efficient, to navigate the space of explanations.

\paragraph{Literals based on the equality operator.}
As can be observed, both running examples use literals of the form
$(x_i=u_i)$. The same applies to the definitions of (weak) AXp's and
CXp's. This need not be the case, as discussed
elsewhere~\cite{iims-jair22}. In the case of DTs, more general
literals have been associated with explanations~\cite{iims-jair22},
e.g.\ by describing literals using set membership.
Nevertheless, and for simplicity, in this document we will use
literals that use solely the equality operator.

\paragraph{Prime implicants \& implicates vs.\ MUSes \& MCSes.}
%
%Given a classifier $\kappa$, and an instance $(\mbf{v},c)$, with
%$c=\kappa(\mbf{v})$, abductive explanations are defined with respect
%to $\mbf{v}$, and so are \emph{localized} given $\mbf{v}$.
%
%(...)\\
%
%Observe that a counterexample can be written as follows,
%I.e., the negation of $\eta$ is a prime implicate of the predicate
%$\kappa(\mbf{x})=c$.
%
%(...)\\
%
%Whereas AXp's for a class $c\in\fml{K}$.
%It is important to highlight the differences between the duality
%results in earlier work~\cite{inms-nips19,inams-aiia20}.
%
%(...)\\
%
%\jnote{Finish formalization.}
%
%
For global abductive explanations, the duality result established in
earlier work~\cite{inms-nips19} essentially relates prime implicants
and implicates of some boolean function $\varsigma:\mbb{F}\to\{0,1\}$,
with $\varsigma(\mbf{x})=(\kappa(\mbf{x})=c)$.
In contrast, the duality results established in more recent
work~\cite{inams-aiia20} relate localized AXp's and CXp's, and can
instead be viewed as relating MUSes and MCSes of some inconsistent
formula (see~\cref{sec:cxps} for additional detail).
These results reveal a more fine-grained relationship between prime
implicants and prime implicates, than what is proposed in earlier
work~\cite{rymon-amai94,pimms-ijcai15}.

\paragraph{Formal explainability and model-based diagnosis.}
Although we approach formal explainability as a problem of abduction,
there are other possible ways to represent the problem of
explainability. One well-known example is model-based
diagnosis (MBD)~\cite{reiter-aij87}.
We consider a system description $\mathsf{SD}$ consisting of a set of
first-order logic statements, and a set of constants
$\mathsf{Comp}$, representing the system's components. Each component
may or may not be operating correctly, and we use a predicate
$\mathsf{Ab}$ to indicate whether the component $C_j$ is operating
incorrectly (i.e.\ $\msf{Ab}(C_j)$ holds, denoting abnormal behavior),
or correctly (i.e.\ $\neg\msf{Ab}(C_j)$ holds, denoting normal
behavior). Furthermore, we also assume some observation $\msf{Obs}$
about the system's expected behavior.
In a diagnosis scenario, where $\msf{Obs}$ disagrees with expected
result, it is the case that,
\begin{equation}
  \msf{SD}\cup\{\neg\msf{Ab}(C_1),\neg\msf{Ab}(C_2),\ldots,\neg\msf{Ab}(C_n)\}
  \cup\msf{Obs} \entails \bot
\end{equation}

A conflict set is a (subset)-minimal set $\msf{CS}$ of $\msf{Comp}$
such that,
\begin{equation}
  \msf{SD}\cup\{\neg\msf{Ab}(C_i)\,|\,C_i\in\msf{CS}\}
  \cup\msf{Obs} \entails \bot
\end{equation}
A diagnosis is a subset-minimal set $\Delta$ of $\msf{Comp}$ such
that,
\begin{equation}
\msf{SD}\cup\{\neg\msf{Ab}(C_i)\,|\,C_i\in\msf{Comp}\setminus\Delta\}
\cup\msf{Obs} \nentails \bot
\end{equation}

A simple reduction of the problem of finding abductive explanations to
model based diagnosis, can be organized as follows:
\begin{enumerate}
\item $\msf{SD}$ is given by,
  $\llbracket\land_{i\in\fml{F}}[(x_i=v_i)\lor\msf{Ab}(i)]\rrbracket$,
  where $\llbracket\cdot\rrbracket$ denotes a logic encoding in a
  suitable logic theory.
\item $\msf{Obs}$ is given by
  $\llbracket\kappa(\mbf{x})\not=c\rrbracket$.
\end{enumerate}

Clearly, if all components operate correctly, then the system
description is inconsistent with the stated observation, as
expected. (And in this case the stated observation is \emph{not}
observing $c$.)
Furthermore, it is easy to see that a minimal conflict is an abductive
explanation, and a minimal diagnosis is a contrastive explanation.
It should be underlined that the proposed reduction aims at
simplicity, but it can also be perceived as somewhat artificial.
Clearly, the computation of $\kappa$ could be deemed part of the
system description, such that its operation was forced to be
correct. Other variations could be envisioned.
Given the progress observed in MBD in recent
years~\cite{codish-jair14,msjim-ijcai15}, it would be interesting to
assess the scalability of MBD tools in the context of explainability.

\jnoteF{Write down ideas presented at DX'22.}

%\subsection{Formal Explainability Timeline \& Research Directions}
\subsection{A Timeline for Formal Explainability}

\begin{figure}[t]
  %\begin{flushleft}
  \hspace*{-0.525cm}\scalebox{0.885}{\begin{tikzpicture}[timespan={},timeline width=13.75,timeline height=1,timeline offset=0.1]
  \timeline[custom interval=true]{\bfseries 2019, \bfseries 2020, \bfseries 2021, \bfseries 2022, \bfseries 2023}
  \begin{phases}
    %2019
    \phase{between week=1 and 2 in -0.385,involvement degree=3.25cm}

    %2020
    \phase{between week=2 and 3 in -0.125,involvement degree=3.75cm,phase color=blue!80!cyan}

    %2021
    \phase{between week=2 and 3 in 0.575,involvement degree=3.5cm,phase color=green!50!black}

    %2022
    \phase{between week=3 and 4 in 0.5,involvement degree=4.0cm,phase color=red!90!black}

    \phase{between week=3 and 4 in 1.0,involvement degree=3.75cm,phase color=red!40!yellow}

    \phase{between week=3 and 4 in 1.75,involvement degree=3.75cm,phase color=red!40!green}

    \phase{between week=3 and 4 in 2.35,involvement degree=3.5cm,phase color=blue!50!white}

  \end{phases}

  %2019
  \addmilestone{at=phase-1.90,direction=90:2.0cm,text={XP definitions},text options={above}}
  \begin{scope}[every node/.append style={alias=mynode-1}]
    \addmilestone{at=phase-1.270,direction=270:1.5cm,text={AXp, CXp,
        duality},text options={below}}
  \end{scope}

  %2020
  \addmilestone{at=phase-2.90,direction=90:1.5cm,text={Tractability},text options={above}}
  \begin{scope}[every node/.append style={alias=mynode-2}]
    \addmilestone{at=phase-2.270,direction=270:2.0cm,text={DTs, NBCs, etc.},text options={below}}
  \end{scope}

  %2021
  \addmilestone{at=phase-3.90,direction=90:2.425cm,text={Efficient solutions},text options={above}}
  \begin{scope}[every node/.append style={alias=mynode-3}]
    \addmilestone{at=phase-3.270,direction=270:1.25cm,text={RFs, DLs,
        BTs, etc.},text options={below}}
  \end{scope}

  %2022
  \addmilestone{at=phase-4.90,direction=90:1.5cm,text={Queries},text options={above}}
  \begin{scope}[every node/.append style={alias=mynode-4}]
    \addmilestone{at=phase-4.270,direction=270:2.85cm,text={Member.,
        Enum., etc.},text options={below}}
  \end{scope}

  \addmilestone{at=phase-5.90,direction=90:2.25cm,text={Input distrib.},text options={above}}
  \begin{scope}[every node/.append style={alias=mynode-5}]
    \addmilestone{at=phase-5.270,direction=270:1.5cm,text={Inp.~constr.},text options={below}}
  \end{scope}

  \addmilestone{at=phase-6.90,direction=90:1.5cm,text={Prob. XPs},text options={above}}
  \begin{scope}[every node/.append style={alias=mynode-6}]
    \addmilestone{at=phase-6.270,direction=270:2.25cm,text={DTs, NBCs, etc.},text options={below}}
  \end{scope}

  \addmilestone{at=phase-7.90,direction=90:2.25cm,text={New topics},text options={above}}
  \begin{scope}[every node/.append style={alias=mynode-7}]
    \addmilestone{at=phase-7.270,direction=270:3.125cm,text={Distil., etc.},text options={below}}
  \end{scope}

  %\node[fit=(mynode-1) (mynode-2) (mynode-3)] (f1){};
  %\draw[decorate,decoration=brace,thick] (f1.south east) -- (f1.south west)
  %node[midway,below]{something};
  %\node[fit=(mynode-4) (mynode-5) (mynode-6)] (f2){};
  %\draw[decorate,decoration=brace,thick] (f2.south east) -- (f2.south west)
  %node[midway,below]{something else};
  %\node[fit=(mynode-7) (mynode-8) (mynode-9)] (f3){};
  %\draw[decorate,decoration=brace,thick] (f3.south east) -- (f3.south west)
  %node[midway,below]{hibernate};
\end{tikzpicture}}
  %\end{flushleft}
  \caption{A timeline of research on formal explainability}
  \label{fig:timeline} 
\end{figure}
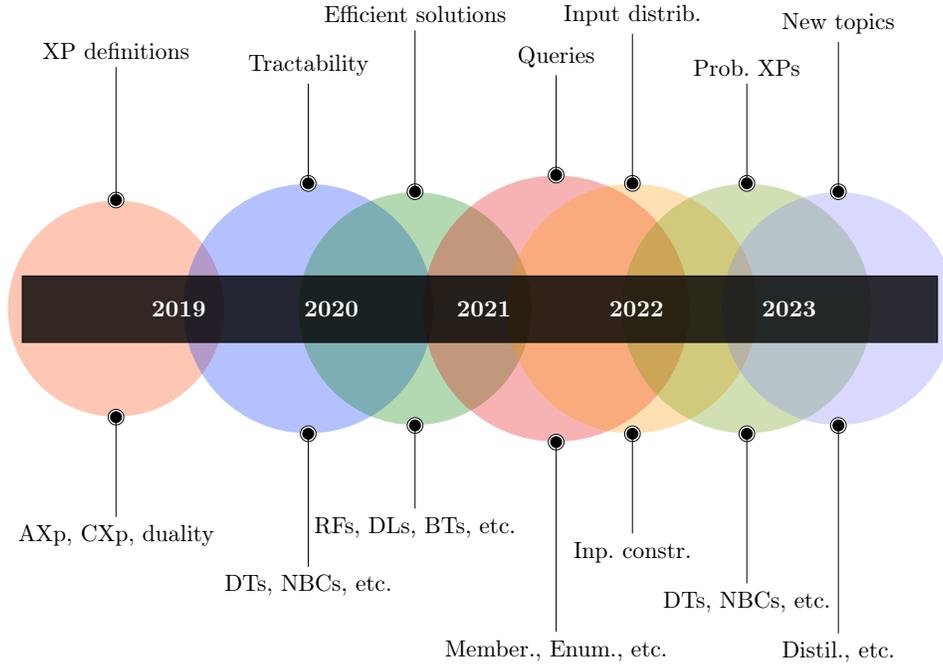

\cref{fig:timeline} depicts the evolution in time of the main areas of
research in formal explainability.
The initial focus (in 2019-2020) was on the definition of
explanations, duality results, but also approaches for the computation
of explanations.
The next major effort (in 2020-2021) was on classifiers exhibiting
tractable computation of one explanation. This was soon followed by
efforts on the efficient computation of explanations even when the
computation of explanations was computationally hard (in 2021-2022).
More recently, there has been research on addressing different
explainability queries (in 2021-2022), tackling input distributions
(started in 2022), and computing probabilistic explanations in practice
(also started in 2022). There are additional topics of research, which
are also discussed in this document.
Although there is ongoing research in most areas of research shown
in~\cref{fig:timeline}, it is also the case that the most recent
topics exhibit more open research questions.
The rest of this paper, overviews the areas of research in formal
explainability shown in~\cref{fig:timeline}.

\jnoteF{Topics: Other kinds of explanations? More duality results?}

\section{Computing Explanations} \label{sec:cxps}

This section covers the computation of explanations, both abductive
and contrastive. The focus are on families of classifiers for which
computing one explanation is computationally hard. The next section
covers families of classifiers for which there exist polynomial-time
algorithms for computing one abductive/contrastive explanation.

%%\subsection{General Case} \label{ssec:gcase}
%
%\begin{enumerate}
%\item Basic ideas
%\item Progress in computing explanations
%\item Neural networks
%\item Decision lists
%\item Random forests
%\item Boosted trees
%\item Topics of research: efficiency of reasoning for NNs/BNs
%\end{enumerate}

\subsection{Progress in Computing Explanations} \label{ssec:xpprog}

\begin{figure}[t]
  \begin{center}
    \scalebox{1.1}{\begin{tikzpicture}
[
  good/.style={
    draw=tgreen3,
    thick,
    fill=tgreen3!35,
    minimum width=0.5cm,
    minimum height=0.5cm,
    rounded corners=1mm,
    font=\tiny\sffamily\bfseries
  },
  bad/.style={
    draw=tred3,
    thick,
    fill=tred3!27,
    minimum width=0.5cm,
    minimum height=0.5cm,
    rounded corners=1mm,
    font=\tiny\sffamily\bfseries
  },
  soso/.style={
    draw=tblue3,
    thick,
    fill=tblue3!27,
    minimum width=0.5cm,
    minimum height=0.5cm,
    rounded corners=1mm,
    font=\tiny\sffamily\bfseries
  }
]

\def\a{1.9}
\def\b{1.9}
\path
(-3,-3.25) node[good]{NBCs~\cite{msgcin-nips20}}
(-1.475,-3.25) node[good]{DTs~\cite{iims-corr20,iims-corr22}}
(-3,-2.575) node[good]{XpGs~\cite{hiims-kr21}}
(-1.65,-2.575) node[good]{GDFs~\cite{hiims-kr21}}
(-2.625,-1.9) node[good]{Monotonic~\cite{msgcin-icml21,cms-cp21}}
(-1.425,-1.225) node[good]{d-DNNF~\cite{hiicams-aaai22}}

(-2.9,0.5) node[soso]{DLs~\cite{ims-sat21}}
(-2.3,1.15) node[soso]{RFs~\cite{ims-ijcai21}}
(-1.5,1.8) node[soso]{BTs~\cite{ignatiev-ijcai20,iisms-aaai22}}

(2,2.075) node[bad]{NNs~\cite{inms-aaai19}}
(2.8,3.05) node[bad]{BNs~\cite{darwiche-ijcai18}}
;
\draw[dotted,thick] (2*\a,0)--(-2*\a,0) (0,2*\b)--(0,-2*\b);
\draw[solid,thick] (-2*\a,-2*\b)--+(0:4*\a);
\draw[solid,thick] (-2*\a,2*\b)--+(0:4*\a);
\draw[solid,thick] (-2*\a,-2*\b)--+(90:4*\b);
\draw[solid,thick] (2*\a,-2*\b)--+(90:4*\b);
\path
(0,-4.5) node{\small Practical scalability (effectiveness)}
(-3,-4.075) node{\scriptsize Effective}
(3,-4.0075) node{\scriptsize Ineffective}
(-4.5,0) node[rotate=90]{\small Computational complexity}
%(-3.9,1.8) node[rotate=90]{\scriptsize NP-hard}
(-4.075,1.8) node[rotate=90]{\scriptsize Computationally hard}
(-4.075,-1.8) node[rotate=90]{\scriptsize Poly-time}
(0,4.075) node{\small Computing one XP}
;
\end{tikzpicture}}
  \end{center}
  \caption{Progress in formal explainability} \label{fig:progress}
\end{figure}
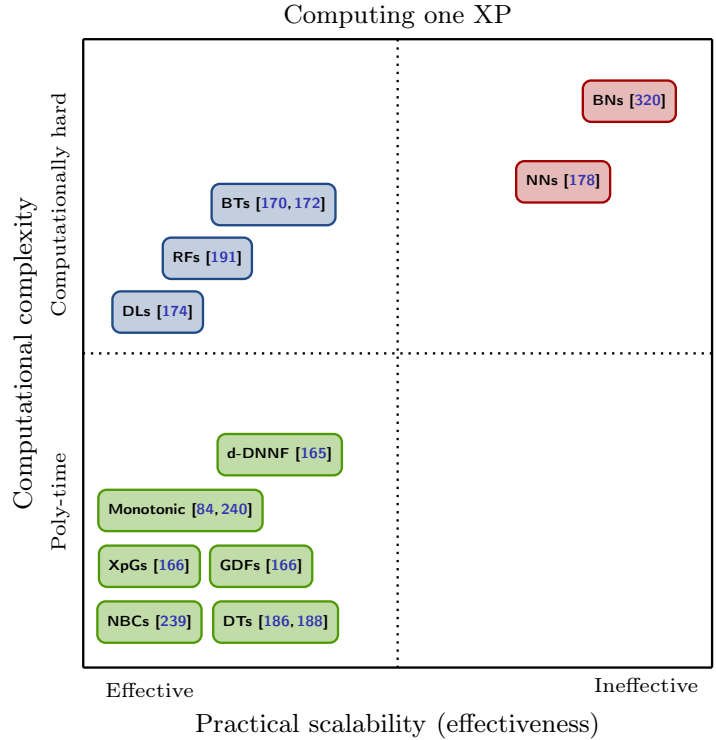

Since 2019, there has been steady progress in the practical
efficiency of computing formal explanations. \cref{fig:progress}
summarizes the observed progress. 
For some families of classifiers, including decision trees, graphs and
diagrams, naive bayes classifiers, monotonic classifiers, restricted
propositional language classifiers and others, it has been shown that
computing one AXp is
tractable~\cite{msgcin-nips20,iims-corr20,msgcin-icml21,hiims-kr21,cms-cp21,hiicams-aaai22,iims-corr22}. For
some other families of classifiers, e.g.\ decision lists and sets,
random forests and tree ensembles, it has been established the
computational hardness of computing one
AXp~\cite{ims-ijcai21,ims-sat21}. However, and also for these families 
of classifiers, existing logical encodings enable the efficient
computation of one
AXp~\cite{inms-corr19,ignatiev-ijcai20,ims-ijcai21,ims-sat21,iisms-aaai22}
in practice.
Finally, for a few other families of
classifiers~\cite{darwiche-ijcai18,inms-aaai19}, computing one AXp is
not only computationally hard, but existing algorithms are not
efficient in practice, at least for large scale ML models.
The next sections analyze some of these results in more detail.

\subsection{General Oracle-Based Approach} \label{ssec:xpgen}

\jnoteF{Outline how to compute explanations in general.}

The main approach for computing explanations is based on exploiting
automated reasoners (e.g.\ SAT, SMT, MILP, etc.) as oracles.
We start by analyzing how to decide whether a subset $\fml{X}$ of
features is a weak AXp. From~\eqref{eq:axp1}, negating twice, we get:
\[
\neg\exists(\mbf{x}\in\mbb{F}).%
\left[\bigwedge\nolimits_{i\in\fml{X}}(x_i=v_i)\right]\land(\kappa(\mbf{x})\not=c)\\
\]
This corresponds to deciding the consistency of a logic formula, as
follows:
\[
\neg
\consistent{\lencode{\left[\bigwedge\nolimits_{i\in\fml{X}}(x_i=v_i)\right]\land(\kappa(\mbf{x})\not=c)}}
\]

The computation of a single AXp or a single CXp can be achieved by
adapting existing algorithms provided a few requirements are met.
First, reasoning in theory $\fml{T}$ is required to be monotone,
i.e.\ \emph{in}consistency is preserved if constraints are added to a
set of constraints, and consistency is preserved if constraints are
removed from a set of constraints.
Second, for computing one AXp, the predicate to consider is:
\begin{equation} \label{eq:predaxp}
  %\begin{align} \label{eq:predaxp}
  \predicate_{\tn{axp}}(\fml{S};\fml{T}, %&
  \fml{F},\kappa,\mbf{v})
  \triangleq %\nonumber \\
  \neg\chkco %&
  \left(\left\llbracket\left(\bigwedge\nolimits_{i\in\fml{S}}(x_i=v_i)\right)\land(\kappa(\mbf{x})\not=c)\right\rrbracket\right)
  %\end{align}
\end{equation}
and for computing one CXp, the predicate to consider is:
\begin{equation} \label{eq:predcxp}
  %\begin{align} \label{eq:predcxp}
  \predicate_{\tn{cxp}}(\fml{S};\fml{T}, %&
  \fml{F},\kappa,\mbf{v})
  \triangleq %\nonumber\\
  \chkco %&
  \left(\left\llbracket\left(\bigwedge\nolimits_{i\in\fml{F}\setminus\fml{S}}(x_i=v_i)\right)\land(\kappa(\mbf{x})\not=c)\right\rrbracket\right)
  %\end{align}
\end{equation}
where, the starting set $\fml{S}$ can be any set that respects the
invariant of the predicate for which it serves as an argument. For
example, for computing a AXp, $\fml{S}$ can represent any weak AXp,
and for computing a CXp, $\fml{S}$ can represent any weak CXp.
(Similar to the case of $\textco$, $\predicate_{\tn{axp}}$ and
$\predicate_{\tn{cxp}}$ are parameterized by $\fml{T}$, $\fml{F}$,
$\kappa$, $\mbf{v}$, and also $c=\kappa(\mbf{v})$. For simplicity,
this parameterization will be left implicit when convenient. Also, the
parameterization on $c=\kappa(\mbf{v})$, given the ones on $\kappa$
and $\mbf{v}$.)
Observe that, since~\eqref{eq:axp1} and~\eqref{eq:cxp1} are monotone,
then $\predicate_{\tn{axp}}$ and $\predicate_{\tn{cxp}}$ are also
monotone with respect to set $\fml{S}$. 
\begin{algorithm}[t]
  \begin{flushleft}
  \hspace*{\algorithmicindent}
  \textbf{Input}: {
    Seed $\fml{S}\subseteq\fml{F}$,
    parameters $\predicate$,
    $\fml{T}$, $\fml{F}$, $\kappa$, $\mbf{v}$}\\
  %Theory $\fml{T}$, Features $\fml{F}$, Classifier $\kappa$, instance $\mbf{v}$}\\
  \hspace*{\algorithmicindent}
  \textbf{Output}: {One XP $\fml{W}$}
\end{flushleft}

\begin{algorithmic}[1]
  \Procedure{$\onexp$}{$\fml{S};\mbb{P},\fml{T},\fml{F},\kappa,\mbf{v}$}
  \State{$\fml{W} \gets \fml{S}$}
  \Comment{Initialization: $\mbb{P}(\fml{W})$ holds}
  \For{$i\in\fml{S}$}
  \Comment{Loop invariant: $\mbb{P}(\fml{W})$ holds}
    \If{$\predicate(\fml{W}\setminus\{i\}; \fml{T},\fml{F},\kappa,\mbf{v})$}
    \State{$\fml{W} \gets \fml{W}\setminus\{i\}$}
    \Comment{Update $\fml{W}$ only if $\mbb{P}(\fml{W}\setminus\{i\})$ holds}
    \EndIf
  \EndFor
  \State{\bfseries{return}~{$\fml{W}$}}
  \Comment{Returned set $\fml{W}$: $\mbb{P}(\fml{W})$ holds}
\EndProcedure
\end{algorithmic}

  \caption{Finding one AXp/CXp} \label{alg:onexp}
\end{algorithm}
Moreover, the monotonicity of $\predicate_{\tn{axp}}$ and
$\predicate_{\tn{cxp}}$ enables adapting standard algorithms for
computing one explanation.
\cref{alg:onexp} illustrates one possible approach%
\footnote{%
This algorithm is referred to as the \emph{deletion-based}
algorithm~\cite{chinneck-jc91}, but it can be traced back to the work 
of Valiant~\cite{valiant-cacm84} (and some authors~\cite{juba-aaai16}
argue that it is implicit in works from the $\text{19}^{\text{th}}$
century~\cite{mill-bk43}).
Although variants of~\cref{alg:onexp} are most often used in practical
settings, there are several alternative algorithms that can also be
used, including QuickXplain~\cite{junker-aaai04},
Progression~\cite{msjb-cav13}, or even
insertion-based~\cite{puget-ecai88}, among
others~\cite{blms-aicom12,bacchus-cav15}.
As illustrated by~\cref{alg:onexp}, it is now known that most of these
algorithms can be formalized in an abstract way, thus allowing them to
be used to solve subset-minimal problems when these problems can be
represented by monotonic predicates~\cite{msjm-aij17}. As argued
earlier, predicates $\predicate_{\tn{axp}}$ (see \cref{eq:predaxp})
and $\predicate_{\tn{cxp}}$ (see \cref{eq:predcxp}) are both
monotonic.}.
For computing one AXp or one CXp, the initial seed set $\fml{S}$
of~\cref{alg:onexp} is set to $\fml{F}$. However, as long as the
precondition $\predicate(\fml{S})$ holds, then any set
$\fml{S}\subseteq\fml{F}$ can be considered.

\begin{example} \label{ex:axpdt}
  We consider the DT of~\cref{fig:runex02a}, and both the computation
  of one AXp and one CXp when $\fml{S}=\{3,4,5\}$.
  For the AXp, it is plain that if features $\{3,4,5\}$ are fixed,
  then the prediction does not change (as shown
  in~\cref{tab:tt02aa}).
  \cref{tab:dtxp02a} summarizes the execution of~\cref{alg:onexp} when
  computing one AXp and one CXp, starting from a set of literals
  (i.e.\ the seed) $\fml{S}=\{3,4,5\}$. (Without additional 
  information, we would start from $\fml{S}=\fml{F}$, and so the table
  would include a few more lines.)
  The difference between the two executions is the result of the
  predicate used.
\end{example}

\begin{table}[t]
  \begin{subtable}[b]{0.485\textwidth}
    \begin{center}
      \begin{tabular}{C{1.5cm}C{0.5cm}C{1.0cm}L{1.5cm}} \toprule
        $\fml{W}$ & $i$ & $\predaxp$ & Decision \\ \toprule
        $\{3,4,5\}$ & 3 & 0          & Keep 3 \\
        $\{3,4,5\}$ & 4 & 1          & Drop 4 \\
        $\{3,5\}$   & 5 & 0          & Keep 5 \\
        \bottomrule
      \end{tabular}
    \end{center}
    \caption{Finding 1 AXp} \label{tab:dtaxp02a}
  \end{subtable}
  \quad
  \begin{subtable}[b]{0.485\textwidth}
    \begin{center}
      \begin{tabular}{C{1.5cm}C{0.5cm}C{1.0cm}L{1.5cm}} \toprule
        $\fml{W}$ & $i$ & $\predcxp$ & Decision \\ \toprule
        $\{3,4,5\}$ & 3 & 1          & Drop 3 \\
        $\{4,5\}$   & 4 & 1          & Drop 4 \\
        $\{5\}$     & 5 & 0          & Keep 5 \\
        \bottomrule
      \end{tabular}
    \end{center}
    \caption{Finding 1 CXp} \label{tab:dtcxp02a}
  \end{subtable}
  \caption{Computation of 1 AXp and 1 CXp starting from seed
    $\fml{S}=\{3,4,5\}$. The partial truth tables
    in~\cref{tab:tt02aa,tab:tt02ab} can be used for computing both
    $\predaxp$ and $\predcxp$.
    \cref{tab:tt02aa} serves to validate whether the prediction
    remains unchanged.
    \cref{tab:tt02ab} serves to assess whether there are points in
    feature space for which the prediction changes.
  }
  \label{tab:dtxp02a}
\end{table}

In some settings, it may be relevant to compute one smallest AXp or
one smallest CXp. Computing one CXp can be achieved by computing a
smallest(-cost) MCS. Hence, a MaxSAT/MaxSMT reasoner can be used in
this case. For AXp's, and given their relationship with MUSes, a
different approach needs to be devised. For a given theory $\fml{T}$,
let
\begin{equation} \label{eq:decide}
  \begin{array}{lcr}
    \decide(\fml{X}) & \quad~{:=}~\quad\quad &
    \consistent{\lencode{\left[\bigwedge\nolimits_{i\in\fml{X}}(x_i=v_i)\right]\land(\kappa(\mbf{x})\not=c)}}\\
  \end{array}
\end{equation}
Furthermore, we assume that $\decide(\fml{X})$ returns a pair
$(\outc,\mu)$, indicating whether the formula is indeed consistent and,
if it is, the computed assignment.
%
%\ifthenelse{\boolean{shownotes}}{
%  \jnote{Include algorithm for computing smallest AXp.}
%}{}
%
\cref{alg:minxp} illustrates the use of dualization for computing one
smallest AXp, and builds on earlier work on computing smallest
MUSes~\cite{iplms-cp15}.
\begin{algorithm}[t]
  \begin{flushleft}
  \hspace*{\algorithmicindent}
  \textbf{Input}: { %Predicate $\predaxp$,
    Algorithm parameterized by $\fml{T}$, $\fml{F}$, $\kappa$, $\mbf{v}$}\\
  %Theory $\fml{T}$, Features $\fml{F}$, Classifier $\kappa$, instance $\mbf{v}$}\\
  \hspace*{\algorithmicindent}
  \textbf{Output}: {Smallest AXp $\fml{M}$}
  \vspace*{-0.175cm}
\end{flushleft}
\begin{algorithmic}[1]
  \Procedure{$\minxp$}{} %$\mbb{P}$ %,\fml{T},\fml{F},\kappa,\mbf{v}
  \State{$\fml{H}\gets\emptyset$}
  \While{\TRUE}
  \State{$\fml{M}\gets\minhs(\fml{H})$}
  \State{$(\outc,\mu)\gets\decide(\fml{M};\fml{T},\fml{F},\kappa,\mbf{v})$}
  \If{$\mathbf{not~}\outc$}
  \State{\bfseries{return}~{$\fml{M}$}}
  \Else
  %%\State{$\mu\gets\model()$}
  \State{$\fml{L}\gets\falselits(\fml{F}\setminus\fml{M},\mu)$}
  \State{$\fml{H}\gets\fml{H}\cup\fml{L}$}
  \EndIf
  \EndWhile
  \EndProcedure
\end{algorithmic}

  \caption{Finding one smallest AXp} \label{alg:minxp}
\end{algorithm}
Any minimum-size hitting set such that the picked (fixed) features
represent a weak AXp must be a smallest AXp.

\paragraph{Relationship with MUSes \& MCSes.}
We can relate the computation of AXp's and CXp's respectively with
the extraction of MUSes and MCSes.
We construct a formula $\varphi=\fml{B}\cup\fml{S}$ in some theory
$\fml{T}$ where the background knowledge $\fml{B}$ corresponds to,
\begin{align}
  \fml{B}\triangleq\lencode{\kappa(\mbf{x})\not=c}
\end{align}
and the soft constraints $\fml{S}$ correspond to,
\begin{align}
  \fml{S}\triangleq\left\{ \lencode{x_i=v_i}\,|\,i\in\fml{F}\right \}
\end{align}
%
%\lencode{\left[\bigwedge\nolimits_{i\in\fml{X}}(x_i=v_i)\right]}
%
Clearly, $\varphi=\fml{B}\cup\fml{S}$ is inconsistent, i.e.\ if all
features are fixed, then the prediction must be $c$.
As a result, an MUS $\fml{U}$ of $\fml{B}\cup\fml{S}$ (i.e.\ a subset
of $\fml{U}$ of $\fml{S}$) is such that $\fml{U}\cup\fml{S}$ is
inconsistent. Thus, $\fml{U}$ represents a subset-minimal set of
features which, if fixed, ensure inconsistency (and so the prediction
must be $c$); hence, $\fml{U}$ is an AXp.
Similarly, an MCS $\fml{C}$ of $\fml{B}\cup\fml{S}$ (i.e.\ a subset of
$\fml{C}$ of $\fml{S}$) is such that
$\fml{B}\cup(\fml{S}\setminus\fml{C})$ is consistent.
Thus, $\fml{C}$ represents a subset-minimal set of features which, if
allowed to change their, ensure consistency (and so the prediction
can be different from $c$); hence, $\fml{C}$ is a CXp.

%\subsection{Decision Lists and Related Classifiers}
\subsection{Explaining Decision Lists}

This section details the computation of AXp's/CXp's in the case of
DLs. Furthermore, it is briefly mentioned the relationship with
computing explanations for DSs and DTs.

\paragraph{Explaining DLs.}
The computation of AXp's has been shown to be computationally hard,
both for DLs and DSs~\cite{ims-sat21}. As a result, the solution
approach is to follow the general approach detailed
in~\cref{ssec:xpgen}. However, we will devise a propositional encoding
and use SAT solvers as NP oracles.

To illustrate the computation of explanations (both AXp's and CXp's),
we consider a DL with the following structure:
\[
\renewcommand{\arraystretch}{1.2}
\begin{array}{lllcl}
  \tn{R$_{1}$:} & \tn{IF~}      & (\tau_1) & \tn{~THEN~} & d_1 \\
  \tn{R$_{2}$:} & \tn{ELSE IF~} & (\tau_2) & \tn{~THEN~} & d_2 \\
  \multicolumn{5}{c}{\cdots} \\
  %%\tn{R$_{2}$:} & \tn{ELSE IF~} & (\tau_2) & \tn{~THEN~} & d_2 \\
  \tn{R$_{j}$:} & \tn{ELSE IF~} & (\tau_j) & \tn{~THEN~} & d_j \\
  \multicolumn{5}{c}{\cdots} \\
  \tn{R$_{n}$:} & \tn{ELSE IF~} & (\tau_n) & \tn{~THEN~} & d_n \\
  \tn{R$_{\tn{\sc{def}}}$:} & \tn{ELSE~} & \quad\quad & \tn{~THEN~} & d_{n+1} %\\
\end{array}
\]
where $d_r\in\fml{K},r=1,\ldots,n+1$, and $\tau_j$ is a conjunction of
literals, or in general some logic formula.

Let $\lencode{\phi}$ denote the propositional CNF encoding of $\phi$.
This encoding can introduce not only a number of additional clauses,
but also fresh propositional variables. The resulting formula will be
represented by $\mathfrak{E}_{\phi}(z_1,\ldots)$, where $z_1$ is a
propositional variable taking value 1 iff $\phi$ takes value 1.
To develop a propositional CNF encoding, we let
$\mathfrak{E}_{\tau_j}(t_j,\ldots)$ represent the clauses associated
with $\lencode{\tau_j}$, i.e.\ the propositional CNF encoding of
$\tau_j, j=1,\ldots,n$, and where $t_j$ represents a new propositional
variable that takes value 1 only in points of feature space where
$\tau_j$ is true. (Additional propositional variables may be used, and
these are represented by $\ldots$ at this stage.)
%
%$\left(t_j\leftrightarrow\lencode{\tau_j}\right)$, %$j=1,\ldots,n$,
%where $t_j$ is a propositional variable, and
%$\lencode{\tau_j}$ is the
%propositional CNF encoding of $\tau_j, j=1,\ldots,n$.
%
Let the target class be $c$ and define the propositional constant
$e_j$ to be 1 iff $d_j$ matches $c$.
%
%%Then, we define the following constraints:

Moreover, since literals may require propositional encodings, let
%$l_i\leftrightarrow\lencode{x_i=v_i}, i=1,\ldots,m$.
$\mathfrak{E}_{x_i=v_i}(l_i,\ldots)$ represent the clauses associated
$\lencode{x_i=v_i}, i=1,\ldots,m$, and where $l_i$ represents a new
propositional variable that takes value 1 only in points of feature
space where $x_i=v_i$. (Additional propositional variables may be
used, and these are represented by $\ldots$ at this stage. Also, we
can envision re-using and sharing common encodings; this will be
discussed in~\cref{ex:dlxp} below.)

Clearly, for some point $\mbf{x}$ in feature space, the prediction
changes if it is the case that,
\begin{enumerate}
\item For $\tau_j$, with $e_j=0$ and $1\le{j}\le{n}$, it is the case
  that $\tau_j$ is true, and for any $1\le{k}<j$, with $e_k=1$,
  $\tau_k$ is false:
  \[
  \left[f_j\leftrightarrow\left(t_j\land\bigwedge\nolimits_{1\le{k}<j,e_k=1}\neg{t_k}\right)\right]
  \]
  where $f_j$ is a new propositional variable, denoting that rule $j$
  with a different prediction would fire (and so it would \emph{flip}
  its previous status). (Clearly, if some other rule $\tn{R}_r$,
  $\tau_r$ with $r<j$ and $e_r=0$, fires then the prediction will also
  change as intended, and this is covered by some other constraint.
  Hence, there is no need to account for such rules.)
\item Moreover, we require that at least one $f_j$, with $e_j=0$ and
  $1\le{j}\le{n}$, to be true:
  \[
  \left(\bigvee\nolimits_{1\le{j}\le{n},e_j=0}{f_j}\right)
  \]
\end{enumerate}

Given the above, we now organize the propositional encoding in two
components, one composed of soft clauses and the other composed of
hard clauses:
\begin{itemize}
\item The set of soft clauses is given by:
  \begin{equation}
    \fml{S}\triangleq\{(l_i), i=1,\ldots,m\}
  \end{equation}
\item The set of hard clauses is given by:
  \begin{align}
    \fml{B}\triangleq&
    \bigwedge\nolimits_{1\le{i}\le{m}}\mathfrak{E}_{x_i=v_i}(l_i,\ldots)
    \land %\\
    %&
    \bigwedge\nolimits_{1\le{j}\le{n}}\mathfrak{E}_{\tau_j}(t_j,\ldots)
    \land \nonumber\\
    &
    \bigwedge\nolimits_{1\le{j}\le{n},e_j=0}\left(f_j\leftrightarrow\left(t_j\land
    \bigwedge\nolimits_{1\le{k}<j,e_k=1}\neg{t_k}\right)\right)
    \land \nonumber\\
    &
    \left(\bigvee\nolimits_{1\le{j}\le{n},e_j=0}{f_j}\right)
  \end{align}
\end{itemize}

It is plain that $\fml{B}\cup\fml{S}$ represents an inconsistent
propositional formula. Any MUS of $\fml{B}\cup\fml{S}$ is a subset of
$\fml{S}$, and represents one AXp. Moreover, any MCS is also a subset
of $\fml{S}$, and represents one CXp. As argued earlier, AXp's are
MHSes of CXp's and
vice-versa~\cite{reiter-aij87,lozinskii-jetai03,inams-aiia20}.
This observation also means that we can use \emph{any} algorithm for
MUS/MCS extraction/enumeration for computing explanations of
DLs~\cite{msm-ijcai20}.

Finally, the propositional encoding proposed above differs slightly
from the one proposed in earlier work~\cite{ims-sat21}, offering a
more streamlined encoding.
%Furthermore, earlier work~\cite{ims-sat21}
%also described the computation of explanations for decision sets.

\begin{example} \label{ex:dlxp}
  Let us investigate how we can encode the computation of one AXp for
  the DL running example (see~\cref{fig:runex01a}).
  The soft clauses are given by,
  \[
  \fml{S}=\{l_1,l_2,l_3,l_4\}
  \]
  where $l_i$ is associated with $\lencode{x_i=v_i}$, i.e.\ it is one
  of the variables used in $\mathfrak{E}_{x_i=v_i}$.
  For simplicity, there is no need to encode $x_i=v_i$ for $i=1,2,3$,
  with $\mbf{v}=(0,0,1,2)$, and so we let 
  $l_1\lequiv\neg{x_1}$, $l_2\lequiv\neg{x_2}$, and
  $l_3\lequiv{x_3}$. However, for $x_4=2$, a dedicated
  propositional encoding is required. Hence, we use three
  propositional variables, $\{x_{41},x_{42},x_{43}\}$, and pick a one
  hot encoding to get
  $\mathfrak{E}_{\tau_3}(l_4,\ldots)\triangleq{l_4}\lequiv{x_{43}}\land\tn{EqualsOne}(x_{41},x_{42},x_{43})$,
  where $x_{43}=1$ iff $x_4=1$.
  There are many propositional encodings for
  $\tn{EqualsOne}(x_{41},x_{42},x_{43})$~\cite{sat-handbook21}
  constraints. One simple solution is,
  \[
  (\neg{x_{41}}\lor\neg{x_{42}})\land
  (\neg{x_{41}}\lor\neg{x_{43}})\land
  (\neg{x_{42}}\lor\neg{x_{43}})\land(x_{41}\lor{x_{42}}\lor{x_{43}})
  \]
  
  For each rule we need to encode $\tau_j$.
  It is immediate to get $t_1\lequiv{x_1}$ for $\tau_1$, and
  $t_2\lequiv{x_2}$ for $\tau_2$. With respect to $\tau_3$, we can use
  the same encoding as above (for $x_4$), and so we get
  $t_3\lequiv{x_{42}}$. (It is clear that the constraints for
  encoding the possible values of $x_4$ would be the same as before,
  and so there is no need to replicate them.)

  Furthermore, we must encode the change of prediction. Since the
  prediction for $(0,0,1,2)$ is 1, then we are interested in rules
  $\tn{R}_1$ and $\tn{R}_3$, since these are the only options to
  change the prediction. As a result, we get,
  \[
  \begin{array}{l}
    f_1\lequiv{t_1}\\
    f_3\lequiv\left({t_3}\land\neg{t_2}\right)\\
  \end{array}
  \]
  indicating that the prediction changes if either $t_1$ is true
  (i.e.\ rule $\tn{R}_1$ fires), or $t_3$ is true \emph{and} $t_2$ is
  false (i.e.\ rule $\tn{R}_3$ fires and rule $\tn{R}_2$ does not
  fire, and earlier rules are already covered by other constraints).
  Finally, we need the constraint $(f_1\lor{f_3})$, to enforce
  that a change of prediction will take place.

  Given the above, we get that,
  \begin{align*}
    \fml{B} = &
    (l_1\lequiv\neg{x_1})\land
    (l_2\lequiv\neg{x_2})\land
    (l_3\lequiv{x_3})\land
    (l_4\lequiv{x_{43}})\land\\
    & \tn{EqualsOne}(x_{41},x_{42},x_{43})\land\\
    & (t_1\lequiv{x_1})\land
    (t_2\lequiv{x_2})\land
    (t_3\lequiv{x_{42}})\land\\
    &
    (f_1\lequiv{t_1})\land(f_3\lequiv(t_3\land\neg{t_2}))\land(f_1\lor{f_3})
  \end{align*}
  
  Moreover, and by inspection, the tuple $(\fml{S},\fml{B})$ can be
  simplified to,
  \[
  \fml{S}=\{\neg{x_1},\neg{x_2},x_3,x_{43}\}
  \]
  and,
  \begin{equation*}
    %\begin{align*}
    \fml{B}= %&
    \tn{EqualsOne}(x_{41},x_{42},x_{43})\land %\\
    %&
    (f_1\lequiv{x_1})\land(f_3\lequiv(x_{42}\land\neg{x_2}))\land(f_1\lor{f_3})
    %\end{align*}
  \end{equation*}
  It is apparent that $\fml{B}\cup\fml{S}$ is inconsistent.
  Finally, we can also observe that for tuple $(\fml{S}',\fml{B})$
  with, 
  \begin{equation*}
    \begin{array}{lcl}
      %\begin{align*}
      \fml{S}' & = & %&
      \{\neg{x_1},x_{43}\} \\[2pt]
      \fml{B} & = & \tn{EqualsOne}(x_{41},x_{42},x_{43})\land %\\
      %&
      (f_1\lequiv{x_1})\land(f_3\lequiv(x_{42}\land\neg{x_2}))\land(f_1\lor{f_3})
      %\end{align*}
    \end{array}
  \end{equation*}
  $\fml{B}\cup\fml{S}'$ is still inconsistent. Thus,
  $\fml{S}'=\{\neg{x_1},x_{43}\}$ is an unsatisfiable subset (which we
  can prove to be irreducible), and so $\fml{X}=\{1,4\}$ is a weak AXp
  (which we can prove to be an AXp).
\end{example}

Existing results indicate that the computation of explanations for DLs
is very efficient in practice~\cite{ims-sat21}.

\subsection{From DLs to DTs \& DSs}

\paragraph{Explaining DTs as DLs.}
A conceptually straightforward approach for explaining DTs is to
represent a DT as a DL. Hence, the propositional encoding proposed
above for explaining DLs can also be used for explaining DTs.
Nevertheless, as argued in~\cref{ssec:xpdt}, in the case of DTs there
are polynomial time algorithms for computing one abductive
explanation, and there are polynomial time algorithms for enumerating
all the contrastive explanations.

\paragraph{The case of DSs.}
The fact that DSs are unordered raises a number of technical
difficulties, including the fact that, if there can be rules that
predict different classes and fire on the same input, then the
classifier does not compute a function. This is referred to as
\emph{overlap}~\cite{ipnms-ijcar18}.
If the rules predicting each class are represented as a DNF, then one
call to an NP oracle suffices to decide whether overlap exists.

\jnoteF{Formalize the identification of overlap.}

Let a DS be represented by a set of DNF formulas, one for each
class in $\fml{K}$. Moreover, let $z_r$ represent the value computed
for DNF $r$, which predicts class $c_r\in\fml{K}$.
There exists \emph{no} overlap is the following condition does
\emph{not} hold:
\[
\sum\nolimits_{r=1}^{K}z_r>1
\]
If we also want to ensure that there is a prediction for any point in
feature space, then we can instead require that the following
constraint is inconsistent:
\begin{equation} \label{eq:dsok}
  \sum\nolimits_{r=1}^{K}z_r\not=1
\end{equation}
i.e.\ we want the sum to be equal to 1 on each point of feature
space. (Clearly, the constraint above must be inconsistent given the 
logic encoding of the classifier. The actual encoding of each $z_r$
will depend on how the DNFs are represented, and there is no
restriction of considering purely boolean classifiers.)

Under the standard assumption that \eqref{eq:dsok} is inconsistent,
then the encoding proposed for DLs can be adapted to the case of DSs.
We will have to encode each term (i.e.\ each unordered rule), and then
encode the disjunction of terms for each DNF. We will briefly outline
a propositional encoding for computing abductive (and contrastive)
explanations.
The approach differs from the DL case since we do not have order in
the rules. Hence, each class is analyzed as a DNF. As usual, we
consider an instance $(\mbf{v},c_s)$, where $c_s\in\fml{K}$ is the
predicted class. The constraints for the encodings are organized as
follows:
\begin{enumerate}
\item The DNF $r$ of class $c_r\in\fml{K}$, it is a disjunction of
  $n_r$ terms $\tau_{rj},r=1,\ldots,n_r$.
\item Each term $\tau_{rj}$ is encoded into
  $\encls{\tau_{rj}}(t_{rj},\ldots)$, such that $t_j=1$ iff the term
  $\tau_{rj}$ takes value 1.
\item The literals of the form $(x_i=v_i)$ will be encoded into the
  clauses $\encls{x_i=v_i}(l_i,\ldots)$.\\
  (As mentioned earlier in this section in the case of DLs, the
  encoding from the original feature variables to propositional
  variables is assumed in all these encodings.)
\item The soft clauses will be $(l_i)$, $i=1,\ldots,m$.
\item A class $c_r$ is picked iff $p_r=1$, where $p_i$ is a fresh
  propositional variable.
  Hence, we define $p_r$, for class $c_r$ as follows:
  $p_r\lequiv\left(\lor_{j=1}^{n_r}{t_{rj}}\right)$ 
\item The prediction changes if $p_s=0$. \\
  (Observe that we could instead introduce another propositional
  variable $s$, defined as follows
  $s\lequiv\lor_{c_r\in\fml{K}\setminus\{c_s\}}p_r$, such that $s=1$
  would mean that the prediction changes. However, this is
  unnecessary.)
%\item $s$ must not hold, since we do not want to change the
%  prediction.
\end{enumerate}
Given the above, we can write down a propositional encoding for a DS
classifier which respects~\eqref{eq:dsok}. The set of soft clauses is
given by:
\begin{equation}
  \fml{S}\triangleq\{(l_i),i=1,\ldots,m\}
\end{equation}
The set of hard clauses is given by:
\begin{align}
  \fml{B}\triangleq&
  \bigwedge\nolimits_{1\le{i}\le{m}}\encls{x_i=v_i}(l_i,\ldots)
  \land %\\
  %&
  \bigwedge_{1\le{r}\le{K}}
  \bigwedge\nolimits_{1\le{j}\le{n_r}}\encls{\tau_{rj}}(t_{rj},\ldots)
  \land \nonumber \\
  &
  \bigwedge\nolimits_{1\le{j}\le{K}}p_r\lequiv\left(\lor_{j=1}^{n_r}{t_{rj}}\right)
  %\nonumber
\end{align}

\jnoteF{%
  \begin{enumerate}
  \item The approach differs from the DL case since we do not have
    order in the rules. Hence, each class is analyzed as a DNF.
  \item The DNF $r$ of class $c_r$, it is a disjunction of $n_r$ terms
    $\tau_j,r=1,\ldots,n_r$. 
  \item Each term $\tau_j$ is encoded into
    $\encls{\tau_j}(t_j,\ldots)$, such that $t_j=1$ iff the term
    $\tau_j$ takes value $\TRUE$
  \item The literals of the form $(x_i=v_i)$ will be encoded into the
    clauses $\encls{x_i=v_i}(l_i,\ldots)$,
  \item The soft clauses will be $(l_i)$, $i=1,\ldots,m$
  \item To pick class $c_r$, $p_r$: $p_r\lequiv\left(\lor_j{t_j}\right)$
  \item To change the prediction, $s$:
    $s\lequiv\lor_{c_r\in\fml{K}\setminus\{c\}}p_r$
  \item Alternatively, it suffices to require $p_s=0$, where $c_s$ is
    the class associated with the instance.
  \item $s$ must not hold, since we do not want to change the
    prediction.
  \item Given the above, we can write down a propositional encoding
    for a DS classifier which respects~\eqref{eq:dsok}.
  \end{enumerate}
}

\jnoteF{Clarify how this is done in the case of DSs.}

\jnoteF{Brief comment on DSs.}

\subsection{Explaining Neural Networks}
%\subsection{Explaining NNs}
%\paragraph{Explaining NNs.}
%
To illustrate the modeling flexibility of the approach proposed in the
previous section, let us develop an MILP/SMT encoding for the problem
of computing one AXp for a neural network.
The encoding to be used is based on the MILP representation of NNs
proposed in earlier work~\cite{fischetti-cj18} and is illustrated with
the NN running example of~\cref{fig:runex03a}. The MILP encoding is
shown in~\cref{fig:runex03b:milp}.

To decide whether a set $\fml{X}\subseteq\fml{F}$ is a weak AXp, we
would have to decide the \emph{in}consistency of (adapted
from~\cref{fig:runex03b:milp}):
\begin{align} \label{eq:nn:axp}
  & \bigwedge\nolimits_{i\in\fml{X}}(x_i=v_i)\land \nonumber \nonumber
  \\
  & [(x_1+x_2-0.5=t_1-s_1)\land
    (z_1=1\limply{t_1}\le0)\land
    (z_1=0\limply{s_1}\le0)\land \nonumber \\
    & (o_1=(t_1>0))\land(t_1\ge0)\land(s_1\ge0)]\land
  \left[(o_1\not=1)\right]
\end{align}
where
$\fml{D}_{x_1}=\fml{D}_{x_2}=\fml{D}_{z_1}=\fml{D}_{z_2}=\{0,1\}$ and
$\fml{D}_{s_1}=\fml{D}_{t_1}=\mbb{R}$. (The definition of domains 
introduces a mild abuse of notation, since the indices used are the
names of feature variables and not the names of features. However, the
meaning is clear.)

\begin{figure}[t]
  %%\begin{adjustbox}{minipage=\linewidth,frame}
  %\medskip\medskip
  \begin{center}
    %\centering
    \begin{subfigure}[b]{0.4\textwidth}
      \centering
      \begin{minipage}{\textwidth}
        \[
        \begin{array}{l}
          x_1+x_2-0.5=t_1-s_1\\
          z_1=1\limply{t_1}\le0\\
          z_1=0\limply{s_1}\le0\\
          o_1=(t_1>0)\\[2pt]
          x_1,x_2,z_1,o_1\in\{0,1\}\\
          t_1,s_1\ge0\\[5pt]
        \end{array}
        \]
        %%\vspace*{0.75cm}
      \end{minipage}
      \caption{Logic representation~\cite{fischetti-cj18}}
      %%of NN~\cite{fischetti-cj18}
      %\caption{Logic representation of NN~\cite{fischetti-cj18}}
      \label{fig:runex03b:milp}
    \end{subfigure}
    \begin{subfigure}[b]{0.15\textwidth}
      ~~~
    \end{subfigure}
    %\medskip
    %
    \begin{subfigure}[b]{0.4\textwidth}
      \centering
      \begin{minipage}{\textwidth}
        \[
        \begin{array}{l}
          1+0-0.5=0.5-0\\
          1\lor{0.5}\le0\\
          0\lor{0}\le0\\
          1=(0.5>0)\\[2pt]
          x_1=1, x_2=0,
          z_1=0, o_1=1\\
          t_1=0.5,s_1=0\\[5pt]
        \end{array}
        \]
      \end{minipage}
      \caption{Instance $(\mbf{x},c)=((1,0),1)$}
      \label{fig:runex03b:chk10}
    \end{subfigure}
    %
    %\begin{subfigure}[b]{0.05\textwidth}
    %  ~~~
    %\end{subfigure}

    \medskip

    \begin{subfigure}[b]{0.4\textwidth}
      \centering
      \begin{minipage}{\textwidth}
        \[
        \begin{array}{l}
          0+0-0.5=0-0.0\\
          0\lor{0}\le0\\
          1\lor{0.5}\le0\\
          0=(0>0)\\[2pt]
          x_1=0, x_2=0,
          z_1=1, o_1=0\\
          t_1=0,s_1=0.5\\[5pt]
        \end{array}
        \]
      \end{minipage}
      \caption{Checking $(x_1,x_2)=(0,0)$}
      \label{exnn01:chk00}
    \end{subfigure}
    \begin{subfigure}[b]{0.15\textwidth}
      ~~~
    \end{subfigure}
    \begin{subfigure}[b]{0.4\textwidth}
      \centering
      \begin{minipage}{\textwidth}
        \[
        \begin{array}{l}
          1+1-0.5=1.5-0\\
          1\lor{1.5}\le0\\
          0\lor{0}\le0\\
          1=(1.5>0)\\[2pt]
          x_1=1, x_2=1,
          z_1=0, o_1=1\\
          t_1=1.5,s_1=0\\[5pt]
        \end{array}
        \]
      \end{minipage}
      \caption{Checking $(x_1,x_2)=(1,1)$}
      \label{fig:runex03b:chk11}
    \end{subfigure}

    %\medskip\medskip
    \caption{Computing one AXp with an NN}
    %\label{exnn01:all}
    \label{fig:runex03b}

  \end{center}
  %%\end{adjustbox}
\end{figure}

\begin{example}
  To compute an AXp for the NN running example, we could iteratively
  call an MILP solver on \eqref{eq:nn:axp}, starting from $\fml{F}$
  and iteratively removing features (see~\cref{alg:onexp}).
  However, the very simple encoding for the example NN allows us to
  analyze the constraints without calling an MILP reasoner. The
  analysis is summarized in~\cref{fig:runex03b}.
  We first consider allowing $x_1$ to take any value. In this case,
  this means allowing $x_1$ to take value 0 (besides the value 1 it is
  assigned to). As can be observed (see~\cref{fig:runex03b:chk10}), the
  prediction is allowed to change (actually, in this case it is forced
  to change). Hence, the feature $1$ must be included in the AXp.
  In contrast, by changing $x_2$ from 0 to 1, the prediction
  cannot change (see~\autoref{fig:runex03b:chk11}).
  This means that, if the other features remain unchanged, the
  prediction is 1, no matter the value taken by $x_2$.
  Hence, the feature $2$ is dropped from the working set of features.
  As a result, the AXp in this case is $\fml{X}=\{1\}$.
\end{example}

The computation of AXp's in the case of NNs was investigated in
earlier work on computing formal explanations~\cite{inms-aaai19}.
However, and in contrast with the families of classifiers studied
earlier in this section, the computation of AXp's/CXp's in the case of
NNs scales up to a few tens of neurons.
It is plain that the ability to efficiently compute AXp's/CXp's for
NNs will track the ability to reason efficiently about NNs.
Although there have been steady improvements on reasoners for
NNs~\cite{barrett-cav17,barrett-cav19,barrett-fto21}, it is also the
case that scalability continues to be a challenge.

\subsection{Other Families of Classifiers}

%\subsection{Tree Ensembles}

\paragraph{Tree ensembles (TEs).}
Based on the general approach detailed in~\cref{ssec:xpgen}, there
have been proposals for computing explanations for boosted
trees (BTs)~\cite{inms-corr19,ignatiev-ijcai20,iisms-aaai22}, and
random forests (RFs)~\cite{ims-ijcai21,mazure-cikm21}.
For RFs, it has been shown that the decision problem of computing one
AXp is complete for $\ddp$~\cite{ims-ijcai21}. Nevertheless, the
proposed encodings~\cite{ims-ijcai21}, which are purely propositional,
enable computing AXp's/CXp's for RFs with thousands of nodes. At
present, such RF sizes are representative of what is commonly deployed
in practical applications.
It should be noted that the existing propositional encodings consider
the organization of RFs as proposed originally~\cite{breiman-ss01},
i.e.\ the class is picked by majority voting. For other ways of
selecting the chosen class, the encoding is not purely propositional.
For BTs, the most recent results also confirm the scalability to
classifiers deployed in practical settings.

%\subsection{Other Models \& Current Research Directions}
%\paragraph{Neural networks \& bayesian network classifiers.}
\paragraph{Bayesian network classifiers.}
The explanations of Bayesian network classifiers (BNCs) have been 
studied since 2018~\cite{darwiche-ijcai18,darwiche-aaai19}. Whereas in
the case of NNs, SMT and MILP solvers were used, and followed the
approach outlined in~\cref{ssec:xpgen}, in the case of BNCs,
explanations are computed using compilation into a canonical
representation (see~\cref{ssec:comp} below). However, and similarly to
NNs, scalability is currently a challenge.

\subsection{An Alternative -- Compilation-Based Approaches}
\label{ssec:comp}

One alternative to the computation of AXp's and CXp's as proposed in
the previous sections is to compile the explanations into some
canonical representation, from which the explanations can then be
queried for. Such compilation-based approaches have been studied in a
number of
works~\cite{darwiche-ijcai18,darwiche-aaai19,darwiche-ecai20,darwiche-kr20,darwiche-pods20,darwiche-aaai22,darwiche-jlli22-web}.

Past work has focused on binary classification with binary
features. The extension to non-binary classification and non-binary 
features raises a number of challenges. Another limitation is that
canonical representations are worst-case exponential, and the
worst-case behavior is commonly observed.
For example, the performance gap between the two approaches in solving
related problems is often
significant~\cite{darwiche-aaai22,iims-jair22}.

\jnoteF{Brief comments on the use in computing AXp's of compilation
  into some canonical representation.}

%\subsection{Research Directions}
%%\paragraph{Current research directions.}
%%~\\
%
%The ability to devise more efficient tools to reason about NNs
%represents a critical topic of research. Significant improvements in
%the tools used to reason about NNs would allow explaining more complex
%classifiers, and so extend the rage of applicability of formal
%explainability.

%\subsection{Random Forests}
%
%\subsection{Boosted Trees}
%
%\subsection{Neural Networks}
%
%\subsection{Research Topics}
%
\jnoteF{Topic: efficiency of reasoning for NNs/BNs}

\section{Tractable Explanations} \label{sec:txps}

%\begin{enumerate}
%\item Decision trees
%\item Explanation graphs
%\item Monotonic classifiers
%\item Naive Bayes classifiers
%\item Propositional languages
%\item Other families of classifiers
%\item Topics of research: other families of classifiers?
%\end{enumerate}

Since 2020, several tractability results have been established in
formal
explainability~\cite{msgcin-nips20,iims-corr20,msgcin-icml21,hiims-kr21,cms-cp21,hiicams-aaai22,iims-jair22}.
Most of these tractability results concern the computation of one
explanation, and apply both to computing one AXp or one
CXp~\cite{cms-cp21}. However, there are examples of families of
classifiers for which there exist polynomial delay algorithms for
enumeration of explanations~\cite{msgcin-nips20}, or even for
computing all (contrastive) explanations~\cite{hiims-kr21,iims-jair22}.

\subsection{Decision Trees} \label{ssec:xpdt} % iims-corr20,iims-jair22

Given a classification problem for a DT, and an instance
$(\mbf{v},c)$, a set of literals is consistent with $c$ as long there
is at least one inconsistent literal for any path that predicts a
class other than $c$.

\paragraph{Abductive explanations.}
Given the observation above, a simple algorithm for computing one AXp
is organized as follows:
\begin{enumerate}
\item For each path $Q_k$ with prediction other than $c$, let $I_k$
  denote the features which take values inconsistent with the path.
\item Pick a subset-minimal hitting set $H$ of all the sets $I_k$.
\item Clearly, as long as the features in $H$ are fixed, then at least
  one literal in each path $Q_k$ will be inconsistent, and so the
  prediction is guaranteed to be $c$.
\end{enumerate}
It is well-known that there exists simple polynomial time algorithms
for computing one subset-minimal hitting set~\cite{gottlob-sjc95}.
Hence, the proposed algorithm runs in polynomial time.

\begin{example} \label{ex:axp02a}
  For the DT of the second running example (see~\cref{fig:02a:dt}),
  and instance $(\mbf{v},c)=((0,0,1,0,1),1)$, we have the following
  sets:
  \begin{itemize}
  \item $Q_1=\langle1,2,4,6\rangle$, with set $I_1=\{3\}$.
  \item $Q_2=\langle1,2,4,7,10,14\rangle$, with set $I_2=\{5\}$.
  \item $Q_3=\langle1,2,5,8,12\rangle$, with set $I_3=\{2,5\}$.
  \end{itemize}
  Clearly, an MHS of $\{I_1,I_2,I_3\}$ is $\{3,5\}$, which represents
  a weak AXp for the given instance. It is simple to conclude that it
  is irreducible, and so it effectively represents an AXp. In
  addition, it is also plain to establish that there are no other
  AXp's.
  Finally, it should be noted that the abductive explanation computed
  above concurs with what was presented in~\cref{ex:xp02a}, where a
  truth-table was used to justify the abductive explanations. (Of
  course, construction of the truth table would not in general be
  realistic, whereas the algorithm proposed above runs in linear time
  on the size of the DT.)
\end{example}

The simple algorithm described above was first proposed in earlier 
work~\cite{iims-corr20}.
Nevertheless, one can envision other algorithms, which offer more
flexibility~\cite{iims-jair22}. (For example, the algorithm described
below allows for constraints on the inputs, in cases for which not all
points in feature space are possible.)

\paragraph{Abductive explanations by propositional Horn encoding.}
A more flexible approach (see~\cref{sec:ixps}) is the representation
of the problem of computing one AXp as the problem of computing one
MUS (or one MCS) of an inconsistent Horn formula\footnote{%
  Since we have tractability, the formulation can be geared towards
  computing one MUS or instead computing one MCS.}.
There are simple encodings that are worst-case quadratic on the size
of the DT~\cite{iims-jair22}. We describe one encoding that is linear
on the size of the DT~\cite{iims-jair22}.

Let us consider a path $P_k$, with prediction $c\in\fml{K}$.
Moreover, let $\fml{Q}$ denote the paths yielding a prediction other
than $c$. Since the prediction is $c$, then any path in $\fml{Q}$ has
some feature for which the allowed values are inconsistent with
$\mbf{v}$. We say that the paths in $\fml{Q}$ are \emph{blocked}.
(To be clear, a path is blocked as long as some of its literals are
inconsistent.)

For each feature $i$ associated with some node of path $P_k$,
introduce a variable $u_i$. $u_i$ denotes whether feature $i$ is
deemed \emph{universal}, i.e.\ feature $i$ is not included in the
AXp that we will be computing.
(Our goal is to find a subset maximal set of features that can be
deemed universal, such that all the paths resulting in a prediction
other than $c$ remain blocked. Alternatively, we seek to find a
subset-minimal set of features to declare non-universal or fixed, such
that paths with a prediction other than $c$ remain blocked.)
Furthermore, for each DT node $r$, introduce variable $b_r$, denoting
that all sub-paths from node $r$ to any terminal node labeled
$d\in\fml{K}\setminus\{c\}$ must be blocked, i.e.\ some literal in the
sub-path must remain inconsistent. (Our goal is to guarantee that all
paths to terminal nodes labeled $d\in\fml{K}\setminus\{c\}$ remain
blocked even when some variables are allowed to become universal.)

%Let $\Phi(P_k)\subseteq\fml{F}$ denote the features tested along path
%$P_k$.
%
%Then, the soft clauses $\fml{S}$ are given by
%$\{(u_i)\,|\,i\in\mrm{\Phi}(P_k)\}$,
%
The soft clauses $\fml{S}$ are given by $\{(u_i)\,|\,i\in\fml{F}\}$,
i.e.\ one would ideally want to
declare universal as many features as possible, thus minimizing the
size of the explanation. (As noted above, we will settle for finding
subset-maximal solutions.)
We describe next the hard constraints $\fml{H}$ for representing
consistent assignments to the $u_i$ variables.

We proceed to describe the proposed Horn encoding. Here, we opt to
describe first the Horn encoding for computing one AXp\footnote{%
  As discussed in recent work~\cite{iims-jair22}, different types of
  AXp's can be computed in the case of DTs; we specifically
  consider the so-called path-unrestricted AXp's.}.
The hard constraints are created as follows:
\begin{enumerate}[label=\textbf{B\arabic*.},ref=\textbf{B\arabic*},
    leftmargin=*] %nosep,
  \label{enum:cases}
\item For the root node $r$, add the constraint $\top\limply{b_r}$.\\
  (The root node must be blocked.)
  \label{enum:cases:stp01}
\item For each terminal node $r$ with prediction $c$, add the
  constraint $\top\limply{b_r}$.\\
  (Each terminal node with prediction $c$ is also blocked. Also,
  observe that this condition is on the node, not on the path.)
  \label{enum:cases:stp02}
\item For each terminal node $r$ with prediction
  $d\in\fml{K}\setminus\{c\}$, add the constraint
  ${b_r}\limply\bot$.\\
  (Terminal nodes predicting $d\not=c$ cannot be blocked. Also, and as
  above, observe that this condition is on the node, not on the path.)
  \label{enum:cases:stp03}
\item For a node $r$ associated with feature $i$, and connected to the
  child node $s$, such that the edge value(s) is(are)
  \emph{consistent} with the value of feature $i$ in $\mbf{v}$, add
  the constraint $b_r\limply{b_s}$.\\
  (If all sub-paths from node $r$ must be blocked, then all sub-paths
  from node $s$ must all be blocked, independently of the value taken
  by feature $i$.)
  \label{enum:cases:stp04}
\item For a node $r$ associated with feature $i$, and connected to the
  child node $s$, such that the edge value(s) is(are)
  \emph{inconsistent} with the value of feature $i$ in $\mbf{v}$, add
  the constraint $b_r\land{u_i}\limply{b_s}$.\\
  (In this case, the blocking condition along an edge inconsistent
  with the value of feature $i$ in $\mbf{v}$ is only relevant if the
  feature is deemed universal.)
  \label{enum:cases:stp05}
\end{enumerate}

\begin{example} \label{ex:runex02d1}
  For the running example of~\cref{fig:02a:dt}, let
  $(\mbf{v},c)=((0,0,1,0,1),1)$.
  As dictated by the proposed Horn encoding, two sets of variables are
  introduced. The first set represents the variables denoting whether
  a feature is universal, corresponding to 5 variables:
  $\{u_1,u_2,u_3,u_4,u_5\}$. The second set represents the variables
  denoting whether a node is blocked, corresponding to 15 variables:
  $\{b_1,b_2,b_3,b_4,b_5,b_6,b_7,b_8,b_9,b_{10},b_{11},b_{12},b_{13},b_{14},b_{15}\}$.
  The resulting propositional Horn encoding contains hard ($\fml{H}$)
  and soft ($\fml{B}$) constraints, and it is organized as shown
  in~\cref{ex:tab01}.
  \begin{table}[t]
    \begin{center}
      %\smallskip
    \begin{tabular}{cc} \toprule 
      Hard constraint type & Horn clauses \\ \toprule
      \Cref{enum:cases:stp01} & $\{(b_1)\}$ %Root must be blocked
      \\ \midrule
      \Cref{enum:cases:stp02} &
      $\{(b_3),(b_9),(b_{11}),(b_{13}),(b_{15})\}$
      \\ \midrule
      \Cref{enum:cases:stp03} &
      $\{(\neg{b_6}),(\neg{b_{12}}),(\neg{b_{14}})\}$
      \\ \midrule
      \Cref{enum:cases:stp04} &
      $\begin{array}{l}
        \{\,
        (b_1\limply{b_2}),(b_2\limply{b_4}),(b_4\limply{b_7}),(b_5\limply{b_8}),\\
        ~~(b_7\limply{b_{10}}),(b_8\limply{b_{13}}),(b_{10}\limply{b_{15}})
        \,\}
      \end{array}$
      \\ \midrule
      \Cref{enum:cases:stp05} &
      $\begin{array}{l}
        \{\,
        (b_1\land{u_1}\limply{b_3}),(b_2\land{u_2}\limply{b_5}),
        (b_4\land{u_3}\limply{b_6}), \\
        ~~(b_5\land{u_4}\limply{b_9}),(b_7\land{u_4}\limply{b_{11}}),
        (b_8\land{u_5}\limply{b_{12}}),\\
        ~~(b_{10}\land{u_5}\limply{b_{14}})
        \,\}
      \end{array}$
      \\
      \toprule
      Soft constraints, $\fml{S}$ & $\{(u_1),(u_2),(u_3),(u_4),(u_5)\}$ \\
      \bottomrule
    \end{tabular}
  \end{center}
  %\smallskip
    \caption{Horn clauses for the DT of~\cref{fig:02a:dt} for
      computing one AXp with $(\mbf{v},c)=((0,0,1,0,1),1)$}
    \label{ex:tab01}
  \end{table}
   
  It is easy to see that, if $u_1=u_2=u_3=u_4=u_5=1$, then $\fml{B}$
  is falsified. Concretely, 
  $(b_1)\land%
  (b_1\limply{b_2})\land%
  (b_2\limply{b_4})\land({u_3})\land%
  (b_4\land{u_3}\limply{b_6})\land(\neg{b_6})\nentails\bot$.
  The goal is then to find a maximal subset $\fml{M}$ of $\fml{S}$
  such that $\fml{M}\cup\fml{B}$ is consistent. (Alternatively, the
  algorithm finds a minimal set $\fml{C}\subseteq\fml{S}$, such that
  $\fml{S}\setminus\fml{C}\cup\fml{H}$ is consistent.)
  For this concrete example, one such minimal set is obtained by
  picking $u_1=u_2=u_4=1$ and $u_3=u_5=0$, and by setting
  $b_1=b_2=b_3=b_4=b_5=b_7=b_8=b_9=b_{10}=b_{11}=b_{13}=b_{15}=1$
  and
  $b_6=b_{12}=b_{14}=0$. Hence, all clauses are satisfied, and so
  $\{3,5\}$ is a weak AXp. An MCS
  extractor~\cite{mshjpb-ijcai13,mpms-ijcai15,mipms-sat16} would
  confirm that $\{3,5\}$ is subset-minimal, and so it is an AXp.
\end{example}

\paragraph{Contrastive explanations.}
In the case of DTs, recent work devised efficient polynomial-time
algorithms for computing (in fact listing all) contrastive
explanations~\cite{hiims-kr21}.
The main ideas can be summarized as follows:
\begin{itemize}
\item For each path $Q_k$ with prediction other than $c$, list the
  features with literals inconsistent with the instance as set $I_k$.
\item Remove any set $I_l$ that is a superset of some other set $I_k$.
\item Each of the remaining sets $I_k$ is a CXp.
\end{itemize}
Since the number of paths is polynomial (in fact linear) on the number
of tree nodes, then we have a polynomial time algorithm for listing
all contrastive explanations.

\begin{example} \label{ex:cxp02a}
  Using the sets $I_k$ computed in~\cref{ex:axp02a}, we observe that
  $I_3$ is a superset of $I_2$, and so it has to be dropped.
  As a result, $I_1$ and $I_2$ each represent an CXp.
  Furthermore, we can confirm again~\cref{prop:xpdual}, since the only
  AXp for this instance, i.e.\ $\{3,5\}$ is an MHS of the two CXp's,
  i.e. $\{3\}$ and $\{5\}$, and vice-versa.
\end{example}

As noted in recent work, the fact that there exists a polynomial time
algorithm to enumerate all CXp's, implies that are quasi-polynomial
algorithms for the enumeration of
AXp's~\cite{hiims-kr21,iims-jair22}. This is discussed in further
detail in~\cref{ssec:enum}.

Moreover, and although the Horn encoding proposed earlier for
computing one AXp could also be used for computing one CXp, there is
no real need for that, given the simplicity of the algorithm for
enumerating all CXp's of a DT.

%\paragraph{Contrastive explanations by propositional Horn encoding.}
%%
%The Horn encoding, that was proposed earlier for computing one AXp,
%can be used for computing one

\subsection{Monotonic Classifiers}
%\ifthenelse{\boolean{shownotes}}{
%  
%  \jnote{Include the case of monotonic classifiers}
%}{}
This section illustrates how one AXp (or CXp) can be computed in the
case of monotonic classifiers.
In the case of classifiers for which computing the prediction runs in
polynomial time on the size of the classifier, recent work proved that
there exist polynomial time algorithms both for computing one AXp and
one CXp~\cite{msgcin-icml21}.
The algorithms are dual of each other; as a result, we will just
detail the computation of one AXp. Nevertheless, we present the
pseudo-code for both algorithms.
For computing one AXp, we maintain two vectors, one yielding a
lower bound on the computed class, i.e.\ $\mbf{v}_L$, and another
yielding an upper bound on the computed class, i.e.\ $\mbf{v}_U$.
In the case of one AXp, the algorithm requires that
$\kappa(\mbf{v}_L)=\kappa(\mbf{v}_U)=\kappa(\mbf{v})$.
The goal is to allow features to take any possible value in their
domain, i.e.\ to make them \emph{universal} and so we will use an
auxiliary function $\mkuniv$:

\begin{algorithmic}
  \State{$\mbf{v}_{\tn{L}}\gets(v_{\tn{L}_1},\ldots,\lambda(i),\ldots,v_{\tn{L}_N})$}
  \State{$\mbf{v}_{\tn{U}}\gets(v_{\tn{U}_1},\ldots,\mu(i),\ldots,v_{\tn{U}_N})$}
  \State{$(\fml{A},\fml{B})\gets(\fml{A}\setminus\{i\},\fml{B}\cup\{i\})$}
  \State{\Return{$(\mbf{v}_{\tn{L}},\mbf{v}_{\tn{U}},\fml{A},\fml{B})$}}
\end{algorithmic}

If making a feature universal allows the prediction to change, then
the feature must be \emph{fixed} again (to the value dictated by
$\mbf{v}$), and for that we use the auxiliary function $\mkfix$:

\begin{algorithmic}
  \State{$\mbf{v}_{\tn{L}}\gets(v_{\tn{L}_1},\ldots,v_i,\ldots,v_{\tn{L}_N})$}
  \State{$\mbf{v}_{\tn{U}}\gets(v_{\tn{U}_1},\ldots,v_i,\ldots,v_{\tn{U}_N})$}
  \State{$(\fml{A},\fml{B})\gets(\fml{A}\setminus\{i\},\fml{B}\cup\{i\})$}
  \State{\Return{$(\mbf{v}_{\tn{L}},\mbf{v}_{\tn{U}},\fml{A},\fml{B})$}}
\end{algorithmic}

Given these auxiliary functions, the computation of one AXp is shown
in~\cref{alg:mono-oneaxp} (and the computation of one CXp is shown
in~\cref{alg:mono-onecxp}).

\begin{algorithm}[t]
  \begin{flushleft}
  \hspace*{\algorithmicindent}
  \textbf{Input}: {
    Features $\fml{F}$,
    Seed $\fml{S}\subseteq\fml{F}$,
    Point in $\mbb{F}$ $\mbf{v}$\\
    \hspace*{\algorithmicindent}}
  \textbf{Output}: {One AXp $\fml{P}$}
\end{flushleft}

\begin{algorithmic}[1]
  \Procedure{$\oneaxp$}{$\fml{F};\fml{S},\mbf{v}$}
  \State{%
    \label{alg:axp:ln01}$\mbf{v}_{\tn{L}}\gets(v_1,\ldots,v_N)$}
  \State{%
    \label{alg:axp:ln02}$\mbf{v}_{\tn{U}}\gets(v_1,\ldots,v_N)$} \Comment{Ensures: $\kappa(\mbf{v}_{\tn{L}})=\kappa(\mbf{v}_{\tn{U}})$}
  \State{%
    \label{alg:axp:ln03}$(\fml{C},\fml{D},\fml{P})\gets(\fml{F},\emptyset,\emptyset)$}
  \ForAll{\label{alg:axp:ln04}$i\in\fml{S}$}
\Comment{Require: $\kappa(\mbf{v}_{\tn{L}})=\kappa(\mbf{v}_{\tn{U}})$, given $\fml{S}$}
  \State{\label{alg:axp:ln05}$(\mbf{v}_{\tn{L}},\mbf{v}_{\tn{U}},\fml{C},\fml{D})\gets\mkuniv(i,\mbf{v},\mbf{v}_{\tn{L}},\mbf{v}_{\tn{U}},\fml{C},\fml{D})$}
  \label{alg:axp:ln06}\EndFor
  \ForAll{%
    \label{alg:axp:ln07}$i\in\fml{F}\setminus\fml{S}$}
  \Comment{Loop~inv.:~$\kappa(\mbf{v}_{\tn{L}})=\kappa(\mbf{v}_{\tn{U}})$}
  \State{\label{alg:axp:ln08}$(\mbf{v}_{\tn{L}},\mbf{v}_{\tn{U}},\fml{C},\fml{D})\gets\mkuniv(i,\mbf{v},\mbf{v}_{\tn{L}},\mbf{v}_{\tn{U}},\fml{C},\fml{D})$}
  \If{\label{alg:axp:ln09}$\kappa(\mbf{v}_{\tn{L}})\not=\kappa(\mbf{v}_{\tn{U}})$}
  \Comment{If invariant broken, fix it}
  \State{\label{alg:axp:ln10}$(\mbf{v}_{\tn{L}},\mbf{v}_{\tn{U}},\fml{D},\fml{P})\gets\mkfix(i,\mbf{v},\mbf{v}_{\tn{L}},\mbf{v}_{\tn{U}},\fml{D},\fml{P})$}
  \EndIf
  \EndFor
  \State{\label{alg:axp:ln13}\Return{$\fml{P}$}}
  \EndProcedure
\end{algorithmic}

  \caption{Computing one AXp for a monotonic classifier}
  \label{alg:mono-oneaxp}
\end{algorithm}

\begin{algorithm}[t]
  \begin{flushleft}
  \hspace*{\algorithmicindent}
  \textbf{Input}: {
    Features $\fml{F}$,
    Seed $\fml{S}\subseteq\fml{F}$,
    Point in $\mbb{F}$ $\mbf{v}$\\
    \hspace*{\algorithmicindent}}
  \textbf{Output}: {One CXp $\fml{P}$}
\end{flushleft}

\begin{algorithmic}[1]
  \Procedure{$\onecxp$}{$\fml{F};\fml{S},\mbf{v}$}
  \State{%
    \label{alg:cxp:ln01}$\mbf{v}_{\tn{L}}\gets(\lambda(1),\ldots,\lambda(N))$}
  \State{%
    \label{alg:cxp:ln02}$\mbf{v}_{\tn{U}}\gets(\mu(1),\ldots,\mu(N))$}\Comment{Ensures: $\kappa(\mbf{v}_{\tn{L}})\not=\kappa(\mbf{v}_{\tn{U}})$}
  \State{%
    \label{alg:cxp:ln03}$(\fml{C},\fml{D},\fml{P})\gets(\fml{F},\emptyset,\emptyset)$}
  \ForAll{\label{alg:cxp:ln04}$i\in\fml{S}$}
  \Comment{Require: $\kappa(\mbf{v}_{\tn{L}})\not=\kappa(\mbf{v}_{\tn{U}})$, given $\fml{S}$}
  \State{%
    \label{alg:cxp:ln05}$(\mbf{v}_{\tn{L}},\mbf{v}_{\tn{U}},\fml{C},\fml{D})\gets\mkfix(i,\mbf{v},\mbf{v}_{\tn{L}},\mbf{v}_{\tn{U}},\fml{C},\fml{D})$}
  \label{alg:cxp:ln06}\EndFor
  \ForAll{%
    \label{alg:cxp:ln07}$i\in\fml{F}\setminus\fml{S}$}
  \Comment{Loop~inv.:~$\kappa(\mbf{v}_{\tn{L}})\not=\kappa(\mbf{v}_{\tn{U}})$}
  \State{\label{alg:cxp:ln08}$(\mbf{v}_{\tn{L}},\mbf{v}_{\tn{U}},\fml{C},\fml{D})\gets\mkfix(i,\mbf{v},\mbf{v}_{\tn{L}},\mbf{v}_{\tn{U}},\fml{C},\fml{D})$}
  \If{\label{alg:cxp:ln09}$\kappa(\mbf{v}_{\tn{L}})=\kappa(\mbf{v}_{\tn{U}})$}\Comment{If invariant broken, fix it}
  \State{\label{alg:cxp:ln10}$(\mbf{v}_{\tn{L}},\mbf{v}_{\tn{U}},\fml{D},\fml{P})\gets\mkuniv(i,\mbf{v},\mbf{v}_{\tn{L}},\mbf{v}_{\tn{U}},\fml{D},\fml{P})$}
  \EndIf
  \EndFor
  \State{\label{alg:cxp:ln13}\Return{$\fml{P}$}}
  \EndProcedure
\end{algorithmic}

  \caption{Computing one CXp for a monotonic classifier}
  \label{alg:mono-onecxp}
\end{algorithm}

The algorithm starts from some set $\fml{S}\subseteq\fml{F}$ (which
can be the empty set) of universal features, which is required to
ensure that $\kappa(\mbf{v}_L)=\kappa(\mbf{v}_U)=\kappa(\mbf{v})$, and
iteratively attempts to add features to set $\fml{S}$, i.e.\ to make
them universal. Monotonicity of entailment (and the discussion in
previous sections) ensures soundness of the algorithm.

\begin{example}

  For the monotonic classifier of~\cref{fig:runex04a}, and instance
  $((Q,X,H,R),\allowbreak{M})=((10,10,5,0),A)$, we show how one AXp can be
  computed.
  For each feature $i$, $1\le{i}\le4$, $\lambda(i)=0$ and $\mu(i)=10$.
  Moreover, features are analyzed in order:
  $\langle1,2,3,4\rangle$; the order is arbitrary.
  The algorithm's execution is summarized in~\autoref{tab:axp-ex}.

  \begin{table}[t]
  %%\begin{center}
  \begin{flushleft}
    \hspace*{-0.175cm}
    \scalebox{0.75125}{ %0.9125
      \renewcommand{\tabcolsep}{0.35em}
      \begin{tabular}{|c|cc|cc|cc|c|cc|} \toprule
        \multirow{2}{*}{Feat.} &
        \multicolumn{2}{c|}{Initial values} &
        \multicolumn{2}{c|}{Changed values} &
        \multicolumn{2}{c|}{Predictions} &
        \multirow{2}{*}{Dec.} &
        \multicolumn{2}{c|}{Resulting values} \\
        \cline{2-7}\cline{9-10}%\midrule
        &
        $\mbf{v}_{\tn{L}}$ & $\mbf{v}_{\tn{U}}$ & 
        $\mbf{v}_{\tn{L}}$ & $\mbf{v}_{\tn{U}}$ & 
        $\kappa(\mbf{v}_{\tn{L}})$ & $\kappa(\mbf{v}_{\tn{U}})$ & 
        &
        $\mbf{v}_{\tn{L}}$ & $\mbf{v}_{\tn{U}}$
        \\ \toprule
        1 &
        $(10{,}10{,}5{,}0)$ & $(10{,}10{,}5{,}0)$ &
        $(0{,}10{,}5{,}0)$ & $(10{,}10{,}5{,}0)$ &
        $C$ & $A$ &
        \keepmark &  %keep 1
        $(10{,}10{,}5{,}0)$ & $(10{,}10{,}5{,}0)$
        \\ \midrule
        2 &
        $(10{,}10{,}5{,}0)$ & $(10{,}10{,}5{,}0)$ &
        $(10{,}0{,}5{,}0)$ & $(10{,}10{,}5{,}0)$ &
        $E$ & $A$ &
        \keepmark &  %keep 2
        $(10{,}10{,}5{,}0)$ & $(10{,}10{,}5{,}0)$
        \\ \midrule
        3 &
        $(10{,}10{,}5{,}0)$ & $(10{,}10{,}5{,}0)$ &
        $(10{,}10{,}0{,}0)$ & $(10{,}10{,}5{,}0)$ &
        $A$ & $A$ &
        \dropmark &  %drop 3 
        $(10{,}10{,}0{,}0)$ & $(10{,}10{,}10{,}0)$
        \\ \midrule
        4 &
        $(10{,}10{,}0{,}0)$ & $(10{,}10{,}10{,}0)$ &
        $(10{,}10{,}0{,}0)$ & $(10{,}10{,}10{,}10)$ &
        $A$ & $A$ &
        \dropmark &  %drop 4
        $(10{,}10{,}0{,}0)$ & $(10{,}10{,}10{,}10)$
        \\
        \bottomrule
      \end{tabular}
    }
    %%\end{center}
  \end{flushleft}
  \caption{Execution of algorithm for finding one AXp} \label{tab:axp-ex}
\end{table}

  As can be observed, features 1 and 2 are kept as part of the
  AXp, and features 3 and 4 are dropped from the AXp.
  Thus, the AXp for the given instance is $\{1,2\}$, representing the
  literals $\{Q=10,X=10\}$.
\end{example}

Besides monotonic classifiers, recent work that similar ideas have
been shown to apply in the case of other (related) families of
classifiers~\cite{cms-cp21}.

%%\subsection{Explanation Graphs} % hiims-kr21

%%\subsection{Monotonic Classifiers} % msgcin-icml21,cms-cp21

%\subsection{Other Families of Classifiers \& Research Directions}
\subsection{Other Families of Classifiers}

A number of additional tractability results have been uncovered.
Recent work~\cite{hiims-kr21} showed that the computation of
explanations for decision graphs~\cite{oliver-tr92}, decision diagrams
and trees could be unified, and explanations computed in polynomial
time.
\jnoteF{Explanation graphs...} % hiims-kr21
For NBCs, it is now known that a smallest explanation can be computed
in polynomial time~\cite{msgcin-nips20}, that there exists a
polynomial delay algorithm for enumeration of abductive explanations
(but it is straightforward to apply the same ideas to the case of
CXp's).
In the case of classifiers represented as propositional languages,
including the broad class d-DNNF, it has been shown that there exist
polynomial time algorithms for computing one
AXp/CXp~\cite{hiicams-aaai22}.
%
%For another broad class of classifiers, concretely monotonic
%classifiers, it has been shown the existence of polynomial-time
%algorithms for computing one AXp/CXp~\cite{msgcin-icml21}. The same
%ideas have been shown to apply in the case of other (related) families
%of classifiers~\cite{cms-cp21}.

%\subsection{Naive Bayes Classifiers} % msgcin-nips20
%
%\subsection{Propositional Languages} % hiicams-aaai22
%
%\subsection{Other Families of Classifiers} % see above
%

%\subsection{Research Directions}
%

\jnoteF{Topic: other families of classifiers?}

\section{Explainability Queries} \label{sec:qxps}

%\begin{enumerate}
%\item Enumeration
%\item Membership
%\item Other queries
%\item Topics of research: counting, enumeration
%\end{enumerate}

Besides the computation of explanations, recent research
considered a number of explainability
queries~\cite{marquis-kr20,msgcin-nips20,inams-aiia20,hiims-kr21,marquis-kr21,hims-aaai23,hcmpms-tacas23}.
This section considers two concrete queries: enumeration of
explanations and feature membership. Additional queries have been
investigated in the listed references.

Enumeration addresses a crucial problem in explainability. If a human
decision maker does not accept the (abductive or contrastive)
explanation provided by an explanation tool, how can one compute some
other explanation, assuming one exists? Most non-formal explainability
approaches do not propose a solution to this problem.
The problem of feature membership is to decide whether some (possibly
sensitive) feature is included in some explanation of an instance,
among all possible explanations. Feature membership is relevant when
assessing whether a classifier can exhibit bias.

\subsection{Enumeration of Explanations} \label{ssec:enum}

Given an explanation problem, and some set of already computed
explanations (AXp's and/or CXp's), the query of enumeration of
explainability is to find one explanation (AXp or CXp) among those
that are not included in the set of explanations.

For NBCs, it has been shown that there is a polynomial-delay algorithm
for the enumeration AXp's~\cite{msgcin-nips20}. A similar approach
yields a solution for the enumeration of CXp's.

For most other families of classifiers, it is conceptually simple to
devise algorithms that enumerate CXp's, without the need of computing
or enumerating AXp's. In contrast, the enumeration of AXp's is
obtained through duality between AXp's and CXp's
(see~\cref{ssec:xpdual} and additional detail
in~\cite{inams-aiia20}).
One solution for enumerating AXp's is to compute all CXp's, and then
use hitting set dualization for computing the AXp's. Unfortunately,
the number of CXp's is often exponential, and this may prevent the
enumeration of any AXp.
Thus, and building on fairly recent work on the enumeration of
MUSes~\cite{lpmms-cj16}, the solution is to iteratively compute
AXp's/CXp's by exploiting hitting set duality, using a SAT solver for
iteratively picking a set of features to serve as a seed for either
computing one AXp or one CXp.
A number of recent works have reported results on the enumeration of
explanations~\cite{msgcin-icml21,ims-sat21,hiicams-aaai22,iisms-aaai22}.
The query of explanation enumeration has also been studied in terms of
its complexity~\cite{marquis-kr20,marquis-kr21}.
%
%
%i.e.\ the MARCO algorithm for MUS enumeration~\cite{lpmms-cj16},
%one solutionfor the enumeration
%
%For most other families of classifiers, practical enumeration of
%explanations has been obtained
%
% The basic algorithmic mimics the MARCO
%algorithm for MUS/MCS enumeration~\cite{lpmms-cj16}.
%
%
%Experimental results have been reported in recent
%work~\cite{msgcin-icml21,ims-sat21,hiicams-aaai22,iisms-aaai22}.
%
One important observation is that, for families of classifiers for 
which computing one explanation is poly-time, then the enumeration of
the next explanation (either AXp or CXp) requires a single call to an
NP oracle~\cite{msgcin-icml21}.

A general-purpose approach for the enumeration of explanations is
shown in~\cref{alg:allxp}\footnote{%
The algorithm mimics the on-demand MUS enumeration algorithm proposed
elsewhere~\cite{lpmms-cj16,pms-aaai13,liffiton-cpaior13}, which
enumerates both MUSes and MCSes. There are several other alternative
MUS enumeration algorithms, which could also be
considered~\cite{reiter-aij87,greiner-aij89,khachiyan-jalg96,kavvadias-wae99,boros-esa03,stuckey-padl05,boros-cocoon05,kavvadias-jgaa05,boros-dam06,liffiton-cpaior13,bacchus-cav15,bacchus-cpaior16,narodytska-ijcai18,bendik-atva18,luo-sc19,kavvadias-tcs20,bendik-tacas20,bendik-lpar20}.
For some of these algorithms, a first required step is the complete
enumeration of MCSes, for which a wealth of algorithms also
exists~\cite{liffiton-jar08,felfernig-aiedam12,mlms-hvc12,mshjpb-ijcai13,lagniez-aaai14,mpms-ijcai15,mipms-sat16,pmjms-aaai18,pmjms-sat17,lagniez-ijcai18}.
Furthermore, there is a tight relationship between MUS/MCS enumeration
and several other computational
problems~\cite{msm-ijcai20,msjm-aij17}, which allows devising generic
algorithms for solving families of related problems.}.
\begin{algorithm}[t]
  \begin{flushleft}
  \hspace*{\algorithmicindent}
  \textbf{Input}: {
    Parameters $\predaxp$, $\predcxp$,
    $\fml{T}$, $\fml{F}$, $\kappa$, $\mbf{v}$} %\\
  %Theory $\fml{T}$, Features $\fml{F}$, Classifier $\kappa$, instance $\mbf{v}$}\\
  \hspace*{\algorithmicindent}
  %\textbf{Output}: {One XP $\fml{W}$}
  %
\end{flushleft}

\begin{algorithmic}[1]
  \State{%
    \label{alg:exp:ln01}$\fml{H}\gets\emptyset$}%
  \Comment{$\fml{H}$ defined on set $U=\{u_1,\ldots,u_m\}$}
  \Repeat\label{alg:exp:ln02} 
  \State{%
    \label{alg:exp:ln03}$(\outc,\mbf{u})\gets\SAT(\fml{H})$}
  \If{\label{alg:exp:ln04}$\outc=\TRUE$}
  \State{\label{alg:exp:ln05}$\fml{S}\gets\{i\in\fml{F}\,|\,u_i=0\}$}%
  \Comment{$\fml{S}$: \emph{fixed} features}
  \State{\label{alg:exp:ln06}$\fml{U}\gets\{i\in\fml{F}\,|\,u_i=1\}$}%
  \Comment{$\fml{U}$: \emph{universal} features; $\fml{F}=\fml{S}\cup\fml{U}$}
  \If{\label{alg:exp:ln07}$\predcxp(\fml{U};\fml{T},\fml{F},\kappa,\mbf{v})$}
  \Comment{$\fml{U}\supseteq$ some CXp}
  \State{\label{alg:exp:ln08}$\fml{P}\gets\onexp(\fml{U};\predcxp,\fml{T},\fml{F},\kappa,\mbf{v})$}
  \State{\label{alg:exp:ln09}$\prtcxp(\fml{P})$}
  \State{\label{alg:exp:ln10}$\fml{H}\gets\fml{H}\cup\{(\lor_{i\in\fml{P}}\neg{u_i})\}$}
  \Else\label{alg:exp:ln11}
  \Comment{$\fml{S}\supseteq$ some AXp}
  \State{\label{alg:exp:ln12}$\fml{P}\gets\onexp(\fml{S};\predaxp,\fml{T},\fml{F},\kappa,\mbf{v})$}
  \State{\label{alg:exp:ln13}$\prtaxp(\fml{P})$}
  \State{\label{alg:exp:ln14}$\fml{H}\gets\fml{H}\cup\{(\lor_{i\in\fml{P}}{u_i})\}$}
  %\State{\label{alg:exp:ln09}$\fml{P}\gets\axp(\fml{F},\fml{S},\mbf{u})$}
  %\State{\label{alg:exp:ln10}$\prtaxp(\fml{P})$}
  %\State{\label{alg:exp:ln11}$\fml{H}\gets\fml{H}\cup\{(\lor_{i\in\fml{P}}u_i)\}$}
  %
  \EndIf
  \EndIf
  \Until{\label{alg:exp:ln15}$\outc=\FALSE$}
\end{algorithmic}
%  \begin{comment}
%  \STATE{$\mbf{v}_{\tn{LB}}\gets\mbf{v}$}
%  \STATE{$\mbf{v}_{\tn{UB}}\gets\mbf{v}$}
%  %(v_1,\ldots,v_N)
%  \STATE{$(\fml{C},\fml{D},\fml{P})\gets(\fml{F},\emptyset,\emptyset)$}
%  \FORALL{$i\in\fml{F}$}
%  \STATE{$\mbf{v}_{\tn{LB}}\gets(v_{\tn{LB}_1},\ldots,\lambda(i),\ldots,v_{\tn{LB}_N})$}
%  \STATE{$\mbf{v}_{\tn{UB}}\gets(v_{\tn{UB}_1},\ldots,\mu(i),\ldots,v_{\tn{UB}_N})$}
%  \STATE{$(\fml{C},\fml{D})\gets(\fml{C}\setminus\{i\},\fml{D}\cup\{i\})$}
%  \If{$\kappa(\mbf{v}_{\tn{LB}})\not=\kappa(\mbf{v}_{\tn{UB}})$}
%  \STATE{$(\fml{D},\fml{P})\gets(\fml{D}\setminus\{i\},\fml{P}\cup\{i\})$}
%  \STATE{$\mbf{v}_{\tn{LB}}\gets(v_{\tn{LB}_1},\ldots,v_i,\ldots,v_{\tn{LB}_N})$}
%  \STATE{$\mbf{v}_{\tn{UB}}\gets(v_{\tn{UB}_1},\ldots,v_i,\ldots,v_{\tn{UB}_N})$}
%  \ENDIf
%  \ENDFOR
%  \end{comment}

  \caption{Finding all AXp/CXp} \label{alg:allxp}
\end{algorithm}
Variants of this algorithm have been studied in recent
work~\cite{inams-aiia20,hiims-kr21,msgcin-icml21,ims-sat21,iisms-aaai22,hiicams-aaai22}.

\begin{example}
  For the DT of~\cref{fig:runex02a}, \cref{tab:enum1,tab:enum2} show
  possible executions of the explanation enumeration algorithm.
  The difference between the two tables is the assignments picked by
  the SAT solver.
  (\cref{tab:tt02aa,tab:tt02ab} are used to decide the values of the
  predicates tested in the algorithm's execution.)
  Depending on that assignment $\mbf{u}$, either there is a pick of
  features that changes the prediction or there is none. If the
  prediction can be changed, then one CXp is computed. Otherwise,
  one AXp is computed. In both cases,~\cref{alg:onexp} is used, but a
  different predicate is considered in each case.
  The clause added after each AXp/CXp is computed prevents the
  repetition of explanations. The algorithm terminates when all
  AXp's/CXp's have been enumerated.
\end{example}

\begin{table}[t]
  \begin{center}
    \renewcommand{\arraystretch}{1.025}
    \begin{tabular}{C{1.0cm}C{2.25cm}C{2.0cm}C{1.5cm}C{1.5cm}C{1.5cm}C{1.5cm}}
      \toprule
      Iter. & $\mbf{u}$ & $\fml{S}$ & $\predcxp(\cdot)$ &
      AXp & CXp & Clause 
      \\ \toprule
      1 & $(1,1,1,1,1)$ & $\emptyset$ & 1 & -- & $\{3\}$ & $(\neg{u_3})$ 
      \\ \toprule
      2 & $(1,1,0,1,1)$ & $\{3\}$     & 1 & -- & $\{5\}$ & $(\neg{u_5})$ 
      \\ \toprule
      3 & $(1,1,0,1,0)$ & $\{3,5\}$   & 0 & $\{3,5\}$ & -- & $({u_3}\lor{u_5})$ 
      \\ \toprule
      5 & $[\outc=\FALSE]$ & -- & -- & -- & -- & --
      \\ %\toprule
      \bottomrule
    \end{tabular}
  \end{center}
  \caption{Example execution of enumeration algorithm} \label{tab:enum1}
\end{table}

\begin{table}[t]
  \begin{center}
    \renewcommand{\arraystretch}{1.025}
    \begin{tabular}{C{1.0cm}C{2.25cm}C{2.0cm}C{1.5cm}C{1.5cm}C{1.5cm}C{1.5cm}}
      \toprule
      Iter. & $\mbf{u}$ & $\fml{S}$ & $\predcxp(\cdot)$ &
      AXp & CXp & Clause 
      \\ \toprule
      1 & $(0,0,0,0,0)$ & $\{1,2,3,4,5\}$ & 0 & $\{3,5\}$ & -- & $({u_3}\lor{u_5})$ 
      \\ \toprule
      2 & $(0,0,1,0,0)$ & $\{1,2,4,5\}$ & 1 & -- & $\{3\}$ & $(\neg{u_3})$ 
      \\ \toprule
      3 & $(0,0,1,0,1)$ & $\{1,2,4\}$ & 1 & -- & $\{5\}$ & $(\neg{u_5})$
      \\ \toprule
      5 & $[\outc=\FALSE]$ & -- & -- & -- & -- & --
      \\ %\toprule
      \bottomrule
    \end{tabular}
  \end{center}
  \caption{Another possible execution of enumeration algorithm} \label{tab:enum2}
\end{table}

As noted in recent work~\cite{hiims-kr21,iims-jair22}, and in the case
of DTs, the fact that all CXp's are computed in poly-time enables
using more efficient quasi-polynomial algorithms for the enumeration
of AXp's (e.g.\ using well-known results in monotone
dualization~\cite{khachiyan-jalg96}).

\jnoteF{Add pseudo-code for enumeration? Also, cite variants of
  enumeration algorithms: MARCO, etc.}

\subsection{Explanation Membership}

The problem of deciding whether a given (possibly sensitive) feature
is included in some explanation is referred to as the \emph{feature
  membership problem} (FMP)~\cite{hiims-kr21}.
\begin{definition}[FMP]
  Given an explanation problem $\fml{E}_L=(\fml{M},\mbf{v},c)$, with
  $\fml{M}=(\fml{F},\mbb{D},\mbb{F},\fml{K},\kappa)$, and some target
  feature $t\in\fml{F}$, the feature membership problem is to decide
  whether there exists an AXp (resp.~CXp) $\fml{X}\subseteq\fml{F}$
  ($\fml{Y}\subseteq\fml{F}$) such that $t\in\fml{X}$ (resp.~$t\in\fml{Y}$)
\end{definition}
%
% Carlos suggests using lean kernels!...
%
It should be observed that FMP is tightly related with queries in
logic-based abduction, namely 
\emph{relevancy}/\emph{irrelevancy}~\cite{selman-aaai90,gottlob-ese90,gottlob-jacm95}.

\begin{example}
  For the DL of~\cref{fig:01a:dl}, and the instance
  $(\mbf{v},c)=((0,0,1,2),1)$, from the list of explanations,
  $\mbb{A}=\{\{1,4\}\}$ (for the AXp's), and $\mbb{C}=\{\{1\},\{4\}\}$
  (for the CXp's), it is plain that features 2 and 3 and not included
  in any explanation, and 1 and 4 are included in some explanation.
\end{example}

The MHS duality between AXp's and CXp's yields the following result: 
\begin{proposition} \label{prop:fmpaxpcxp}
  Given an explanation problem $\fml{E}_L=(\fml{M},\mbf{v},c)$, with
  $\fml{M}=(\fml{F},\mbb{D},\mbb{F},\fml{K},\kappa)$, and some target
  feature $t\in\fml{F}$, $t$ is included in some AXp of $\fml{E}_L$
  iff  $t$ is included in some CXp of $\fml{E}_L$.
\end{proposition}
Hence, when devising algorithms for FMP, one can either study the
membership in some AXp or the membership in some CXp.

FMP has a simple QBF formulation:
\[
\exists(\fml{X}\subseteq\fml{F}).%
\left[({t}\in\fml{X})\land\axp(\fml{X})\land%
  \left(\forall(\fml{X}'\subsetneq\fml{X}).\neg\axp(\fml{X}')\right)
  \right]
\]
There are several optimizations that can be introduced to this basic
QBF formulation, but that is beyond the scope of this document. 
%
%\paragraph{Basic results.}
%~\\
More importantly, there are some known results about the complexity of
FMP. These can be briefly summarized as follows,

\begin{proposition}[\cite{hiims-kr21}]
  FMP for a DNF classifier is $\stwop$-hard.
\end{proposition}

Since a DNF classifier can be reduced to more expressive classifiers,
like RFs and other tree ensembles like BTs, but also NNs, then we have
the following result,

\begin{proposition}
  FMP is $\stwop$-hard for RFs, BTs and NNs.
\end{proposition}

One important recent result has been the proof of membership in
$\stwop$. As a result, one solution approach for FMP is the use of
QBF/2QBF solvers~\cite{sat-handbook21}\footnote{%
There have been observable improvements in the performance of QBF
solvers in recent years, which can largely be attributed to the use of 
abstraction refinement
methods~\cite{jms-sat11,jms-ijcai15,rabe-fmcad15,jkmsc-aij16,rabe-sat16}.}.

Despite the complexity of FMP in general settings, there are families
of classifiers for which deciding FMP is in P~\cite{hiims-kr21}.
An immediate consequence of the fact that CXp's can be enumerated in
polynomial time for DTs is:

\begin{proposition}[\cite{hiims-kr21}]
  FMP for a DT is in P.
\end{proposition}

More recently, additional results on FMP have been proved,

\begin{proposition}[\cite{hms-corr22}]
  For a classifier for which it is in P to decide whether a set of
  features is a WAXp, then deciding FMP is in NP.
\end{proposition}

\begin{Proof}[Sketch]
  To prove that FMP is in NP in this case, one proceeds are follows.
  First, one guesses (non-deterministically) a set $\fml{X}$
  containing the target feature $t$.
  By hypothesis, this set is decided to be a weak AXp in poly-time,
  Next, we show that removing any feature causes the resulting set to
  no longer represent a weak AXp. Once again, by hypothesis there
  exists a polynomial time algorithm for deciding whether such a
  reduced set is a WAXp. Thus, deciding FMP is in P.
\end{Proof}

\jnoteF{DNF classifier: FMP is $\stwop$.}

\jnoteF{DTs: FMP is in P.}

\begin{comment}  %%% DECIDE WHETHER TO ADD
%
%
%
\paragraph{General-purpose FMP algorithm.}
%~\\
Since FMP is $\stwop$-complete for a wide range of classifiers, a
natural question is how to solve the problem in practice.
%
One solution is to represent FMP as a QBF formula (in fact a 2QBF
formula), and decide FMP with an existing QBF reasoner.
%
An alternative approach is to devise a dedicated algorithm...

\jnote{CEGAR approach to solve FMP. Am I sure this should be added?}
%
%
%
\end{comment}

%\paragraph{Open questions.}
%~\\
%
%\jnote{Move to last section.}

\subsection{Additional Explainability Queries}

A wealth of additional explainability queries have been studied in
recent years~\cite{marquis-kr20,marquis-kr21,darwiche-jair21}.
Examples include finding mandatory and/or forbidden features, 
counting and/or enumerating instances, among others.
Queries can be broadly categorized as class queries or explanation
queries~\cite{marquis-kr20,marquis-kr21}.
Examples of class queries include mandatory/forbidden features for a
class and necessary features for a class. Examples of explanation
queries include finding smallest AXp's, finding one AXp and finding
one CXp. Some of these queries have been studied earlier in this
document as well.
Complexity-wise, ~\cite{marquis-kr21} proves the NP-hardness of these
queries for the families of classifiers DLs, RFs, BTs, boolean NNs,
and binarized NNs (BNNs). In contrast, and also as shown in this
paper, for DTs, most queries can be answered in polynomial time.
%
%One important observation of this work is that for most families of
%classifiers, most queries are computationally hard to answer, with the
%exception being DTs.
%
It should be noted that some of the queries studied in recent work can
also be related with queries in logic-based
abduction~\cite{gottlob-ese90,selman-aaai90,gottlob-jacm95},
concretely relevancy/irrelevancy but also necessity.
More recent work on feature relevancy in explanations includes
dedicated algorithms for arbitrary classifiers~\cite{hims-aaai23}, and
NP-hardness proofs for some families of
classifiers~\cite{hcmpms-tacas23}.

%%\ifthenelse{\boolean{shownotes}}{
%%  \jnote{Add more detail, e.g.\ from the slides.}
%%}{}

%\subsection{Research Directions}

\jnoteF{Topics: counting, enumeration}

\begin{figure}[t]
  \begin{center}
    %\scalebox{0.925}{
    % Example from Tanner et al. PLoS'08 paper
%%
%
\forestset{
  BDT/.style={
    for tree={
      %l=1.5cm,s sep=1.15cm,
      l=1.5cm,s sep=1.0cm,
      if n children=0{}{circle}, %rectangle
      %if n children=0{}{draw},
      draw=black,%draw=midblue,
      text=black,%text=midblue,
      edge={-{Stealth[]}},
      %edge={
      %  my edge
      %},
      %if n=1{}{
      %  edge+={0 my edge},
      %},
      %edge=thick,
      %font=\sffamily,
    }
  },
}
\begin{forest}
  BDT
  [{$A$}, label={[yshift=-6.75ex]{{\xtiny1}}} %middle-middle=x
    [{$P$}, label={[yshift=-6.75ex]{{\xtiny2}}}, %top-left=x
      %%edge=thick,
      edge label={node[midway,left,xshift=-0.5pt] {{\xscriptsize$=0$}}}
          [\rhlight{\tbf{Y}}, label={[yshift=-5.375ex]{{\xtiny4}}},
            %%edge=thick,
            edge label={node[midway,left,xshift=-0.5pt] {{\xscriptsize$=1$}}},
            rectangle, fill={tred3!20}
          ]
          [\dghlight{\tbf{N}}, label={[yshift=-5.375ex]{{\xtiny5}}},
            %%edge=thick,
            edge label={node[midway,right,xshift=0.5pt] {{\xscriptsize$=0$}}},
            rectangle, fill={tgreen3!25}
          ]
    ]
    [{$P$}, label={[yshift=-6.75ex]{{\xtiny3}}}, %top-left=x
      %%edge=thick,
      edge label={node[midway,right,xshift=0.5pt] {{\xscriptsize$=1$}}}
      [{$N$}, label={[yshift=-6.885ex]{{\xtiny6}}}, %top-left=x
        %%edge=thick,
        edge label={node[midway,left,xshift=-2.25pt] {{\xscriptsize$=0$}}}
        [{$V$}, label={[yshift=-6.75ex]{{\xtiny8}}}, %top-left=x
          %%edge=thick,
          edge label={node[midway,left,xshift=-1.75pt] {{\xscriptsize$=0$}}}
          [{$Z$}, label={[xshift=-3.35ex,yshift=-3.5ex]{{\xtiny10}}}, %top-left=x
            %%edge=thick,
            edge label={node[midway,left,xshift=-2.0pt] {{\xscriptsize$=1$}}}
            [\dghlight{\tbf{N}}, label={[yshift=-5.375ex]{{\xtiny12}}},
              %%edge=thick,
              edge label={node[midway,left,xshift=-0.5pt] {{\xscriptsize$=1$}}},
              rectangle, fill={tgreen3!25}
            ]
            [{$S$}, label={[yshift=-6.75ex]{{\xtiny13}}}, %top-left=x
              %%edge=thick,
              edge label={node[near end,right,xshift=0.5pt] {{\xscriptsize$=2$}}}
              [{$G$}, label={[yshift=-6.75ex]{{\xtiny15}}}, %top-left=x
                %%edge=thick,
                edge label={node[midway,left,xshift=-2.5pt] {{\xscriptsize$=1$}}}
                [\dghlight{\tbf{N}}, label={[yshift=-5.375ex]{{\xtiny17}}},
                  %%edge=thick,
                  edge label={node[midway,left,xshift=0.5pt] {{\xscriptsize$=0$}}},
                  rectangle, fill={tgreen3!25}
                ]
                [\rhlight{\tbf{Y}}, label={[yshift=-5.375ex]{{\xtiny18}}},
                  %%edge=thick,
                  edge label={node[midway,right,xshift=0.5pt] {{\xscriptsize$=1$}}},
                  rectangle, fill={tred3!20}
                ]
              ]
              [{$H$}, label={[yshift=-6.885ex]{{\xtiny16}}}, %top-left=x
                %%edge=thick,
                edge label={node[midway,right,xshift=0.5pt] {{\xscriptsize$=0$}}}
                [\dghlight{\tbf{N}}, label={[yshift=-5.375ex]{{\xtiny19}}},
                  %%edge=thick,
                  edge label={node[midway,left,xshift=0.5pt] {{\xscriptsize$=1$}}},
                  rectangle, fill={tgreen3!25}
                ]
                [{$C$}, label={[yshift=-6.75ex]{{\xtiny20}}}, %top-left=x
                  %%edge=thick,
                  edge label={node[midway,right,xshift=0.5pt] {{\xscriptsize$=0$}}}
                  [\rhlight{\tbf{Y}}, label={[yshift=-5.375ex]{{\xtiny21}}},
                    %%edge=thick,
                    edge label={node[midway,left,xshift=-0.5pt] {{\xscriptsize$=1$}}},
                    rectangle, fill={tred3!20}
                  ]
                  [{$G$}, label={[yshift=-6.75ex]{{\xtiny22}}}, %top-left=x
                    %%edge=thick,
                    edge label={node[midway,right,xshift=0.5pt] {{\xscriptsize$=0$}}}
                    [\rhlight{\tbf{Y}}, label={[yshift=-5.375ex]{{\xtiny23}}},
                      %%edge=thick,
                      edge label={node[midway,left,xshift=-0.5pt] {{\xscriptsize$=1$}}},
                      rectangle, fill={tred3!20}
                    ]
                    [\dghlight{\tbf{N}}, label={[yshift=-5.375ex]{{\xtiny24}}},
                      %%edge=thick,
                      edge label={node[midway,right,xshift=0.5pt] {{\xscriptsize$=0$}}},
                      rectangle, fill={tgreen3!25}
                    ]
                  ]
                ]
              ]
            ]
            [\rhlight{\tbf{Y}}, label={[yshift=-5.375ex]{{\xtiny14}}},
              %%edge=thick,
              edge label={node[midway,right,xshift=0.5pt] {{\xscriptsize$=0$}}},
              rectangle, fill={tred3!20}
            ]
          ]
          [\rhlight{\tbf{Y}}, label={[yshift=-5.375ex]{{\xtiny11}}},
            %%edge=thick,
            edge label={node[midway,right,xshift=0.5pt] {{\xscriptsize$=0$}}},
            rectangle, fill={tred3!20}
          ]
        ]
        [\rhlight{\tbf{Y}}, label={[yshift=-5.375ex]{{\xtiny9}}},
          %%edge=thick,
          edge label={node[midway,right,xshift=0.5pt] {{\xscriptsize$=1$}}},
          rectangle, fill={tred3!20}
        ]
      ]
      [\rhlight{\tbf{Y}}, label={[yshift=-5.375ex]{{\xtiny7}}},
        %%edge=thick,
        edge label={node[midway,right,xshift=0.5pt] {{\xscriptsize$=1$}}},
        rectangle, fill={tred3!20}
      ]
    ]
  ]
\end{forest}
    %}
  \end{center}
  \caption{Decision tree adapted
    from~\cite[Fig.~9]{belmonte-ieee-access20}} \label{fig:dt:issue}
\end{figure}

\paragraph{Validation of ML models.}
%~\\
Recent work~\cite{bmpms-unpub23} illustrates the use of formal
explanations for identifying apparent flaws in ML models.
For example, the DT shown in~\cref{fig:dt:issue} has been proposed in
the field of medical diagnosis~\cite{belmonte-ieee-access20}, aiming
at providing a solution for non-invasive diagnosis of Meningococcal
Disease (MD) meningitis. (The actual feature names, and their domains,
are shown in~\cref{tab:dt:issue}.)
Unfortunately, the DT has a number of issues, in that it allows MD
meningitis to be diagnosed for patients that exhibit no symptoms 
whatsoever. The use of formal explanations, namely AXp's, allows
demonstrating these issues.

\begin{table}[t]
  \begin{center}
    \renewcommand{\arraystretch}{1.0125}
    \begin{tabular}{cccccc} \toprule
      Feat.~\# & Name & Meaning & Definition & Domain & Trait/Symp. \\ \toprule 
      1 & $A$ & Age & $\tn{Age}>5$? & $\{0,1\}$ & T \\ \midrule
      2 & $P$ & Petechiae & Petechiae? & $\{0,1\}$ & S \\ \midrule 
      3 & $N$ & Stiff Neck & Stiff Neck? & $\{0,1\}$ & S \\ \midrule 
      4 & $V$ & Vomiting & Vomiting? & $\{0,1\}$ & S \\ \midrule 
      %5 & $Z$ & Local Zone & Local Zone & $\{0,1,2\}$ & T\\ \midrule 
      5 & $Z$ & Zone & Zone${=}$? & $\{0,1,2\}$ & T\\ \midrule 
      6 & $S$ & Seizures & Seizures? & $\{0,1\}$ & S \\ \midrule
      7 & $G$ & Gender & Gender? & $\{0\,\tn{(F)},1\,\tn{(M)}\}$ & T \\ \midrule
      8 & $H$ & Headache & Headache? & $\{0,1\}$ & S \\ \midrule
      9 & $C$ & Coma & Coma? & $\{0,1\}$ & S \\
      \bottomrule
    \end{tabular}
  \end{center}
  \caption{Features and domains
    from~\cite{belmonte-ieee-access20}} \label{tab:dt:issue}
\end{table}

As one concrete example, the computation of AXp's allows concluding
that MD meningitis will be predicted whenever a patient has more than
5 years of age and lives in a rural area, i.e.\ without exhibiting any 
symptoms of the disease, at least among those tested for.
To prove that this is the case, one considers the path
$\langle1,3,6,8,10,14\rangle$, and confirms that there is an
explanation that does not include any of the symptoms
(i.e.\ Petechiae, Stiff neck, and Vomiting).
Here, the query is to assess the existence of explanations for which
the symptoms need not be tested for. The conclusion is that one
can diagnose MD meningitis without testing any of the symptoms of
meningitis.

As the previous example illustrates, reasoning about formal
explanations, including different kinds of queries, can serve to to
help decision makers in assessing whether an ML model offers
sufficient guarantees of quality to be deployed.
The previous example also illustrates the fundamental importance of
formal verification of ML models~\cite{msi-aaai22,seshia-cacm22} in
reduce the likelihood of accidents~\cite{bianchi-jair23}.

\section{Probabilistic Explanations} \label{sec:pxps}

%\begin{enumerate}
%\item General case
%\item Decision trees
%\item Naive Bayes classifiers
%\item Topics of research: other families of classifiers?
%\end{enumerate}

The cognitive limits of human beings are
well-known~\cite{miller-pr56}.
Unfortunately, it is also the case that formal explanations are often
larger than such cognitive limits.
One possible solution is to compute explanations that are not as
rigorous as AXp's, but which offer strong probabilistic guarantees of
rigor.
We refer to these explanations as \emph{probabilistic explanations}.
There is recent initial work on the complexity of computing
probabilistic explanations~\cite{kutyniok-jair21,waldchen-phd22},
where the name \emph{probabilistic prime implicants} is used.
Following more recent
work~\cite{iincms-corr21,iincms-corr22,ims-corr22,ihincms-corr22}, we
will use the term(s) (weak) \emph{probabilistic abductive
explanations} ((W)PAXp's).
To simplify the section contents, features are assumed to be
categorical or ordinal, in which case the values are restricted to
being boolean or integer.

\subsection{Problem Formulation}

A probabilistic (weak) AXp generalizes the definition of weak AXp, by
allowing the prediction to change in some points of feature space,
where a weak AXp would require the prediction not to change, but such 
that those changes have small probability.
One is thus interested in sets $\fml{X}\subseteq\fml{F}$ such that,
\begin{equation} \label{eq:probxp}
  \prob_{\mbf{x}}(\kappa(\mbf{x})=c\,|\,\mbf{x}_{\fml{X}}=\mbf{v}_{\fml{X}})
  \ge \delta
\end{equation}
where $0<\delta\le1$ is some given threshold, and
$\mbf{x}_{\fml{X}}=\mbf{v}_{\fml{X}}$ holds for any point in feature
space for which $\wedge_{i\in\fml{X}}(x_i=v_i)$.
Clearly, for $\delta=1$, \eqref{eq:probxp} corresponds to stating
that,
\begin{equation} \label{eq:xp}
  \forall(\mbf{x}\in\mbb{F}).%
  \left[\bigwedge\nolimits_{i\in\fml{X}}(x_i=v_i)\right]\limply(\kappa(\mbf{x})=c)
\end{equation}
Recent work~\cite{kutyniok-jair21,waldchen-phd22} established that,
for binary classifiers represented by boolean circuits, it is
$\tn{NP}^\tn{PP}$-complete to decide the existence of a set
$\fml{X}\subseteq\fml{F}$, with $|\fml{X}|\le{k}$, such that 
\eqref{eq:probxp} holds.
Despite this unwieldy complexity, it has been shown that for specific
families of
classifiers~\cite{iincms-corr22,ims-corr22,ihincms-corr22}, it is
computationally easier and practically efficient to compute 
(approximate) subset-minimal sets such that \eqref{eq:probxp} holds.
Concretely, instead of~\eqref{eq:axp1}, we will instead consider:
\begin{equation} \label{eq:wpaxp}
  \wpaxp(\fml{X};\mbb{F},\kappa,\mbf{v},c,\delta)
  %\:\, := %\:\:
  \quad := \quad
  % :=\,\:
  \prob_{\mbf{x}}(\kappa(\mbf{x})=c\,|\,\mbf{x}_{\fml{X}}=\mbf{v}_{\fml{X}})
  \ge \delta
\end{equation}
where $\wpaxp$ denotes a weak probabilistic AXp (PAXp).
Similarly to the deterministic case, a PAXp is a subset-minimal weak
AXp.
A set $\fml{X}\subseteq\fml{F}$ such that \eqref{eq:wpaxp} holds is
also referred to as \emph{relevant set}.
In the next section, we will illustrate how PAXp's are computed in the
case of DTs.

\subsection{Probabilistic Explanations for Decision Trees}
\label{sec:ppdt}

\paragraph{Path Probabilities for DTs.}
Next, we investigate how to compute, in the case of DTs, the
conditional probability,
%
%\begin{equation} \label{eq:cprob}
%  \prob_{\mbf{x}}(\kappa(\mbf{x})=c\,|\,\mbf{x}_{\fml{X}}\in\mbb{E}_{\fml{X}})\ge\delta
%\end{equation}
%
\begin{equation} \label{eq:cprob}
  \prob_{\mbf{x}}(\kappa(\mbf{x})=c\,|\,\mbf{x}_{\fml{X}}=\mbf{v}_{\fml{X}})
\end{equation}
where $\fml{X}$ is a set of \emph{fixed} features (whereas the other
features are not fixed, being deemed \emph{universal}), and $P_t$ is a
path in the DT consistent with the instance $(\mbf{v},c)$. (Also, note
that \eqref{eq:cprob} is the left-hand side of the definition of
$\wpaxp$ in~\eqref{eq:wpaxp} above.)
%%~\eqref{eq:drsdt}.
%
To motivate the proposed approach, let us first analyze how we can
compute $\prob_{\mbf{x}}(\kappa(\mbf{x})=c)$, where 
$\fml{P}\subseteq\fml{R}$ is the set of paths in the DT with
prediction $c$.
Let $\mrm{\Lambda}(R_k)$ denote the set of literals (each of the form
$x_i\in\mbb{E}_i$) in some path $R_k\in\fml{R}$. If a feature $i$ is
tested multiple times along path $R_k$, then $\mbb{E}_i$ is the
intersection of the sets in each of the literals of $R_k$ on $i$.
The number of values of $\mbb{D}_i$ consistent with literal
$x_i\in\mbb{E}_i$ is $|\mbb{E}_i|$.
Finally, the features \emph{not} tested along $R_k$ are denoted by
$\mrm{\Psi}(R_k)$.
For path $R_k$, the probability that a randomly chosen point in
feature space is consistent with $R_k$ (i.e.\ the \emph{path
  probability} of $R_k$) is given by,
\begin{equation} \label{eq:probrk}
\prob(R_k) =
\nicefrac{\left[\prod_{(x_i\in\mbb{E}_i)\in\mrm{\Lambda}(R_k)}|\mbb{E}_i|
    \times\prod_{i\in\mrm{\Psi}(R_k)}|\mbb{D}_i|\right]}{|\mbb{F}|}
\end{equation}
%
%%(Observe that, for instance-based explanations, each $|\mbb{E}_i|$
%%would be replaced by 1.)
%
As a result, we get that,
\begin{equation} \label{eq:probpred}
\prob_{\mbf{x}}(\kappa(\mbf{x})=c)={\textstyle\sum\nolimits}_{R_k\in\fml{P}}\prob(R_k)
\end{equation}

Given an instance $(\mbf{v},c)$ and a set of fixed features $\fml{X}$
(and so a set of universal features $\fml{F}\setminus\fml{X}$), we now
detail how to compute~\eqref{eq:cprob}.
Since some features will now be declared universal, multiple paths
with possibly different conditions can become consistent.
%For example, in~\cref{fig:runex01:dt} if feature 1 and 2 are declared
%universal, then (at least) paths $P_1$, $P_2$ and $Q_1$ are
%consistent with some of the possible assignments.
%
Although universal variables might seem to complicate the computation
of the conditional probability, this is not the case.

A key observation is that the feature values that make a path
consistent are disjoint from the values that make other paths
consistent. This observation allows us to compute the models 
consistent with each path and, as a result, to
compute~\eqref{eq:wpaxp}.
Let $R_k\in\fml{R}$ represent some path in the decision tree. (Recall
that $P_t\in\fml{P}$ is the target path, which is consistent with
$\mbf{v}$.)
Let $n_{ik}$ represent the (integer) number of assignments to feature
$i$ that are consistent with path $R_k\in\fml{R}$, given
$\mbf{v}\in\mbb{F}$ and $\fml{X}\subseteq\fml{F}$.
For a feature $i$, let $\mbb{E}_{i}$ denote the set of domain values of
feature $i$ that is consistent with path $R_k$. Hence, for path $R_k$,
we consider a literal $(x_i\in{\mbb{E}_{i}})$.
Given the above, the value of $n_{ik}$ is defined as follows:
\begin{enumerate}[nosep]
\item If $i$ is fixed:
  \begin{enumerate}[nosep]
  \item If $i$ is tested along $R_k$ and the value of $x_i$ is
    inconsistent with $\mbf{v}$, i.e.\ there exists a literal
    $(x_i\in\mbb{E}_i)\in\mrm{\Lambda}(R_k)$ and
    $\{v_i\}\cap{\mbb{E}_i}=\emptyset$, then $n_{ik}=0$;
  \item If $i$ is tested along $R_k$ and the value of $x_i$ is
    consistent with $R_k$, i.e.\ there exists a literal
    $(x_i\in\mbb{E}_i)\in\mrm{\Lambda}(R_k)$ and
    $\{v_i\}\cap{\mbb{E}_i}\not=\emptyset$, then $n_{ik}=1$;
  \item If $i$ is not tested along $R_k$, then $n_{ik}=1$.
  \end{enumerate}
\item Otherwise, $i$ is universal:
  \begin{enumerate}[nosep]
  \item If $i$ is tested along $R_k$, with some literal
    $x_i\in\mbb{E}_i$, then $n_{ik}=|\mbb{E}_i|$;
  \item If $i$ is not tested along $R_k$, then $n_{ik}=|\mbb{D}_i|$.
  \end{enumerate}
\end{enumerate}
Using the definition of $n_{ik}$, we can then compute the number of 
assignments consistent with $R_k$ as follows:
\begin{equation}
  \#(R_k;\mbf{v},\fml{X})={\textstyle\prod\nolimits}_{i\in\fml{F}}n_{ik}
\end{equation}
Finally,~\eqref{eq:cprob} is given by,
%
%\begin{align} \label{eq:cprob2}
%\prob_{\mbf{x}}(\kappa(\mbf{x})=c\,|&\,\mbf{x}_{\fml{X}}=\mbf{v}_{\fml{X}})=
%%\prob_{\mbf{x}}(\kappa(\mbf{x})=c\,|&\,\mbf{x}_{\fml{X}}\in\mbf{E}_{\fml{X}})=
%\nonumber\\
%&\nicefrac%
%    {\sum_{P_k\in\fml{P}}\#(P_k;\fml{F}\setminus\fml{X},\mbf{v})}
%    {\sum_{R_k\in\fml{R}}\#(R_k;\fml{F}\setminus\fml{X},\mbf{v})}
%\end{align}
%%
\begin{equation} \label{eq:cprob2}
\prob_{\mbf{x}}(\kappa(\mbf{x})=c\,|\,\mbf{x}_{\fml{X}}=\mbf{v}_{\fml{X}})\,=\,
\nicefrac%
    {\sum_{P_k\in\fml{P}}\#(P_k;\mbf{v},\fml{X})} %\fml{F}\setminus\fml{X},
    {\sum_{R_k\in\fml{R}}\#(R_k;\mbf{v},\fml{X})} %\fml{F}\setminus\fml{X},
\end{equation}
As can be concluded, and in the case of a decision tree,
both
$\prob_{\mbf{x}}(\kappa(\mbf{x})=c\,|\,\mbf{x}_{\fml{X}}=\mbf{v}_{\fml{X}})$
and $\wpaxp(\fml{X};\mbb{F},\kappa,\mbf{v},c,\delta)$ are computed in
polynomial time on the size of the DT.

\begin{example}
  For the DT of~\cref{fig:02a:dt}, with instance
  $(\mbf{v},c)=((0,0,1,0,1),1)$, we know that an AXp is $\{3,5\}$. Let
  $\delta=0.85$, and let us assess whether $\{5\}$ represents a weak
  probabilistic explanation.
  \cref{tab:02a:wpaxp} shows the path counts given $\fml{X}=\{5\}$ and
  $\mbf{v}=(0,0,1,0,1)$.
  From the table, we get that,
  \[
  \prob_{\mbf{x}}(\kappa(\mbf{x})=c\,|\,\mbf{x}_{\fml{X}}=\mbf{v}_{\fml{X}})\,=\,
  \nicefrac%
      {\sum_{P_k\in\fml{P}}\#(P_k;\mbf{v},\fml{X})} %\fml{F}\setminus\fml{X},
      {\sum_{R_k\in\fml{R}}\#(R_k;\mbf{v},\fml{X})}
      %\fml{F}\setminus\fml{X},
      \,=\, \nicefrac{14}{16} \,=\, 0.875
      \]
      And so, $\{5\}$ is a weak PAXp.
\end{example}

\begin{table}[t]
  \begin{center}
    \renewcommand{\tabcolsep}{0.5em}
    \renewcommand{\arraystretch}{1.075}
    \begin{tabular}{C{2.25cm}C{3cm}C{2.25cm}C{2.25cm}} \toprule
      Path ($R_k$) & Nodes of $R_k$ & $\#(R_k;\mbf{v},\fml{X})$ & Obs \\
      \toprule
      $P_1$ & $\langle1,2,4,7,10,15\rangle$ & 1 \\
      $P_2$ & $\langle1,2,4,7,11\rangle$    & 1 \\
      $P_3$ & $\langle1,2,5,8,13\rangle$    & 2 \\
      $P_4$ & $\langle1,2,5,9\rangle$       & 2 \\
      $P_5$ & $\langle1,3\rangle$           & 8 \\
      \midrule
            &                               & 14 & Total for $\fml{P}$ \\
      \midrule
      $Q_1$ & $\langle1,2,4,6\rangle$       & 2 \\
      $Q_2$ & $\langle1,2,4,7,10,14\rangle$ & 0 \\
      $Q_3$ & $\langle1,2,5,8,12\rangle$    & 0 \\
      \midrule
            &                               & 2 & Total for $\fml{Q}$ \\
      \bottomrule
    \end{tabular}
  \end{center}
  \caption{Path probabilities for DT of~\cref{fig:02a:dt}}
  \label{tab:02a:wpaxp}
\end{table}

%%\ifthenelse{\boolean{shownotes}}{
%%  \jnote{Add more detail, and example.}
%%}{}

%\subsection{Other Classifiers}
%\subsection{Additional Topics}
\subsection{Additional Results}

For classifiers represented as boolean circuits, the computation of
probabilistic abductive explanations is
$\tn{NP}^\tn{PP}$-hard~\cite{kutyniok-jair21,waldchen-phd22}.
Motivated by this complexity, expected to be beyond the reach of
modern reasoners, recent efforts studied specific families of
classifiers. In the case of DTs, the approach summarized in the
previous section was proposed
elsewhere~\cite{iincms-corr21,iincms-corr22,ihincms-corr22}, and shown
to be effective in practice. Also in the case of DTs, computational
hardness results have been proved in more recent
work~\cite{barcelo-corr22,barcelo-nips22}.
Furthermore, an approach based on dynamic programming was used for
computing probabilistic explanations in the case of
NBCs~\cite{ims-corr22,ihincms-corr22}.

%\subsection{Boolean Circuits}

%\subsection{Decision Trees}

%\subsection{Naive Bayes Classifiers}

%\subsection{Research Directions}

\jnoteF{Topics: other families of classifiers?}

\section{Input Constraints \& Distributions} \label{sec:ixps}

A critical assumption implicit on most work on formal explainability
is that all inputs are possible (and equally likely).
Unfortunately, this is often not the case. For example, consider a
classification problem with features 'order', denoting the animal
order, and 'winged', denoting whether the animal has wings. It might
be expected that points in feature space having 'order=Proboscidea'
(i.e.\ that includes elephants) and 'winged=true' would be
disallowed. In contrast, 'order=Chiroptera' (i.e.\ that includes
bats) and 'winged=true' would be allowed.
%
%'gender' and 'pregnant', it might be expected that points in feature
%space having 'gender=male' and 'pregnant=yes' would be disallowed.
%
%
If such constraints on the features are known a priori, then one can
take them into account when computing explanations. However, in most
cases, such constraints are unknown.
This section summarizes recent work on the general topic of handling
of input constraints.

%\subsection{Constraints on the Inputs}
\paragraph{Constraints on the features.}
Let us assume that a given classifier $\mbb{C}$ is characterized by a
constraint set $\fml{I}_{\mbb{C}}$ capturing the allowed points in
feature space. (In general, we view $\fml{I}_{\mbb{C}}$ as a
predicate, mapping points in feature space into $\{0,1\}$.)
The definition of weak AXp can be adapted to account for such
constraint set as follows,
\begin{equation} \label{eq:axp3}
  \forall(\mbf{x}\in\mbb{F}).%
  \left[\bigwedge\nolimits_{i\in\fml{X}}(x_i=v_i)\right]\limply\left[\fml{I}_{\mbb{C}}(\mbf{x})\limply(\kappa(\mbf{x})=c)\right] %\nolimits
\end{equation}
Similarly, the definition of weak CXp can be adapted to account for
$\fml{I}_{\mbb{C}}$,
\begin{equation} \label{eq:cxp3}
  \exists(\mbf{x}\in\mbb{F}).%
  \left[\bigwedge\nolimits_{i\not\in\fml{Y}}(x_i=v_i)\right]\land
  \left[\fml{I}_{\mbb{C}}(\mbf{x})\land(\kappa(\mbf{x})\not=c)\right] %\nolimits
\end{equation}
A number of observations can be made:
\begin{enumerate}
\item The definitions of AXp and CXp, given the definitions of input
  constraint aware weak AXp/CXp's, remain unchanged.
\item Duality between AXp's and CXp's (see~\cref{ssec:xpdual}) still
  holds.
\item Depending on the family of classifiers and the representation of
  the constraints, the complexity of computing one explanation need
  not change. For example, if the constraints are represented as
  propositional Horn clauses (and this is the case with propositional
  rules), then the propositional encoding for computing explanations
  of DTs will still enable computing explanations in polynomial time.
\item Finally, the same approach can also be used with probabilistic
  explanations.
\end{enumerate}
Given the above, and as long as the allowed points in feature space
are represented by a constraint set, then we can take those
constraints into account when computing AXp's and CXp's.
The original ideas on accounting for input constraints were presented
in recent work~\cite{rubin-aaai22}, and extended more recently for
contrastive explanations~\cite{yisnms-corr22}.
However, a major difficulty with the handling input constraints is how
to infer those input constraints in the first place. A possible
solution to this challenge has been proposed in recent
work~\cite{yisnms-corr22}.

\paragraph{Inferring constraints.}
When given a dataset and an ML classifier, one can exploit standard ML
learning approaches for inferring constraints that are consistent
with training data.
Recent work studied the learning of rules on the features given the
training data~\cite{yisnms-corr22}.
The experimental results substantiate the importance of inferring
constraints on the features, that also lead to smaller abductive
explanations and \emph{larger} contrastive explanations.

\paragraph{Research directions.}
Inferring \emph{good} constraints from training data is a promising
direction of research. The goal will be to find the best possible
rules, that improve the accuracy of explanations, but that do not
impact significantly the performance of formal explainers.
Another line of research is to account input distributions when these
are either known or can be inferred.

\jnoteF{From constraints to distributions.}

%\subsection{Problem Formulation}
%
%\subsection{Research Topics}

%\section{Model Approximations -- Surrogate Models} \label{sec:axps}
\section{Formal Explanations with Surrogate Models} \label{sec:axps}

%\begin{enumerate}
%\item Approximations with RFs
%\item Approximations with DTs??
%\end{enumerate}

As briefly discussed in~\cref{ssec:nfxp}, the most visible
(non-formal) explainability approaches consist of approximating a
complex classifier with a much simpler classifier (e.g.\ a linear
classifier or a decision tree) which, due to its simplicity, is
interpretable and so represents an explanation for the complex
classifier~\cite{guestrin-kdd16,lundberg-nips17}.

Despite the numerous shortcomings of such line of research
(see~\cref{ssec:nfxp}), it is also the case that approximating complex
ML models \emph{locally} with much simpler (or surrogate) ML models,
has been studied in several other
works~\cite{hinton-cexaiia17,bastani-corr17a,bastani-corr17b}.
Furthermore, there has been work on finding surrogate models, that
locally approximate a complex ML model, such that computing formal
explanations for the surrogate model is efficient in
practice~\cite{mazure-cikm21}.

Let $\mbb{C}$ represent a complex ML model, e.g.\ a neural network,
and let $(\mbf{v},c)$ represent a target instance. Moreover, let
$\mbb{A}$ represent an approximating (surrogate) ML model, e.g.\ a
random forest, which \emph{approximates} $\mbb{C}$ in points of
feature space that are sufficiently close to $\mbf{v}$, i.e.\ for 
point $\mbf{x}\in\mbb{F}$, with $||\mbf{x}-\mbf{v}||\le\epsilon$, for
some small $\epsilon>0$, it is the case that
$\kappa_{\mbb{C}}(\mbf{x})$ and $\kappa_{\mbb{A}}(\mbf{x})$ coincide
with high probability.
The conjecture out forward in recent work~\cite{mazure-cikm21} is that
a (rigorous) explanation (either AXp or CXp) of the instance
$(\mbf{v},c)$ computed for $\mbb{A}$ is also a sufficiently accurate
explanation for $\mbb{C}$ on the same instance.
At present, this novel line of research requires further validation.
For example, past work has not shown in practice that formal
explanations computed for the surrogate model are sufficiently
accurate for the complex model.
Although past work considered random forests as the surrogate model,
it is plain to conclude that other surrogate models can be considered,
e.g.\ decision trees or NBCs. The reason for considering simpler ML
models is that probabilistic explanations can be computed efficiently,
and this is not the case with random forests.

%\subsection{Problem Formulation}
%
%\subsection{Approximations with Random Forests}
%
%\subsection{Approximations with Decision Trees}
%
%\subsection{Research Topics}

\section{Additional Topics \& Extensions} \label{sec:exts}

%\begin{enumerate}
%\item Explanation literals
%\item Feature aggregation
%\item ...
%\end{enumerate}

%\subsection{Links with Fairness, Robustness, etc.}
\paragraph{Links with fairness, robustness, etc.}
Formal explainability has been related with robustness and
fairness. For global explanation problems, it is now known that
the minimal hitting sets of abductive explanations (which have been
referred to as counterexamples) contain one or more adversarial
examples~\cite{inms-nips19}.
Moreover, initial links with fairness were investigated in more recent
work~\cite{icshms-cp20}.
Finally, the relationship between model learning and explainability is
a topic of future research.

%\subsection{Explanation Literals}
\paragraph{Explanation literals.}
By definition, the definition of AXp and CXp assumes literals based on
equality. This is justified by the fact that AXp's and CXp's are
computed with respect to a concrete point in feature space. 
In some settings, it has been shown that literals based on equality
can be generalized to the literals that occur in the model itself.
This is the case with decision trees~\cite{iims-jair22}, where
literals can be defined using the set membership operator, and so
explanations can be related with such literals.
Similar ideas are yet to be investigated in the case of other ML
models.

%\subsection{Localized Explanations}
\paragraph{Localized explanations.}

Non-formal explainability methods emphasize the \emph{local} nature of
their explanations. In situations where such locality if of interest,
one may wonder whether formal explainability can be adapted to further
emphasize locality.
Given the diverse nature of features, we opt to define the Hamming
distance between two points in feature space,
\begin{equation}
  \hamd(\mbf{x}_1,\mbf{x}_2)=\sum\nolimits_{i=1}^m\tn{ITE}(x_{1i}=x_{2i},1,0)
\end{equation}

Given the definition of Hamming distance, we can now propose a
definition of \emph{localized} weak abductive and contrastive
explanations.
%%
%\begin{equation} \label{eq:laxp}
%  \begin{array}{rcl}
%    \wlaxp(\fml{X}) & ~:=~~ &
%    \forall(\mbf{x}\in\mbb{F}).%
%    \left[\hamd(\mbf{x},\mbf{v})\land\bigwedge_{i\in\fml{X}}(x_i=v_i)\right]\limply(\kappa(\mbf{x})=c)%\\ %\nolimits
%  \end{array}
%\end{equation}
%%
%%
%\begin{equation} \label{eq:lcxp}
%  \begin{array}{rcl}
%    \wlcxp(\fml{Y}) & ~:=~~ & \exists(\mbf{x}\in\mbb{F}).%
%    \left[\hamd(\mbf{x},\mbf{v})\land\bigwedge_{i\not\in\fml{Y}}(x_i=v_i)\right]\land(\kappa(\mbf{x})\not=c)\\ %\nolimits
%  \end{array}
%\end{equation}
%%
\begin{align} %%\label{eq:lxp}
  \wlaxp(\fml{X}) :=~ 
  \forall&(\mbf{x}\in\mbb{F}). \nonumber \\
  & \left[\hamd(\mbf{x},\mbf{v})\le\epsilon\land\bigwedge\nolimits_{i\in\fml{X}}(x_i=v_i)\right]\limply(\kappa(\mbf{x})=c)
  \label{eq:lxp1}
  \\[2.5pt] %\nolimits
  \wlcxp(\fml{Y}) :=~ \exists& (\mbf{x}\in\mbb{F}). \nonumber \\
  & \left[\hamd(\mbf{x},\mbf{v})\le\epsilon\land\bigwedge\nolimits_{i\not\in\fml{Y}}(x_i=v_i)\right]\land(\kappa(\mbf{x})\not=c)%\\ %\nolimits
  \label{eq:lxp2}
\end{align}
for some target $\epsilon>0$.
The definitions of subset-minimal sets remain unchanged,
i.e.\ localized AXp's and CXp's can be computed using~\eqref{eq:axp2b}
and~ \eqref{eq:cxp2b}, by replacing $\waxp$ and $\waxp$, respectively
by $\wlaxp$ and $\wlcxp$.
Finally, although we opted to use Hamming distance,~\eqref{eq:lxp1}
and~\eqref{eq:lxp2} could consider other measures of distance.

\jnoteF{How to control distance to point of feature space associated
  with instance?}

%\subsection{Explanations Beyond Machine Learning}
\paragraph{Explanations beyond ML.}
Although at present ML model explainability of ML models is the most
studied theme in the general field of explainability, it is also the
case that explainability has been studied in AI for
decades~\cite{swartout-ijcai77,swartout-aij83,shanahan-ijcai89,selman-aaai90,gottlob-ese90,gottlob-jacm95,simari-aij02,uzcategui-aij03,amgoud-aaai06,amgoud-aamasj08,amgoud-aij09,toni-ecai14},
with a renewed interest in recent years.
For example, explanations have recently been studied in AI planning~\cite{
  fox-corr17,%AI planning
  kambhampati-ijcai19,%AI planning
  hoffmann-rw19,%AI planning
  hoffmann-aaai20,%AI planning
  kambhampati-ijcai20,%AI planning
  kambhampati-aij21a,%AI planning
  magazzeni-jair21,%AI planning
  magazzeni-jair22,%AI planning
  kambhampati-iclr22,%AI planning, etc.
  hoffmann-ijcai22%,%AI planning
},
constraint satisfaction and problem solving~\cite{%
  guns-aij21,%constraint satisfaction
  osullivan-ijcai21,%constraint satisfaction
  guns-ijcai21,%constraint satisfaction
  gent-corr21,%puzzles (problem solving)
  kambhampati-iclr22%,%AI planning, etc.
},
among other examples~\cite{%
  kambhampati-aij21b%,%AI agents
}.
Furthermore, there is some agreement that regulations like EU's
General Data Protection Regulation (GDPR)~\cite{eu-gdpr16} effectively
impose the obligation of explanations for any sort of algorithmic
decision making~\cite{goodman-aimag17,routledge-beq22}.
Despite representing fairly distinct areas of research, it is the case
that most explainability approaches focus on computing explanations by
computing MUSes or variants thereof. This is the case in
planning~\cite{hoffmann-aaai20}, in constraint
solving~\cite{guns-aij21}, besides explanations in ML as detailed in
earlier sections of this paper (see~\cref{sec:cxps}).

%\ifthenelse{\boolean{shownotes}}{
%  \jnote{Highlight connections between different areas by relating
%    with MUS/MCS extraction/enumeration.}
%}{}

\jnoteF{Explanations beyond classification: regression
  AI planning, scheduling, optimization, constraint solving, etc.\\
  Algorithmic decision making in general.
}

\jnoteF{
  Cite planning references: Fox, Hoffmann, Kambhampati, etc. \\
  Cite Guns, Gent, etc. \\
  Cite O'Sullivan\\
  \cite{
  fox-corr17,%AI planning
  kambhampati-ijcai19,%AI planning
  hoffmann-rw19,%AI planning
  hoffmann-aaai20,%AI planning
  kambhampati-ijcai20,%AI planning
  guns-aij21,%constraint satisfaction
  kambhampati-aij21a,%AI planning
  kambhampati-aij21b,%AI agents
  magazzeni-jair21,%AI planning
  osullivan-ijcai21,%constraint satisfaction
  guns-ijcai21,%constraint satisfaction
  gent-corr21,%puzzles (problem solving)
  magazzeni-jair22,%AI planning
  kambhampati-iclr22,%AI planning, etc.
  hoffmann-ijcai22%,%AI planning
  }
  But also,\\
  \cite{goodman-aimag17}
}

\section{Future Research \& Conclusions} \label{sec:conc}

This section concludes the paper. As the previous sections illustrate,
formal explainability has blossomed into a number of important areas
of research. Thus, we start by overviewing a number of research
directions. Afterwards, we summarize the paper's contributions.

\subsection{Research Directions}
%~\\

As the second part of the paper reveals
(see~\crefrange{sec:qxps}{sec:exts}), there exist a vast number of
ongoing research topics in the field of formal explainability.

\paragraph{Definitions of explanations.}
%~\\
Although the existing definitions of (formal) explanations offer
important theoretical advantages, e.g.\ duality of explanations,
researchers have looked at alternative definitions, with the purpose
of improving the efficiency of algorithms for computing explanations,
or improving the expressiveness of
explanations~\cite{marquis-aaai22,marquis-ijcai22a}.

\paragraph{Computation of explanations.}
%~\\
The ability to devise more efficient tools to reason about NNs
represents a critical topic of research. Significant improvements in
the tools used to reason about NNs would allow explaining more complex
classifiers, and so extend the rage of applicability of formal
explainability.
The grand challenges in the computation of explanations is to devise
novel methods for efficiently computing explanations of neural
networks and bayesian network classifiers.
Recent progress in the analysis of
NNs~\cite{barrett-cav17,barrett-cav19,barrett-fto21} suggests initial
directions.
A related line of research is the computation of approximate
explanations with formal guarantees~\cite{katz-tacas23}.

\paragraph{Explainability queries.}
%~\\
Besides enumeration of explanations, a related question is the
enumeration of explanations that are preferred or that take user
suggestions into account. This is the subject of future research.
Regarding the feature membership, several research problems can be
envisioned. One is to efficiently decide membership in the case of
arbitrary classifiers, e.g.\ random forests and other tree ensembles.
Another direction of research is to chart the complexity of FMP for
the many families of classifiers that can be used in practical
settings. For example, given the result that FMP is in NP for families
of classifiers for which computing one explanation is in P, then
proving/disproving hardness results would allow selecting the most
adequate tools to use when solving FMP in practice.

\paragraph{Probabilistic explanations.}
%~\\
One key difficulty of computing probabilistic explanations is the
computational complexity of the problem~\cite{kutyniok-jair21}.
Although researchers have made progress in devising efficient
algorithms for efficiently computing probabilistic explanations for
specific families of
classifiers~\cite{iincms-corr22,ims-corr22}, but also in understanding
the computational hardness of computing probabilistic explanations in  
such cases~\cite{barcelo-corr22}, a number of topics of research can
be envisioned. Concretely, one topic is to devise a more complete
chart of the computational complexity of the problem, and a second
topic is to devise practically efficient algorithms for families of
classifiers that have not yet been investigated, e.g.\ decision lists
and sets and tree ensembles, among others.

%\subsection{Explanation Certification}
\paragraph{Explanation certification.}
It is well-known that algorithms proved correct can be implemented
incorrectly.
In areas where the rigor of results is paramount, there have been
efforts to devise mechanisms for ascertaining the correctness of
either implemented algorithms or their computed
results~\cite{weber-jal09,dfms-ictac10,mehlhorn-cav11,mehlhorn-csr11,mehlhorn-jar14,heule-cacm17,cfmssk-tacas17,heule-aaai18,cfmssk-jar19,nordstrom-aaai20,nordstrom-ijcai20,nordstrom-cp20,nordstrom-aaai21}.
%
%One well-known example is SAT solving, ...
%
A natural topic of research is to apply similar solutions in the case
of the computation of explanations, but also in the case of
explainability queries.
For example, existing algorithms for computing one explanation can be
formalized in a proof assistant (e.g.~~\cite{coq-bk04}), from which a
certified executable can then be extracted.
Explanation of certification is expected to be relevant in settings
that are deemed high-risk or safety-critical.

\jnoteF{Certified implementations; checking; etc.}

\jnoteF{Automated checking of explanations}

\paragraph{Additional topics.}
%~\\
%
% Input constraints
The accounting for input constraints can play a key role in formal
explainability. As a result, the inference of \emph{good} constraints
from  training data is a promising  direction of research. The goal
will be to find the best possible rules, that improve the accuracy of
explanations, but that do not impact significantly the performance of
formal explainers.
Another line of research is to account input distributions when these
are either known or can be inferred.
%
% Surrogate models.
As noted in~\cref{sec:axps}, the use of surrogate models to compute
explanations of complex models holds great promise, but it also
requires further assessment. It is open such assessment is to be
made.
%
% Feature aggregation.
%\paragraph{Feature aggregation.}
In classification problems with a large number of features, it is
often important to be able to aggregate features. In formal
explainability, this issue has not yet been addressed, and it is a
topic of future research.
The previous sections also mentioned in passing several topics of
research, that could contribute to raising the impact of formal
explainability.

\subsection{Concluding Remarks}
This paper summarizes the recent developments in the emerging field of
formal explainability. The paper overviews the definition of
explanations, and covers the computation of explanations, addressing
specific families of classifiers, both families for which computing
one explanation is computationally hard, and families for which
computing one explanation is tractable. The paper also covers a wide
range of ongoing topics of active research including explainability
queries, probabilistic explanations, accounting for input constraints,
and formal explainability using surrogate models. In most cases, the
paper also highlights existing topics of research.

As shown throughout the paper, formal explainability borrows
extensively from a number of areas of research in AI, including
automated reasoning and model-based diagnosis. Different reasoners,
including SAT, MILP, and SMT, among others, have been and continue to
be exploited in devising ways of computing explanations, both exact
and probabilistic, but also answering explainability queries.

%\clearpage
%\input{todo}

%\clearpage\input{plan}
%\clearpage\input{wip}
%
\subsubsection*{Acknowledgements.}
This document was motivated by the opportunity to give a short course
on formal XAI at the $18^{\tn{th}}$ Reasoning Web Summer
School\footnote{%
\url{https://2022.declarativeai.net/events/reasoning-web}.
},
organized by Leopoldo Bertossi and Guohui Xiao.
The work summarized in this document results in part from
collaborations and discussions with several colleagues, including
F.~Arenas,
N.~Asher,
R.~Béjar,
M.~Cooper,
B.~German,
T.~Gerspacher,
E.~Hebrard,
X.~Huang,
A.~Hurault,
A.~Ignatiev,
Y.~Izza,
O.~Létoffé,
X.~Liu,
E.~Lorini,
C.~Menc\'{\i}a,
A.~Morgado,
N.~Narodytska,
J.~Planes,
R.~Passos,
M.~Siala,
M.~Tavassoli,
%M.~Janota,
J.~Veron, %%but also
among others.
Some colleagues offered detailed comments on earlier drafts of this
document, namely 
Y.~Izza,
C.~Menc\'{\i}a,
A.~Morgado, and
J.~Planes.
%\\
%
This work was supported by the AI Interdisciplinary Institute ANITI,
funded by the French program ``Investing for the Future -- PIA3''
under Grant agreement no.\ ANR-19-PI3A-0004, and by the H2020-ICT38
project COALA ``Cognitive Assisted agile manufacturing for a Labor
force supported by trustworthy Artificial intelligence''.
Finally, I acknowledge 
% Alt: JMS
the incentive provided by the ERC who, by not funding this research
nor a handful of other grant applications between 2012 and 2022, has
had a lasting impact in framing the research presented in this paper.

%

% RequiredL: \usepackage{etoolbox}
%\providetoggle{mkbbl}
\newtoggle{mkbbl}
% Contents if using bibtex: "\settoggle{mkbbl}{true}"
% Contents if inputing pre-generated file: "\settoggle{mkbbl}{false}"

\settoggle{mkbbl}{false}
 % file is automatically generated

% ---- Bibliography ----
%%\cleardoublepage %% TENTATIVE, and required if bibliography starts page...
\addcontentsline{toc}{section}{References}
\vskip 0.2in
% For arxix paper production, and since arXiv does not allow for
% bibtex, we need to create a .bbl file to include upon submission
% to arXiv.
\iftoggle{mkbbl}{
  % Run bibtex, i.e. generate .bbl gile
  \bibliographystyle{splncs04}
  \bibliography{refs,dts,team,rai,xtra,gov}

\begin{thebibliography}{100}
\providecommand{\url}[1]{\texttt{#1}}
\providecommand{\urlprefix}{URL }
\providecommand{\doi}[1]{https://doi.org/#1}

\bibitem{berrada-ieee-access18}
Adadi, A., Berrada, M.: Peeking inside the black-box: {A} survey on explainable
  artificial intelligence {(XAI)}. {IEEE} Access  \textbf{6},  52138--52160
  (2018), \url{https://doi.org/10.1109/ACCESS.2018.2870052}

\bibitem{adebayo-nips18}
Adebayo, J., Gilmer, J., Muelly, M., Goodfellow, I.J., Hardt, M., Kim, B.:
  Sanity checks for saliency maps. In: NeurIPS. pp. 9525--9536 (2018)

\bibitem{newscientist13}
Adee, S.: Bad bugs: The worst disasters caused by software fails.
  \url{https://www.newscientist.com/gallery/software-bugs/} (2013)

\bibitem{schaus-aaai20}
Aglin, G., Nijssen, S., Schaus, P.: Learning optimal decision trees using
  caching branch-and-bound search. In: AAAI. pp. 3146--3153 (2020)

\bibitem{schaus-ijcai20b}
Aglin, G., Nijssen, S., Schaus, P.: {PyDL8.5}: a library for learning optimal
  decision trees. pp. 5222--5224 (2020)

\bibitem{aiidb23}
{AI Incident Database}: \url{https://incidentdatabase.ai/} (2023)

\bibitem{algwatch20}
{Algorithm Watch}: {AI} ethics guidelines global inventory.
  \url{https://inventory.algorithmwatch.org/} (2020)

\bibitem{mehlhorn-cav11}
Alkassar, E., B{\"{o}}hme, S., Mehlhorn, K., Rizkallah, C.: Verification of
  certifying computations. In: CAV. pp. 67--82 (2011)

\bibitem{mehlhorn-jar14}
Alkassar, E., B{\"{o}}hme, S., Mehlhorn, K., Rizkallah, C.: A framework for the
  verification of certifying computations. J. Autom. Reasoning  \textbf{52}(3),
   241--273 (2014), \url{https://doi.org/10.1007/s10817-013-9289-2}

\bibitem{ansotegui-corr21}
Alos, J., Ansotegui, C., Torres, E.: Learning optimal decision trees using
  {MaxSAT}. CoRR  \textbf{abs/2110.13854} (2021),
  \url{https://arxiv.org/abs/2110.13854}

\bibitem{amgoud-ecsqaru21}
Amgoud, L.: Non-monotonic explanation functions. In: ECSQARU. pp. 19--31 (2021)

\bibitem{amgoud-ijar23}
Amgoud, L.: Int. J. Approx. Reason.  (2023), in Press

\bibitem{amgoud-ijcai22}
Amgoud, L., Ben{-}Naim, J.: Axiomatic foundations of explainability. In: IJCAI.
  pp. 636--642 (2022)

\bibitem{amgoud-aaai06}
Amgoud, L., Prade, H.: Explaining qualitative decision under uncertainty by
  argumentation. In: AAAI. pp. 219--224 (2006)

\bibitem{amgoud-aij09}
Amgoud, L., Prade, H.: Using arguments for making and explaining decisions.
  Artif. Intell.  \textbf{173}(3-4),  413--436 (2009)

\bibitem{amgoud-aamasj08}
Amgoud, L., Serrurier, M.: Agents that argue and explain classifications.
  Auton. Agents Multi Agent Syst.  \textbf{16}(2),  187--209 (2008)

\bibitem{rudin-jmlr17a}
Angelino, E., Larus{-}Stone, N., Alabi, D., Seltzer, M., Rudin, C.: Learning
  certifiably optimal rule lists for categorical data. J. Mach. Learn. Res.
  \textbf{18},  234:1--234:78 (2017)

\bibitem{rudin-kdd17}
Angelino, E., Larus{-}Stone, N., Alabi, D., Seltzer, M.I., Rudin, C.: Learning
  certifiably optimal rule lists. In: KDD. pp. 35--44 (2017)

\bibitem{barcelo-nips21}
Arenas, M., Baez, D., Barcel{\'{o}}, P., P{\'{e}}rez, J., Subercaseaux, B.:
  Foundations of symbolic languages for model interpretability. In: NeurIPS.
  pp. 11690--11701 (2021)

\bibitem{barcelo-aaai21}
Arenas, M., Barcel{\'{o}}, P., Bertossi, L.E., Monet, M.: The tractability of
  {SHAP}-score-based explanations for classification over deterministic and
  decomposable boolean circuits. In: AAAI. pp. 6670--6678 (2021)

\bibitem{barcelo-nips22}
Arenas, M., Barcel{\'{o}}, P., Romero, M., Subercaseaux, B.: On computing
  probabilistic explanations for decision trees. In: NeurIPS (2022)

\bibitem{barcelo-corr22}
Arenas, M., Barcel{\'{o}}, P., Romero, M., Subercaseaux, B.: On computing
  probabilistic explanations for decision trees. CoRR  \textbf{abs/2207.12213}
  (2022), \url{https://doi.org/10.48550/arXiv.2207.12213}

\bibitem{arora-bk09}
Arora, S., Barak, B.: Computational Complexity - {A} Modern Approach. Cambridge
  University Press (2009),
  \url{http://www.cambridge.org/catalogue/catalogue.asp?isbn=9780521424264}

\bibitem{asher-cdmake21}
Asher, N., Paul, S., Russell, C.: Fair and adequate explanations. In: CD-MAKE.
  pp. 79--97 (2021)

\bibitem{marquis-kr21}
Audemard, G., Bellart, S., Bounia, L., Koriche, F., Lagniez, J., Marquis, P.:
  On the computational intelligibility of boolean classifiers. In: KR. pp.
  74--86 (2021)

\bibitem{marquis-ijcai22a}
Audemard, G., Bellart, S., Bounia, L., Koriche, F., Lagniez, J., Marquis, P.:
  On preferred abductive explanations for decision trees and random forests.
  In: IJCAI. pp. 643--650 (2022)

\bibitem{marquis-dke22}
Audemard, G., Bellart, S., Bounia, L., Koriche, F., Lagniez, J., Marquis, P.:
  On the explanatory power of boolean decision trees. Data Knowl. Eng.
  \textbf{142},  102088 (2022)

\bibitem{marquis-aaai22}
Audemard, G., Bellart, S., Bounia, L., Koriche, F., Lagniez, J., Marquis, P.:
  Trading complexity for sparsity in random forest explanations. In: AAAI. pp.
  5461--5469 (2022)

\bibitem{marquis-kr20}
Audemard, G., Koriche, F., Marquis, P.: On tractable {XAI} queries based on
  compiled representations. In: KR. pp. 838--849 (2020)

\bibitem{au-ai-plan21}
{Australian Gov.}: Australia's {AU} action plan. tiny.cc/hy8juz (2021),
  accessed: 2021-12-01

\bibitem{au-ai-ethics21}
{Australian Gov.}: Australia’s artificial intelligence ethics framework.
  \url{tiny.cc/ey8juz} (2021), accessed: 2021-12-01

\bibitem{avellaneda-corr19}
Avellaneda, F.: Learning optimal decision trees from large datasets. CoRR
  \textbf{abs/1904.06314} (2019), \url{http://arxiv.org/abs/1904.06314}

\bibitem{avellaneda-aaai20}
Avellaneda, F.: Efficient inference of optimal decision trees. In: AAAI. pp.
  3195--3202 (2020)

\bibitem{bacchus-cav15}
Bacchus, F., Katsirelos, G.: Using minimal correction sets to more efficiently
  compute minimal unsatisfiable sets. In: CAV. pp. 70--86 (2015)

\bibitem{bacchus-cpaior16}
Bacchus, F., Katsirelos, G.: Finding a collection of {MUSes} incrementally. In:
  CPAIOR. pp. 35--44 (2016)

\bibitem{muller-plosone15}
Bach, S., Binder, A., Montavon, G., Klauschen, F., M{\"u}ller, K.R., Samek, W.:
  On pixel-wise explanations for non-linear classifier decisions by layer-wise
  relevance propagation. PloS one  \textbf{10}(7),  e0130140 (2015)

\bibitem{stuckey-padl05}
Bailey, J., Stuckey, P.J.: Discovery of minimal unsatisfiable subsets of
  constraints using hitting set dualization. In: PADL. pp. 174--186 (2005)

\bibitem{barcelo-nips20}
Barcel{\'{o}}, P., Monet, M., P{\'{e}}rez, J., Subercaseaux, B.: Model
  interpretability through the lens of computational complexity. In: NeurIPS
  (2020)

\bibitem{zdnet07}
Barker, C.: The top 10 {IT} disasters of all time.
  \url{https://www.zdnet.com/article/the-top-10-it-disasters-of-all-time/}
  (2007)

\bibitem{katz-tacas23}
Bassan, S., Katz, G.: Towards formal approximated minimal explanations of
  neural networks. In: TACAS (2023)

\bibitem{bastani-corr17b}
Bastani, O., Kim, C., Bastani, H.: Interpretability via model extraction. CoRR
  \textbf{abs/1706.09773} (2017)

\bibitem{bastani-corr17a}
Bastani, O., Kim, C., Bastani, H.: Interpreting blackbox models via model
  extraction. CoRR  \textbf{abs/1705.08504} (2017)

\bibitem{rudin-cikmw22}
Behrouz, A., L{\'{e}}cuyer, M., Rudin, C., Seltzer, M.I.: Fast optimization of
  weighted sparse decision trees for use in optimal treatment regimes and
  optimal policy design. In: CIKM Workshops (2022)

\bibitem{blms-aicom12}
Belov, A., Lynce, I., Marques{-}Silva, J.: Towards efficient {MUS} extraction.
  {AI} Commun.  \textbf{25}(2),  97--116 (2012),
  \url{https://doi.org/10.3233/AIC-2012-0523}

\bibitem{mlcs-bk12}
Ben{-}Ari, M.: Mathematical Logic for Computer Science. Springer (2012),
  \url{https://doi.org/10.1007/978-1-4471-4129-7}

\bibitem{bendik-tacas20}
Bend{\'{\i}}k, J., Cern{\'{a}}, I.: {MUST:} minimal unsatisfiable subsets
  enumeration tool. In: TACAS. pp. 135--152 (2020)

\bibitem{bendik-lpar20}
Bend{\'{\i}}k, J., Cerna, I.: Rotation based {MSS/MCS} enumeration. In: LPAR.
  pp. 120--137 (2020)

\bibitem{bendik-atva18}
Bend{\'{\i}}k, J., Cern{\'{a}}, I., Benes, N.: Recursive online enumeration of
  all minimal unsatisfiable subsets. In: ATVA. pp. 143--159 (2018)

\bibitem{bengio-cacm21}
Bengio, Y., LeCun, Y., Hinton, G.E.: Deep learning for {AI}. Commun. {ACM}
  \textbf{64}(7),  58--65 (2021), \url{https://doi.org/10.1145/3448250}

\bibitem{coq-bk04}
Bertot, Y., Cast{\'{e}}ran, P.: Interactive Theorem Proving and Program
  Development - Coq'Art: The Calculus of Inductive Constructions. Texts in
  Theoretical Computer Science. An {EATCS} Series, Springer (2004),
  \url{https://doi.org/10.1007/978-3-662-07964-5}

\bibitem{bertsimas-ml17}
Bertsimas, D., Dunn, J.: Optimal classification trees. Mach. Learn.
  \textbf{106}(7),  1039--1082 (2017),
  \url{https://doi.org/10.1007/s10994-017-5633-9}

\bibitem{hebrard-cp09}
Bessiere, C., Hebrard, E., O'Sullivan, B.: Minimising decision tree size as
  combinatorial optimisation. In: CP. pp. 173--187 (2009)

\bibitem{bianchi-jair23}
Bianchi, F., Curry, A.C., Hovy, D.: Viewpoint: Artificial intelligence
  accidents waiting to happen? J. Artif. Intell. Res.  \textbf{76},  193--199
  (2023), \url{https://doi.org/10.1613/jair.1.14263}

\bibitem{bienvenu-jair09}
Bienvenu, M.: Prime implicates and prime implicants: From propositional to
  modal logic. J. Artif. Intell. Res.  \textbf{36},  71--128 (2009),
  \url{https://doi.org/10.1613/jair.2754}

\bibitem{sat-handbook21}
Biere, A., Heule, M., van Maaren, H., Walsh, T. (eds.): Handbook of
  Satisfiability. {IOS} Press (2021)

\bibitem{lozinskii-jetai03}
Birnbaum, E., Lozinskii, E.L.: Consistent subsets of inconsistent systems:
  structure and behaviour. J. Exp. Theor. Artif. Intell.  \textbf{15}(1),
  25--46 (2003), \url{https://doi.org/10.1080/0952813021000026795}

\bibitem{bixby-or02}
Bixby, R.E.: Solving real-world linear programs: {A} decade and more of
  progress. Oper. Res.  \textbf{50}(1),  3--15 (2002),
  \url{https://doi.org/10.1287/opre.50.1.3.17780}

\bibitem{bixby-aor07}
Bixby, R.E., Rothberg, E.: Progress in computational mixed integer programming
  - {A} look back from the other side of the tipping point. Ann. Oper. Res.
  \textbf{149}(1),  37--41 (2007),
  \url{https://doi.org/10.1007/s10479-006-0091-y}

\bibitem{tan-stoc22}
Blanc, G., Koch, C., Lange, J., Tan, L.: The query complexity of certification.
  In: STOC. pp. 623--636 (2022)

\bibitem{tan-icml22a}
Blanc, G., Koch, C., Lange, J., Tan, L.: A query-optimal algorithm for finding
  counterfactuals. In: ICML. pp. 2075--2090 (2022)

\bibitem{tan-nips21}
Blanc, G., Lange, J., Tan, L.: Provably efficient, succinct, and precise
  explanations. In: NeurIPS (2021)

\bibitem{guns-aij21}
Bogaerts, B., Gamba, E., Guns, T.: A framework for step-wise explaining how to
  solve constraint satisfaction problems. Artif. Intell.  \textbf{300},  103550
  (2021), \url{https://doi.org/10.1016/j.artint.2021.103550}

\bibitem{boros-esa03}
Boros, E., Elbassioni, K.M., Gurvich, V., Khachiyan, L.: An efficient
  implementation of a quasi-polynomial algorithm for generating hypergraph
  transversals. In: ESA. pp. 556--567 (2003)

\bibitem{mazure-sum20}
Boumazouza, R., Alili, F.C., Mazure, B., Tabia, K.: A symbolic approach for
  counterfactual explanations. In: SUM. pp. 270--277 (2020)

\bibitem{mazure-cikm21}
Boumazouza, R., Alili, F.C., Mazure, B., Tabia, K.: {ASTERYX:} {A}
  model-agnostic sat-based approach for symbolic and score-based explanations.
  In: CIKM. pp. 120--129 (2021)

\bibitem{bradley-cc07}
Bradley, A.R., Manna, Z.: The Calculus of Computation: Decision Procedures with
  Applications to Verification. Springer (2007)

\bibitem{bradley-fmcad07}
Bradley, A.R., Manna, Z.: Checking safety by inductive generalization of
  counterexamples to induction. In: FMCAD. pp. 173--180 (2007)

\bibitem{breiman-ml01}
Breiman, L.: Random forests. Mach. Learn.  \textbf{45}(1),  5--32 (2001),
  \url{https://doi.org/10.1023/A:1010933404324}

\bibitem{breiman-ss01}
Breiman, L.: Statistical modeling: The two cultures. Statistical science
  \textbf{16}(3),  199--231 (2001)

\bibitem{breiman-bk84}
Breiman, L., Friedman, J.H., Olshen, R.A., Stone, C.J.: Classification and
  Regression Trees. Wadsworth (1984)

\bibitem{utgoff-ml95}
Brodley, C.E., Utgoff, P.E.: Multivariate decision trees. Mach. Learn.
  \textbf{19}(1),  45--77 (1995)

\bibitem{vandenbroeck-aaai21}
den Broeck, G.V., Lykov, A., Schleich, M., Suciu, D.: On the tractability of
  {SHAP} explanations. In: AAAI. pp. 6505--6513 (2021)

\bibitem{bylander-aij91}
Bylander, T., Allemang, D., Tanner, M.C., Josephson, J.R.: The computational
  complexity of abduction. Artif. Intell.  \textbf{49}(1-3),  25--60 (1991),
  \url{https://doi.org/10.1016/0004-3702(91)90005-5}

\bibitem{bmpms-unpub23}
Béjar, R., Morgado, A., Planes, J., Marques-Silva, J.: Logic-based
  explainability with missing data (2023), working paper

\bibitem{lukasiewicz-corr19}
Camburu, O., Giunchiglia, E., Foerster, J., Lukasiewicz, T., Blunsom, P.: Can
  {I} trust the explainer? verifying post-hoc explanatory methods. CoRR
  \textbf{abs/1910.02065} (2019), \url{http://arxiv.org/abs/1910.02065}

\bibitem{kambhampati-ijcai20}
Chakraborti, T., Sreedharan, S., Kambhampati, S.: The emerging landscape of
  explainable automated planning {\&} decision making. In: IJCAI. pp.
  4803--4811 (2020)

\bibitem{rudin-dss22}
Chen, C., Lin, K., Rudin, C., Shaposhnik, Y., Wang, S., Wang, T.: A holistic
  approach to interpretability in financial lending: Models, visualizations,
  and summary-explanations. Decis. Support Syst.  \textbf{152},  113647 (2022),
  \url{https://doi.org/10.1016/j.dss.2021.113647}

\bibitem{rudin-aistats18}
Chen, C., Rudin, C.: An optimization approach to learning falling rule lists.
  In: AISTATS. pp. 604--612 (2018)

\bibitem{chinneck-jc91}
Chinneck, J.W., Dravnieks, E.W.: Locating minimal infeasible constraint sets in
  linear programs. {INFORMS} J. Comput.  \textbf{3}(2),  157--168 (1991)

\bibitem{clark-ml89}
Clark, P., Niblett, T.: The {CN2} induction algorithm. Machine Learning
  \textbf{3},  261--283 (1989), \url{https://doi.org/10.1007/BF00116835}

\bibitem{mc-handbook18}
Clarke, E.M., Henzinger, T.A., Veith, H., Bloem, R. (eds.): Handbook of Model
  Checking. Springer (2018), \url{https://doi.org/10.1007/978-3-319-10575-8}

\bibitem{cnn-tesla21}
{CNN}: Tesla is under investigation because its cars keep hitting emergency
  vehicles. \url{https://cnn.it/3z6NXGF} (August 2021)

\bibitem{cook-stoc71}
Cook, S.A.: The complexity of theorem-proving procedures. In: Harrison, M.A.,
  Banerji, R.B., Ullman, J.D. (eds.) STOC. pp. 151--158 (1971)

\bibitem{cms-cp21}
Cooper, M.C., Marques{-}Silva, J.: On the tractability of explaining decisions
  of classifiers. In: Michel, L.D. (ed.) CP. pp. 21:1--21:18 (2021)

\bibitem{cms-aij23}
Cooper, M.C., Marques-Silva, J.: Tractability of explaining classifier
  decisions. Artif. Intell.  (2023),
  \url{https://www.sciencedirect.com/science/article/pii/S0004370222001813}

\bibitem{kavvadias-tcs20}
Cosmadakis, S.S., Kavvadias, D.J., Panagopoulou, L.: Resolution based
  algorithms for the transversal hypergraph generation problem. Theor. Comput.
  Sci.  \textbf{815},  1--10 (2020),
  \url{https://doi.org/10.1016/j.tcs.2020.02.033}

\bibitem{crama-bk11}
Crama, Y., Hammer, P.L.: Boolean Functions - Theory, Algorithms, and
  Applications. Cambridge University Press (2011),
  \url{http://www.cambridge.org/gb/knowledge/isbn/item6222210/?site\_locale=en\_GB}

\bibitem{cfmssk-tacas17}
Cruz{-}Filipe, L., Marques{-}Silva, J., Schneider{-}Kamp, P.: Efficient
  certified resolution proof checking. In: TACAS. pp. 118--135 (2017)

\bibitem{cfmssk-jar19}
Cruz{-}Filipe, L., Marques{-}Silva, J., Schneider{-}Kamp, P.: Formally
  verifying the solution to the boolean pythagorean triples problem. J. Autom.
  Reason.  \textbf{63}(3),  695--722 (2019),
  \url{https://doi.org/10.1007/s10817-018-9490-4}

\bibitem{fomin-bk15}
Cygan, M., Fomin, F.V., Kowalik, L., Lokshtanov, D., Marx, D., Pilipczuk, M.,
  Pilipczuk, M., Saurabh, S.: Parameterized Algorithms. Springer (2015),
  \url{https://doi.org/10.1007/978-3-319-21275-3}

\bibitem{trivedi-ieee-comp17}
Daily, M., Medasani, S., Behringer, R., Trivedi, M.M.: Self-driving cars.
  Computer  \textbf{50}(12),  18--23 (2017),
  \url{https://doi.org/10.1109/MC.2017.4451204}

\bibitem{dfms-ictac10}
Darbari, A., Fischer, B., Marques{-}Silva, J.: Industrial-strength certified
  {SAT} solving through verified {SAT} proof checking. In: ICTAC. pp. 260--274
  (2010)

\bibitem{darpa-xai16}
DARPA: {DARPA} explainable {A}rtificial {I}ntelligence {(XAI)} program.
  \url{https://www.darpa.mil/program/explainable-artificial-intelligence}
  (2016), accessed: 2021-12-01

\bibitem{darwiche-jacm01}
Darwiche, A.: Decomposable negation normal form. J. {ACM}  \textbf{48}(4),
  608--647 (2001), \url{https://doi.org/10.1145/502090.502091}

\bibitem{darwiche-jancl01}
Darwiche, A.: On the tractable counting of theory models and its application to
  truth maintenance and belief revision. J. Appl. Non Class. Logics
  \textbf{11}(1-2),  11--34 (2001),
  \url{https://doi.org/10.3166/jancl.11.11-34}

\bibitem{darwiche-pods20}
Darwiche, A.: Three modern roles for logic in {AI}. In: PODS. pp. 229--243
  (2020)

\bibitem{darwiche-ecai20}
Darwiche, A., Hirth, A.: On the reasons behind decisions. In: ECAI. pp.
  712--720 (2020)

\bibitem{darwiche-jlli22-web}
Darwiche, A., Hirth, A.: On the (complete) reasons behind decisions. Journal of
  Logic, Language and Information pp. 1--26 (2022), in press

\bibitem{darwiche-aaai22}
Darwiche, A., Ji, C.: On the computation of necessary and sufficient
  explanations. In: AAAI. pp. 5582--5591 (2022)

\bibitem{darwiche-jair02}
Darwiche, A., Marquis, P.: A knowledge compilation map. J. Artif. Intell. Res.
  \textbf{17},  229--264 (2002), \url{https://doi.org/10.1613/jair.989}

\bibitem{darwiche-jair21}
Darwiche, A., Marquis, P.: On quantifying literals in boolean logic and its
  applications to explainable {AI}. J. Artif. Intell. Res.  (2021)

\bibitem{davis-cacm62}
Davis, M., Logemann, G., Loveland, D.W.: A machine program for theorem-proving.
  Commun. {ACM}  \textbf{5}(7),  394--397 (1962),
  \url{https://doi.org/10.1145/368273.368557}

\bibitem{davis-jacm60}
Davis, M., Putnam, H.: A computing procedure for quantification theory. J.
  {ACM}  \textbf{7}(3),  201--215 (1960),
  \url{http://doi.acm.org/10.1145/321033.321034}

\bibitem{dearman-alr19}
DeArman, A.: The wild, wild west: A case study of self-driving vehicle testing
  in arizona. Ariz. L. Rev.  \textbf{61}, ~983 (2019)

\bibitem{demirovic-jmlr22}
Demirovic, E., Lukina, A., Hebrard, E., Chan, J., Bailey, J., Leckie, C.,
  Ramamohanarao, K., Stuckey, P.J.: Murtree: Optimal decision trees via dynamic
  programming and search. J. Mach. Learn. Res.  \textbf{23},  26:1--26:47
  (2022), \url{http://jmlr.org/papers/v23/20-520.html}

\bibitem{demirovic-aaai21}
Demirovic, E., Stuckey, P.J.: Optimal decision trees for nonlinear metrics. In:
  AAAI. pp. 3733--3741 (2021)

\bibitem{mcmillan-cav12}
Dillig, I., Dillig, T., McMillan, K.L., Aiken, A.: Minimum satisfying
  assignments for {SMT}. In: CAV. pp. 394--409 (2012)

\bibitem{weller-ecai20}
Dimanov, B., Bhatt, U., Jamnik, M., Weller, A.: You shouldn't trust me:
  Learning models which conceal unfairness from multiple explanation methods.
  In: ECAI. pp. 2473--2480 (2020)

\bibitem{hoffmann-aaai20}
Eifler, R., Cashmore, M., Hoffmann, J., Magazzeni, D., Steinmetz, M.: A new
  approach to plan-space explanation: Analyzing plan-property dependencies in
  oversubscription planning. In: AAAI. pp. 9818--9826 (2020)

\bibitem{hoffmann-ijcai22}
Eifler, R., Frank, J., Hoffmann, J.: Explaining soft-goal conflicts through
  constraint relaxations. In: IJCAI. pp. 4621--4627 (2022),
  \url{https://doi.org/10.24963/ijcai.2022/641}

\bibitem{gottlob-jacm95}
Eiter, T., Gottlob, G.: The complexity of logic-based abduction. J. {ACM}
  \textbf{42}(1),  3--42 (1995), \url{https://doi.org/10.1145/200836.200838}

\bibitem{gottlob-sjc95}
Eiter, T., Gottlob, G.: Identifying the minimal transversals of a hypergraph
  and related problems. {SIAM} J. Comput.  \textbf{24}(6),  1278--1304 (1995)

\bibitem{nordstrom-aaai20}
Elffers, J., Gocht, S., McCreesh, C., Nordstr{\"{o}}m, J.: Justifying all
  differences using pseudo-boolean reasoning. In: AAAI. pp. 1486--1494 (2020)

\bibitem{gent-corr21}
Espasa, J., Gent, I.P., Hoffmann, R., Jefferson, C., Lynch, A.M.: Using small
  {MUSes} to explain how to solve pen and paper puzzles. CoRR
  \textbf{abs/2104.15040} (2021), \url{https://arxiv.org/abs/2104.15040}

\bibitem{eu-gdpr16}
{EU}: {General Data Protection Regulation}.
  \url{https://eur-lex.europa.eu/eli/reg/2016/679/oj} (2016), accessed:
  2021-12-01

\bibitem{eu-aiact21}
EU: {Artificial Intelligence Act}. \url{tiny.cc/wy8juz} (2021), accessed:
  2021-12-01

\bibitem{eu-coplan21}
EU: Coordinated plan on artificial intelligence -- 2021 review.
  \url{https://bit.ly/3hJG2HF} (2021), accessed: 2021-12-01

\bibitem{simari-aij02}
Falappa, M.A., Kern{-}Isberner, G., Simari, G.R.: Explanations, belief revision
  and defeasible reasoning. Artif. Intell.  \textbf{141}(1/2),  1--28 (2002)

\bibitem{toni-ecai14}
Fan, X., Toni, F.: On computing explanations in abstract argumentation. In:
  ECAI. pp. 1005--1006 (2014)

\bibitem{gupta-nips16}
Fard, M.M., Canini, K.R., Cotter, A., Pfeifer, J., Gupta, M.R.: Fast and
  flexible monotonic functions with ensembles of lattices. In: NeurIPS. pp.
  2919--2927 (2016)

\bibitem{felfernig-aiedam12}
Felfernig, A., Schubert, M., Zehentner, C.: An efficient diagnosis algorithm
  for inconsistent constraint sets. {AI} {EDAM}  \textbf{26}(1),  53--62
  (2012), \url{https://doi.org/10.1017/S0890060411000011}

\bibitem{leite-kr22}
Ferreira, J., de~Sousa~Ribeiro, M., Gon{\c{c}}alves, R., Leite, J.: Looking
  inside the black-box: Logic-based explanations for neural networks. In: KR.
  p. 432–442 (2022)

\bibitem{fischetti-cj18}
Fischetti, M., Jo, J.: Deep neural networks and mixed integer linear
  optimization. Constraints An Int. J.  \textbf{23}(3),  296--309 (2018),
  \url{https://doi.org/10.1007/s10601-018-9285-6}

\bibitem{flach-bk12}
Flach, P.A.: Machine Learning - The Art and Science of Algorithms that Make
  Sense of Data. CUP (2012)

\bibitem{fomin-bk10}
Fomin, F.V., Kratsch, D.: Exact Exponential Algorithms. Texts in Theoretical
  Computer Science. An {EATCS} Series, Springer (2010),
  \url{https://doi.org/10.1007/978-3-642-16533-7}

\bibitem{fox-corr17}
Fox, M., Long, D., Magazzeni, D.: Explainable planning. CoRR
  \textbf{abs/1709.10256} (2017), \url{http://arxiv.org/abs/1709.10256}

\bibitem{khachiyan-jalg96}
Fredman, M.L., Khachiyan, L.: On the complexity of dualization of monotone
  disjunctive normal forms. J. Algorithms  \textbf{21}(3),  618--628 (1996)

\bibitem{gottlob-ese90}
Friedrich, G., Gottlob, G., Nejdl, W.: Hypothesis classification, abductive
  diagnosis and therapy. In: ESE. pp. 69--78 (1990)

\bibitem{hinton-cexaiia17}
Frosst, N., Hinton, G.E.: Distilling a neural network into a soft decision
  tree. In: CExAIIA (2017)

\bibitem{guns-ijcai21}
Gamba, E., Bogaerts, B., Guns, T.: Efficiently explaining {CSPs} with
  unsatisfiable subset optimization. In: IJCAI. pp. 1381--1388 (2021)

\bibitem{vardi-cup20}
Ganesh, V., Vardi, M.Y.: On the unreasonable effectiveness of {SAT} solvers.
  In: Roughgarden, T. (ed.) Beyond the Worst-Case Analysis of Algorithms, pp.
  547--566. Cambridge University Press (2020),
  \url{https://doi.org/10.1017/9781108637435.032}

\bibitem{korovin-lics03}
Ganzinger, H., Korovin, K.: New directions in instantiation-based theorem
  proving. In: LICS. pp. 55--64 (2003)

\bibitem{wired05}
Garfinkel, S.: History's worst software bugs.
  \url{https://www.wired.com/2005/11/historys-worst-software-bugs/} (2005)

\bibitem{schaub-bk12}
Gebser, M., Kaminski, R., Kaufmann, B., Schaub, T.: Answer Set Solving in
  Practice. Synthesis Lectures on Artificial Intelligence and Machine Learning,
  Morgan {\&} Claypool Publishers (2012)

\bibitem{ghassemi-ldh21}
Ghassemi, M., Oakden-Rayner, L., Beam, A.L.: The false hope of current
  approaches to explainable artificial intelligence in health care. The Lancet
  Digital Health  \textbf{3}(11),  e745--e750 (2021)

\bibitem{meel-ecai20}
Ghosh, B., Malioutov, D., Meel, K.S.: Classification rules in relaxed logical
  form. In: ECAI. pp. 2489--2496 (2020)

\bibitem{meel-jair22}
Ghosh, B., Malioutov, D., Meel, K.S.: Efficient learning of interpretable
  classification rules. J. Artif. Intell. Res.  \textbf{74},  1823--1863
  (2022), \url{https://doi.org/10.1613/jair.1.13482}

\bibitem{meel-aies19}
Ghosh, B., Meel, K.S.: {IMLI:} an incremental framework for maxsat-based
  learning of interpretable classification rules. In: AIES. pp. 203--210 (2019)

\bibitem{nordstrom-cp20}
Gocht, S., McBride, R., McCreesh, C., Nordstr{\"{o}}m, J., Prosser, P.,
  Trimble, J.: Certifying solvers for clique and maximum common (connected)
  subgraph problems. In: CP. pp. 338--357 (2020)

\bibitem{nordstrom-ijcai20}
Gocht, S., McCreesh, C., Nordstr{\"{o}}m, J.: Subgraph isomorphism meets
  cutting planes: Solving with certified solutions. In: IJCAI. pp. 1134--1140
  (2020)

\bibitem{nordstrom-aaai21}
Gocht, S., Nordstr{\"{o}}m, J.: Certifying parity reasoning efficiently using
  pseudo-boolean proofs. In: AAAI. pp. 3768--3777 (2021)

\bibitem{bengio-bk16}
Goodfellow, I.J., Bengio, Y., Courville, A.C.: Deep Learning. Adaptive
  computation and machine learning, {MIT} Press (2016),
  \url{http://www.deeplearningbook.org/}

\bibitem{bengio-cacm20}
Goodfellow, I.J., Pouget{-}Abadie, J., Mirza, M., Xu, B., Warde{-}Farley, D.,
  Ozair, S., Courville, A.C., Bengio, Y.: Generative adversarial networks.
  Commun. {ACM}  \textbf{63}(11),  139--144 (2020),
  \url{https://doi.org/10.1145/3422622}

\bibitem{goodman-aimag17}
Goodman, B., Flaxman, S.R.: European union regulations on algorithmic
  decision-making and a "right to explanation". {AI} Mag.  \textbf{38}(3),
  50--57 (2017), \url{https://doi.org/10.1609/aimag.v38i3.2741}

\bibitem{rubin-aaai22}
Gorji, N., Rubin, S.: Sufficient reasons for classifier decisions in the
  presence of domain constraints. In: {AAAI} (February 2022)

\bibitem{lagniez-ijcai18}
Gr{\'{e}}goire, {\'{E}}., Izza, Y., Lagniez, J.: Boosting {MCSes} enumeration.
  In: IJCAI. pp. 1309--1315 (2018)

\bibitem{lagniez-aaai14}
Gr{\'{e}}goire, {\'{E}}., Lagniez, J., Mazure, B.: An experimentally efficient
  method for {(MSS, CoMSS)} partitioning. In: AAAI. pp. 2666--2673 (2014)

\bibitem{greiner-aij89}
Greiner, R., Smith, B.A., Wilkerson, R.W.: A correction to the algorithm in
  reiter's theory of diagnosis. Artif. Intell.  \textbf{41}(1),  79--88 (1989),
  \url{https://doi.org/10.1016/0004-3702(89)90079-9}

\bibitem{pedreschi-acmcs19}
Guidotti, R., Monreale, A., Ruggieri, S., Turini, F., Giannotti, F., Pedreschi,
  D.: A survey of methods for explaining black box models. {ACM} Comput. Surv.
  \textbf{51}(5),  93:1--93:42 (2019)

\bibitem{osullivan-ijcai21}
Gupta, S.D., Genc, B., O'Sullivan, B.: Explanation in constraint satisfaction:
  {A} survey. In: ICJAI. pp. 4400--4407 (2021)

\bibitem{somenzi-bk06}
Hachtel, G.D., Somenzi, F.: Logic synthesis and verification algorithms.
  Springer (2006), \url{https://doi.org/10.1007/0-387-31005-3}

\bibitem{peirce-works31}
Hartshorne, C., Weiss, P. (eds.): Collected Papers of {C}harles {S}anders
  {P}eirce. Harvard University Press (1931)

\bibitem{oppenheim-ps48}
Hempel, C.G., Oppenheim, P.: Studies in the logic of explanation. Philosophy of
  science  \textbf{15}(2),  135--175 (1948)

\bibitem{heule-aaai18}
Heule, M.J.H.: Schur number five. In: AAAI. pp. 6598--6606 (2018)

\bibitem{heule-cacm17}
Heule, M.J.H., Kullmann, O.: The science of brute force. Commun. {ACM}
  \textbf{60}(8),  70--79 (2017), \url{https://doi.org/10.1145/3107239}

\bibitem{eu-hlegai19}
{{HLEG} AI}: Ethics guidelines for trustworthy {AI}.
  \url{https://digital-strategy.ec.europa.eu/en/library/ethics-guidelines-trustworthy-ai}
  (2019), accessed: 2021-12-01

\bibitem{eu-hlegai20}
{{HLEG} AI}: Assessment list for trustworthy artificial intelligence {(ALTAI)}
  for self-assessment. \url{https://bit.ly/3jAeHds} (2020), accessed:
  2021-12-01

\bibitem{hoffmann-rw19}
Hoffmann, J., Magazzeni, D.: Explainable {AI} planning {(XAIP):} overview and
  the case of contrastive explanation (extended abstract). In: Kr{\"{o}}tzsch,
  M., Stepanova, D. (eds.) RW. pp. 277--282 (2019)

\bibitem{holzinger-bk22}
Holzinger, A., Goebel, R., Fong, R., Moon, T., M{\"{u}}ller, K., Samek, W.
  (eds.): xxAI - Beyond Explainable {AI} - International Workshop, Held in
  Conjunction with {ICML} 2020, July 18, 2020, Vienna, Austria, Revised and
  Extended Papers, Lecture Notes in Computer Science, vol. 13200. Springer
  (2022), \url{https://doi.org/10.1007/978-3-031-04083-2}

\bibitem{holzinger-xxai22b}
Holzinger, A., Saranti, A., Molnar, C., Biecek, P., Samek, W.: Explainable {AI}
  methods - {A} brief overview. In: xxAI. pp. 13--38 (2020)

\bibitem{hebrard-ijcai20}
Hu, H., Siala, M., Hebrard, E., Huguet, M.: Learning optimal decision trees
  with {MaxSAT} and its integration in {AdaBoost}. In: IJCAI. pp. 1170--1176
  (2020)

\bibitem{rudin-nips19}
Hu, X., Rudin, C., Seltzer, M.I.: Optimal sparse decision trees. In: NeurIPS.
  pp. 7265--7273 (2019)

\bibitem{hcmpms-tacas23}
Huang, X., Cooper, M.C., Morgado, A., Planes, J., Marques{-}Silva, J.: Feature
  necessity \& relevancy in {ML} classifier explanations. In: TACAS (2023)

\bibitem{hiicams-corr21}
Huang, X., Izza, Y., Ignatiev, A., Cooper, M.C., Asher, N., Marques{-}Silva,
  J.: Efficient explanations for knowledge compilation languages. CoRR
  \textbf{abs/2107.01654} (2021), \url{https://arxiv.org/abs/2107.01654}

\bibitem{hiicams-aaai22}
Huang, X., Izza, Y., Ignatiev, A., Cooper, M.C., Asher, N., Marques{-}Silva,
  J.: Tractable explanations for {d-DNNF} classifiers. In: AAAI. pp. 5719--5728
  (2022)

\bibitem{hiims-kr21}
Huang, X., Izza, Y., Ignatiev, A., Marques{-}Silva, J.: On efficiently
  explaining graph-based classifiers. In: KR. pp. 356--367 (2021)

\bibitem{hims-aaai23}
Huang, X., Izza, Y., Marques-Silva, J.: Solving explainability queries with
  quantification: The case of feature membership. In: {AAAI} (2023)

\bibitem{hms-corr22}
Huang, X., Marques{-}Silva, J.: On deciding feature membership in explanations
  of {SDD} {\&} related classifiers. CoRR  \textbf{abs/2202.07553} (2022),
  \url{https://arxiv.org/abs/2202.07553}

\bibitem{bengio-nips16}
Hubara, I., Courbariaux, M., Soudry, D., El{-}Yaniv, R., Bengio, Y.: Binarized
  neural networks. In: NeurIPS. pp. 4107--4115 (2016)

\bibitem{ignatiev-ijcai20}
Ignatiev, A.: Towards trustable explainable {AI}. In: IJCAI. pp. 5154--5158
  (2020)

\bibitem{icshms-cp20}
Ignatiev, A., Cooper, M.C., Siala, M., Hebrard, E., Marques{-}Silva, J.:
  Towards formal fairness in machine learning. In: CP. pp. 846--867 (2020)

\bibitem{iisms-aaai22}
Ignatiev, A., Izza, Y., Stuckey, P.J., Marques{-}Silva, J.: Using {MaxSAT} for
  efficient explanations of tree ensembles. In: AAAI. pp. 3776--3785 (2022)

\bibitem{ilsms-aaai21}
Ignatiev, A., Lam, E., Stuckey, P.J., Marques{-}Silva, J.: A scalable two stage
  approach to computing optimal decision sets. In: AAAI. pp. 3806--3814 (2021)

\bibitem{ims-sat21}
Ignatiev, A., Marques{-}Silva, J.: {SAT}-based rigorous explanations for
  decision lists. In: SAT. pp. 251--269 (2021)

\bibitem{imms-ecai16}
Ignatiev, A., Morgado, A., Marques{-}Silva, J.: Propositional abduction with
  implicit hitting sets. pp. 1327--1335 (2016)

\bibitem{imms-sat18}
Ignatiev, A., Morgado, A., Marques{-}Silva, J.: {PySAT}: {A} python toolkit for
  prototyping with {SAT} oracles. In: SAT. pp. 428--437 (2018)

\bibitem{inams-aiia20}
Ignatiev, A., Narodytska, N., Asher, N., Marques{-}Silva, J.: From contrastive
  to abductive explanations and back again. In: AIxIA. pp. 335--355 (2020)

\bibitem{inms-aaai19}
Ignatiev, A., Narodytska, N., Marques{-}Silva, J.: Abduction-based explanations
  for machine learning models. In: AAAI. pp. 1511--1519 (2019)

\bibitem{inms-nips19}
Ignatiev, A., Narodytska, N., Marques{-}Silva, J.: On relating explanations and
  adversarial examples. In: NeurIPS. pp. 15857--15867 (2019)

\bibitem{inms-corr19}
Ignatiev, A., Narodytska, N., Marques{-}Silva, J.: On validating, repairing and
  refining heuristic {ML} explanations. CoRR  \textbf{abs/1907.02509} (2019),
  \url{http://arxiv.org/abs/1907.02509}

\bibitem{ipnms-ijcar18}
Ignatiev, A., Pereira, F., Narodytska, N., Marques{-}Silva, J.: A {SAT}-based
  approach to learn explainable decision sets. In: IJCAR. pp. 627--645 (2018)

\bibitem{iplms-cp15}
Ignatiev, A., Previti, A., Liffiton, M.H., Marques{-}Silva, J.: Smallest {MUS}
  extraction with minimal hitting set dualization. In: CP. pp. 173--182 (2015)

\bibitem{it-tesla21}
{Irish Times}: {‘It happened so fast’: Inside a fatal Tesla Autopilot
  crash}. \url{https://bit.ly/3mnvzFJ}

\bibitem{israni-jorlt17}
Israni, E.: Algorithmic due process: Mistaken accountability and attribution in
  {S}tate v. {L}oomis. Harvard Journal of Law and Technology  (2017), published
  online: \url{https://bit.ly/389PYcq}

\bibitem{ihincms-corr22}
Izza, Y., Huang, X., Ignatiev, A., Narodytska, N., Cooper, M.C.,
  Marques{-}Silva, J.: On computing probabilistic abductive explanations. CoRR
  \textbf{abs/2212.05990} (2022). \doi{10.48550/arXiv.2212.05990},
  \url{https://doi.org/10.48550/arXiv.2212.05990}

\bibitem{iims-corr20}
Izza, Y., Ignatiev, A., Marques{-}Silva, J.: On explaining decision trees. CoRR
   \textbf{abs/2010.11034} (2020), \url{https://arxiv.org/abs/2010.11034}

\bibitem{iims-jair22}
Izza, Y., Ignatiev, A., Marques{-}Silva, J.: On tackling explanation redundancy
  in decision trees. J. Artif. Intell. Res.  \textbf{75},  261--321 (2022),
  \url{https://jair.org/index.php/jair/article/view/13575/}

\bibitem{iims-corr22}
Izza, Y., Ignatiev, A., Marques{-}Silva, J.: On tackling explanation redundancy
  in decision trees. CoRR  \textbf{abs/2205.09971} (2022),
  \url{https://doi.org/10.48550/arXiv.2205.09971}

\bibitem{iincms-corr21}
Izza, Y., Ignatiev, A., Narodytska, N., Cooper, M.C., Marques{-}Silva, J.:
  Efficient explanations with relevant sets. CoRR  \textbf{abs/2106.00546}
  (2021), \url{https://arxiv.org/abs/2106.00546}

\bibitem{iincms-corr22}
Izza, Y., Ignatiev, A., Narodytska, N., Cooper, M.C., Marques{-}Silva, J.:
  Provably precise, succinct and efficient explanations for decision trees.
  CoRR  \textbf{abs/2205.09569} (2022),
  \url{https://doi.org/10.48550/arXiv.2205.09569}

\bibitem{ims-ijcai21}
Izza, Y., Marques{-}Silva, J.: On explaining random forests with {SAT}. In:
  IJCAI. pp. 2584--2591 (2021)

\bibitem{ims-corr22}
Izza, Y., Marques{-}Silva, J.: On computing relevant features for explaining
  {NBCs}. CoRR  \textbf{abs/2207.04748} (2022),
  \url{https://doi.org/10.48550/arXiv.2207.04748}

\bibitem{jkmsc-aij16}
Janota, M., Klieber, W., Marques{-}Silva, J., Clarke, E.M.: Solving {QBF} with
  counterexample guided refinement. Artif. Intell.  \textbf{234},  1--25
  (2016), \url{https://doi.org/10.1016/j.artint.2016.01.004}

\bibitem{jms-sat11}
Janota, M., Marques-Silva, J.: Abstraction-based algorithm for {2QBF}. In: SAT.
  pp. 230--244 (2011)

\bibitem{jms-ijcai15}
Janota, M., Marques{-}Silva, J.: Solving {QBF} by clause selection. In: IJCAI.
  pp. 325--331 (2015)

\bibitem{janota-sat20}
Janota, M., Morgado, A.: {SAT}-based encodings for optimal decision trees with
  explicit paths. In: SAT. pp. 501--518 (2020)

\bibitem{juba-aaai16}
Juba, B.: Learning abductive reasoning using random examples. In: AAAI. pp.
  999--1007 (2016)

\bibitem{junker-aaai04}
Junker, U.: {QUICKXPLAIN:} preferred explanations and relaxations for
  over-constrained problems. In: AAAI. pp. 167--172 (2004)

\bibitem{kahn-fortune22}
Kahn, J.: {What's wrong with "explainable A.I."}. \url{https://bit.ly/3rLENgZ}
  (2016)

\bibitem{valera-corr20}
Karimi, A., Barthe, G., Sch{\"{o}}lkopf, B., Valera, I.: A survey of
  algorithmic recourse: definitions, formulations, solutions, and prospects.
  CoRR  \textbf{abs/2010.04050} (2020), \url{https://arxiv.org/abs/2010.04050},
  accepted for publication at ACM Computing Surveys.

\bibitem{valera-facct21}
Karimi, A., Sch{\"{o}}lkopf, B., Valera, I.: Algorithmic recourse: from
  counterfactual explanations to interventions. In: FAccT. pp. 353--362 (2021)

\bibitem{barrett-cav17}
Katz, G., Barrett, C.W., Dill, D.L., Julian, K., Kochenderfer, M.J.: Reluplex:
  An efficient {SMT} solver for verifying deep neural networks. In: CAV. pp.
  97--117 (2017), \url{https://doi.org/10.1007/978-3-319-63387-9\_5}

\bibitem{barrett-cav19}
Katz, G., Huang, D.A., Ibeling, D., Julian, K., Lazarus, C., Lim, R., Shah, P.,
  Thakoor, S., Wu, H., Zeljic, A., Dill, D.L., Kochenderfer, M.J., Barrett,
  C.W.: The marabou framework for verification and analysis of deep neural
  networks. In: CAV. pp. 443--452 (2019)

\bibitem{kavvadias-wae99}
Kavvadias, D.J., Stavropoulos, E.C.: Evaluation of an algorithm for the
  transversal hypergraph problem. In: WAE. pp. 72--84 (1999)

\bibitem{kavvadias-jgaa05}
Kavvadias, D.J., Stavropoulos, E.C.: An efficient algorithm for the transversal
  hypergraph generation. J. Graph Algorithms Appl.  \textbf{9}(2),  239--264
  (2005), \url{https://doi.org/10.7155/jgaa.00107}

\bibitem{boros-cocoon05}
Khachiyan, L., Boros, E., Elbassioni, K.M., Gurvich, V.: A new algorithm for
  the hypergraph transversal problem. In: COCOON. pp. 767--776 (2005)

\bibitem{boros-dam06}
Khachiyan, L., Boros, E., Elbassioni, K.M., Gurvich, V.: An efficient
  implementation of a quasi-polynomial algorithm for generating hypergraph
  transversals and its application in joint generation. Discret. Appl. Math.
  \textbf{154}(16),  2350--2372 (2006),
  \url{https://doi.org/10.1016/j.dam.2006.04.012}

\bibitem{routledge-beq22}
Kim, T.W., Routledge, B.R.: Why a right to an explanation of algorithmic
  decision-making should exist: A trust-based approach. Business Ethics
  Quarterly  \textbf{32}(1),  75--102 (2022)

\bibitem{adebayo-xai19}
Kindermans, P., Hooker, S., Adebayo, J., Alber, M., Sch{\"{u}}tt, K.T.,
  D{\"{a}}hne, S., Erhan, D., Kim, B.: The (un)reliability of saliency methods.
  In: Samek, W., Montavon, G., Vedaldi, A., Hansen, L.K., M{\"{u}}ller, K.
  (eds.) Explainable {AI:} Interpreting, Explaining and Visualizing Deep
  Learning, Lecture Notes in Computer Science, vol. 11700, pp. 267--280.
  Springer (2019), \url{https://doi.org/10.1007/978-3-030-28954-6\_14}

\bibitem{buning-bk99}
Kleine{ }B{\"{u}}ning, H., Lettmann, T.: Propositional logic - deduction and
  algorithms, Cambridge tracts in theoretical computer science, vol.~48.
  Cambridge University Press (1999)

\bibitem{knight-icse02}
Knight, J.C.: Safety critical systems: challenges and directions. In: ICSE. pp.
  547--550 (2002)

\bibitem{berthold-eurojco22}
Koch, T., Berthold, T., Pedersen, J., Vanaret, C.: Progress in mathematical
  programming solvers from 2001 to 2020. EURO Journal on Computational
  Optimization p. 100031 (2022)

\bibitem{korovin-ijcar08}
Korovin, K.: {iProver} - an instantiation-based theorem prover for first-order
  logic (system description). In: IJCAR. pp. 292--298 (2008)

\bibitem{voronkov-cav13}
Kov{\'{a}}cs, L., Voronkov, A.: First-order theorem proving and vampire. In:
  CAV. pp. 1--35 (2013)

\bibitem{magazzeni-jair21}
Krarup, B., Krivic, S., Magazzeni, D., Long, D., Cashmore, M., Smith, D.E.:
  Contrastive explanations of plans through model restrictions. J. Artif.
  Intell. Res.  \textbf{72},  533--612 (2021),
  \url{https://doi.org/10.1613/jair.1.12813}

\bibitem{lakkaraju-corr22a}
Krishna, S., Han, T., Gu, A., Pombra, J., Jabbari, S., Wu, S., Lakkaraju, H.:
  The disagreement problem in explainable machine learning: {A} practitioner's
  perspective. CoRR  \textbf{abs/2202.01602} (2022),
  \url{https://arxiv.org/abs/2202.01602}

\bibitem{hinton-cacm17}
Krizhevsky, A., Sutskever, I., Hinton, G.E.: Imagenet classification with deep
  convolutional neural networks. Commun. {ACM}  \textbf{60}(6),  84--90 (2017),
  \url{http://doi.acm.org/10.1145/3065386}

\bibitem{kroening-bk16}
Kroening, D., Strichman, O.: Decision Procedures - An Algorithmic Point of
  View, Second Edition. Texts in Theoretical Computer Science. An {EATCS}
  Series, Springer (2016), \url{https://doi.org/10.1007/978-3-662-50497-0}

\bibitem{labreuche-sum22}
Labreuche, C.: Explanation of pseudo-boolean functions using cooperative game
  theory and prime implicants. In: SUM. pp. 295--308 (2022)

\bibitem{leskovec-kdd16}
Lakkaraju, H., Bach, S.H., Leskovec, J.: Interpretable decision sets: {A} joint
  framework for description and prediction. In: KDD. pp. 1675--1684 (2016)

\bibitem{lakkaraju-aies20b}
Lakkaraju, H., Bastani, O.: "how do {I} fool you?": Manipulating user trust via
  misleading black box explanations. In: AIES. pp. 79--85 (2020)

\bibitem{bengio-nature15}
LeCun, Y., Bengio, Y., Hinton, G.: Deep learning. nature  \textbf{521}(7553),
  436--444 (2015)

\bibitem{arstech18}
Lee, T.B.: Report: Software bug led to death in {U}ber's self-driving crash.
  \url{https://arstechnica.com/tech-policy/2018/05/report-software-bug-led-to-death-in-ubers-self-driving-crash/}
  (2018)

\bibitem{belmonte-ieee-access20}
Lelis, V.M., Guzm{\'{a}}n, E., Belmonte, M.: Non-invasive meningitis diagnosis
  using decision trees. {IEEE} Access  \textbf{8},  18394--18407 (2020).
  \doi{10.1109/ACCESS.2020.2966397},
  \url{https://doi.org/10.1109/ACCESS.2020.2966397}

\bibitem{liffiton-cpaior13}
Liffiton, M.H., Malik, A.: Enumerating infeasibility: Finding multiple {MUSes}
  quickly. In: CPAIOR. pp. 160--175 (2013)

\bibitem{lpmms-cj16}
Liffiton, M.H., Previti, A., Malik, A., Marques{-}Silva, J.: Fast, flexible
  {MUS} enumeration. Constraints An Int. J.  \textbf{21}(2),  223--250 (2016),
  \url{https://doi.org/10.1007/s10601-015-9183-0}

\bibitem{liffiton-jar08}
Liffiton, M.H., Sakallah, K.A.: Algorithms for computing minimal unsatisfiable
  subsets of constraints. J. Autom. Reason.  \textbf{40}(1),  1--33 (2008),
  \url{https://doi.org/10.1007/s10817-007-9084-z}

\bibitem{rudin-icml20}
Lin, J., Zhong, C., Hu, D., Rudin, C., Seltzer, M.I.: Generalized and scalable
  optimal sparse decision trees. In: ICML. pp. 6150--6160 (2020)

\bibitem{barrett-fto21}
Liu, C., Arnon, T., Lazarus, C., Strong, C.A., Barrett, C.W., Kochenderfer,
  M.J.: Algorithms for verifying deep neural networks. Found. Trends Optim.
  \textbf{4}(3-4),  244--404 (2021), \url{https://doi.org/10.1561/2400000035}

\bibitem{rudin-nips22}
Liu, J., Zhong, C., Li, B., Seltzer, M., Rudin, C.: Fasterrisk: Fast and
  accurate interpretable risk scores. In: NeurIPS (2022)

\bibitem{lin-nips20}
Liu, X., Han, X., Zhang, N., Liu, Q.: Certified monotonic neural networks. In:
  NeurIPS (2020)

\bibitem{lorini-clar21}
Liu, X., Lorini, E.: A logic for binary classifiers and their explanation. In:
  CLAR (2021)

\bibitem{lorini-wollic22}
Liu, X., Lorini, E.: A logic of "black box" classifier systems. In: WoLLIC. pp.
  158--174 (2022)

\bibitem{lundberg-naturemi20}
Lundberg, S.M., Erion, G., Chen, H., DeGrave, A., Prutkin, J.M., Nair, B.,
  Katz, R., Himmelfarb, J., Bansal, N., Lee, S.I.: From local explanations to
  global understanding with explainable {AI} for trees. Nature machine
  intelligence  \textbf{2}(1),  56--67 (2020)

\bibitem{lundberg-nips17}
Lundberg, S.M., Lee, S.: A unified approach to interpreting model predictions.
  In: NeurIPS. pp. 4765--4774 (2017)

\bibitem{luo-sc19}
Luo, J., Liu, S.: Accelerating {MUS} enumeration by inconsistency graph
  partitioning. {SCIENCE} {CHINA} Information Sciences  \textbf{62}(11),
  212104:1--212104:11 (2019)

\bibitem{kwiatkowska-ijcai21}
Malfa, E.L., Michelmore, R., Zbrzezny, A.M., Paoletti, N., Kwiatkowska, M.: On
  guaranteed optimal robust explanations for {NLP} models. In: IJCAI. pp.
  2658--2665 (2021)

\bibitem{meel-cp18}
Malioutov, D., Meel, K.S.: {MLIC:} {A} maxsat-based framework for learning
  interpretable classification rules. In: Hooker, J.N. (ed.) CP. pp. 312--327
  (2018)

\bibitem{msgcin-nips20}
Marques{-}Silva, J., Gerspacher, T., Cooper, M.C., Ignatiev, A., Narodytska,
  N.: Explaining naive bayes and other linear classifiers with polynomial time
  and delay. In: NeurIPS (2020)

\bibitem{msgcin-icml21}
Marques{-}Silva, J., Gerspacher, T., Cooper, M.C., Ignatiev, A., Narodytska,
  N.: Explanations for monotonic classifiers. In: ICML. pp. 7469--7479 (2021)

\bibitem{mshjpb-ijcai13}
Marques{-}Silva, J., Heras, F., Janota, M., Previti, A., Belov, A.: On
  computing minimal correction subsets. In: IJCAI. pp. 615--622 (2013)

\bibitem{msi-aaai22}
Marques{-}Silva, J., Ignatiev, A.: Delivering trustworthy {AI} through formal
  {XAI}. In: AAAI. pp. 12342--12350 (2022)

\bibitem{msjb-cav13}
Marques{-}Silva, J., Janota, M., Belov, A.: Minimal sets over monotone
  predicates in boolean formulae. In: CAV. pp. 592--607 (2013)

\bibitem{msjim-ijcai15}
Marques{-}Silva, J., Janota, M., Ignatiev, A., Morgado, A.: Efficient model
  based diagnosis with maximum satisfiability. In: IJCAI. pp. 1966--1972 (2015)

\bibitem{msjm-aij17}
Marques{-}Silva, J., Janota, M., Menc{\'{\i}}a, C.: Minimal sets on
  propositional formulae. problems and reductions. Artif. Intell.
  \textbf{252},  22--50 (2017),
  \url{https://doi.org/10.1016/j.artint.2017.07.005}

\bibitem{msm-ijcai20}
Marques{-}Silva, J., Menc{\'{\i}}a, C.: Reasoning about inconsistent formulas.
  In: IJCAI. pp. 4899--4906 (2020)

\bibitem{mss-iccad96}
Marques-Silva, J., Sakallah, K.A.: {GRASP} - a new search algorithm for
  satisfiability. In: Rutenbar, R.A., Otten, R.H.J.M. (eds.) ICCAD. pp.
  220--227 (1996)

\bibitem{mss-tcomp99}
Marques-Silva, J., Sakallah, K.A.: {GRASP:} {A} search algorithm for
  propositional satisfiability. {IEEE} Trans. Computers  \textbf{48}(5),
  506--521 (1999), \url{https://doi.org/10.1109/12.769433}

\bibitem{marquis-fair91}
Marquis, P.: Extending abduction from propositional to first-order logic. In:
  FAIR. pp. 141--155 (1991)

\bibitem{mehlhorn-csr11}
McConnell, R.M., Mehlhorn, K., N{\"{a}}her, S., Schweitzer, P.: Certifying
  algorithms. Comput. Sci. Rev.  \textbf{5}(2),  119--161 (2011),
  \url{https://doi.org/10.1016/j.cosrev.2010.09.009}

\bibitem{mcgregor-corr20}
McGregor, S.: Preventing repeated real world {AI} failures by cataloging
  incidents: The {AI} incident database. CoRR  \textbf{abs/2011.08512} (2020),
  \url{https://arxiv.org/abs/2011.08512}

\bibitem{mcgregor-aaai21}
McGregor, S.: Preventing repeated real world {AI} failures by cataloging
  incidents: The {AI} incident database. In: AAAI. pp. 15458--15463 (2021)

\bibitem{mcgregor-corr22b}
McGregor, S., Paeth, K., Lam, K.: Indexing {AI} risks with incidents, issues,
  and variants. CoRR  \textbf{abs/2211.10384} (2022),
  \url{https://doi.org/10.48550/arXiv.2211.10384}

\bibitem{mcquaid-jsep12}
McQuaid, P.A.: Software disasters - understanding the past, to improve the
  future. J. Softw. Evol. Process.  \textbf{24}(5),  459--470 (2012),
  \url{https://doi.org/10.1002/smr.500}

\bibitem{rudin-aaai22}
McTavish, H., Zhong, C., Achermann, R., Karimalis, I., Chen, J., Rudin, C.,
  Seltzer, M.: How smart guessing strategies can yield massive scalability
  improvements for sparse decision tree optimization. In: {AAAI} (February
  2022)

\bibitem{mipms-sat16}
Menc{\'{\i}}a, C., Ignatiev, A., Previti, A., Marques{-}Silva, J.: {MCS}
  extraction with sublinear oracle queries. In: SAT. pp. 342--360 (2016)

\bibitem{mpms-ijcai15}
Menc{\'{\i}}a, C., Previti, A., Marques{-}Silva, J.: Literal-based {MCS}
  extraction. In: IJCAI. pp. 1973--1979 (2015)

\bibitem{codish-jair14}
Metodi, A., Stern, R., Kalech, M., Codish, M.: A novel {SAT}-based approach to
  model based diagnosis. J. Artif. Intell. Res.  \textbf{51},  377--411 (2014),
  \url{https://doi.org/10.1613/jair.4503}

\bibitem{mill-bk43}
Mill, J.S.: A System of Logic, Ratiocinative and Inductive, vol.~1. John W.
  Parker (1843)

\bibitem{miller-pr56}
Miller, G.A.: The magical number seven, plus or minus two: Some limits on our
  capacity for processing information. Psychological review  \textbf{63}(2),
  81--97 (1956)

\bibitem{miller-aij19}
Miller, T.: Explanation in artificial intelligence: Insights from the social
  sciences. Artif. Intell.  \textbf{267},  1--38 (2019)

\bibitem{molnar-bk20}
Molnar, C.: Interpretable Machine Learning. Leanpub (2020),
  \url{http://tiny.cc/6c76tz}

\bibitem{muller-dsp18}
Montavon, G., Samek, W., M{\"{u}}ller, K.: Methods for interpreting and
  understanding deep neural networks. Digit. Signal Process.  \textbf{73},
  1--15 (2018)

\bibitem{mhlpms-cj13}
Morgado, A., Heras, F., Liffiton, M.H., Planes, J., Marques{-}Silva, J.:
  Iterative and core-guided maxsat solving: {A} survey and assessment.
  Constraints An Int. J.  \textbf{18}(4),  478--534 (2013)

\bibitem{mlms-hvc12}
Morgado, A., Liffiton, M.H., Marques{-}Silva, J.: {MaxSAT}-based {MCS}
  enumeration. In: HVC. pp. 86--101 (2012)

\bibitem{morgan-aij71}
Morgan, C.G.: Hypothesis generation by machine. Artif. Intell.  \textbf{2}(2),
  179--187 (1971)

\bibitem{salzberg-jair94}
Murthy, S.K., Kasif, S., Salzberg, S.: A system for induction of oblique
  decision trees. J. Artif. Intell. Res.  \textbf{2},  1--32 (1994)

\bibitem{nagendran-bmj20}
Nagendran, M., Chen, Y., Lovejoy, C.A., Gordon, A.C., Komorowski, M., Harvey,
  H., Topol, E.J., Ioannidis, J.P., Collins, G.S., Maruthappu, M.: Artificial
  intelligence versus clinicians: systematic review of design, reporting
  standards, and claims of deep learning studies. Bmj  \textbf{368} (2020)

\bibitem{hinton-icml10}
Nair, V., Hinton, G.E.: Rectified linear units improve restricted boltzmann
  machines. In: ICML. pp. 807--814 (2010)

\bibitem{narodytska-ijcai18}
Narodytska, N., Bj{\o}rner, N.S., Marinescu, M.V., Sagiv, M.: Core-guided
  minimal correction set and core enumeration. In: IJCAI. pp. 1353--1361 (2018)

\bibitem{nipms-ijcai18}
Narodytska, N., Ignatiev, A., Pereira, F., Marques{-}Silva, J.: Learning
  optimal decision trees with {SAT}. In: IJCAI. pp. 1362--1368 (2018)

\bibitem{nsmims-sat19}
Narodytska, N., Shrotri, A.A., Meel, K.S., Ignatiev, A., Marques{-}Silva, J.:
  Assessing heuristic machine learning explanations with model counting. In:
  SAT. pp. 267--278 (2019)

\bibitem{us-nraird19}
{National Science and Technology Council (US). Select Committee on Artificial
  Intelligence}: The national artificial intelligence research and development
  strategic plan: 2019 update  (2019),
  \url{https://www.nitrd.gov/pubs/National-AI-RD-Strategy-2019.pdf}

\bibitem{newman-brookings21}
Newman, J.: {Explainability won't save AI}. \url{} (2021)

\bibitem{nijssen-kdd07}
Nijssen, S., Fromont, {\'{E}}.: Mining optimal decision trees from itemset
  lattices. In: KDD. pp. 530--539 (2007)

\bibitem{nijssen-dmkd10}
Nijssen, S., Fromont, {\'{E}}.: Optimal constraint-based decision tree
  induction from itemset lattices. Data Min. Knowl. Discov.  \textbf{21}(1),
  9--51 (2010), \url{https://doi.org/10.1007/s10618-010-0174-x}

\bibitem{oecd-rcai21}
OECD: Recommendation of the council on artificial intelligence.
  \url{https://legalinstruments.oecd.org/en/instruments/OECD-LEGAL-0449}
  (2021), accessed: 2021-12-01

\bibitem{stuckey-cj09}
Ohrimenko, O., Stuckey, P.J., Codish, M.: Propagation via lazy clause
  generation. Constraints  \textbf{14}(3),  357--391 (2009).
  \doi{10.1007/s10601-008-9064-x},
  \url{https://doi.org/10.1007/s10601-008-9064-x}

\bibitem{oliver-tr92}
Oliver, J.J.: Decision graphs -- an extension of decision trees. Tech. Rep.
  92/173, Monash University (1992)

\bibitem{szeider-aaai21b}
Ordyniak, S., Szeider, S.: Parameterized complexity of small decision tree
  learning. In: AAAI. pp. 6454--6462 (2021)

\bibitem{papadimitriou-bk94}
Papadimitriou, C.H.: Computational Complexity. Addison Wesley (1994)

\bibitem{papadimitriou-jcss88}
Papadimitriou, C.H., Wolfe, D.: The complexity of facets resolved. J. Comput.
  Syst. Sci.  \textbf{37}(1),  2--13 (1988),
  \url{https://doi.org/10.1016/0022-0000(88)90042-6}

\bibitem{uzcategui-aij03}
P{\'{e}}rez, R.P., Uzc{\'{a}}tegui, C.: Preferences and explanations. Artif.
  Intell.  \textbf{149}(1),  1--30 (2003)

\bibitem{perrow-bk84}
Perrow, C.: Normal accidents: Living with high risk technologies. Basic books
  (1984)

\bibitem{mcgregor-corr22a}
Pittaras, N., McGregor, S.: A taxonomic system for failure cause analysis of
  open source {AI} incidents. CoRR  \textbf{abs/2211.07280} (2022),
  \url{https://doi.org/10.48550/arXiv.2211.07280}

\bibitem{plaisted-jsc86}
Plaisted, D.A., Greenbaum, S.: A structure-preserving clause form translation.
  J. Symb. Comput.  \textbf{2}(3),  293--304 (1986).
  \doi{10.1016/S0747-7171(86)80028-1}

\bibitem{pople-ijcai73}
Pople, H.E.: On the mechanization of abductive logic. In: IJCAI. pp. 147--152
  (1973)

\bibitem{pimms-ijcai15}
Previti, A., Ignatiev, A., Morgado, A., Marques{-}Silva, J.: Prime compilation
  of non-clausal formulae. In: IJCAI. pp. 1980--1988 (2015)

\bibitem{pms-aaai13}
Previti, A., Marques{-}Silva, J.: Partial {MUS} enumeration. In: AAAI (2013)

\bibitem{pmjms-sat17}
Previti, A., Menc{\'{\i}}a, C., J{\"{a}}rvisalo, M., Marques{-}Silva, J.:
  Improving {MCS} enumeration via caching. In: SAT. pp. 184--194 (2017)

\bibitem{pmjms-aaai18}
Previti, A., Menc{\'{\i}}a, C., J{\"{a}}rvisalo, M., Marques{-}Silva, J.:
  Premise set caching for enumerating minimal correction subsets. In: AAAI. pp.
  6633--6640 (2018)

\bibitem{propublica16}
ProPublica: Machine bias. \url{https://bit.ly/3zaHb2F} (2016)

\bibitem{quinlan-ml86}
Quinlan, J.R.: Induction of decision trees. Mach. Learn.  \textbf{1}(1),
  81--106 (1986)

\bibitem{quinlan-bk93}
Quinlan, J.R.: C4.5: programs for machine learning. Morgan-Kaufmann (1993)

\bibitem{rabe-sat16}
Rabe, M.N., Seshia, S.A.: Incremental determinization. In: SAT. pp. 375--392
  (2016)

\bibitem{rabe-fmcad15}
Rabe, M.N., Tentrup, L.: {CAQE:} {A} certifying {QBF} solver. In: FMCAD. pp.
  136--143 (2015)

\bibitem{toni-aij21}
Rago, A., Cocarascu, O., Bechlivanidis, C., Lagnado, D.A., Toni, F.:
  Argumentative explanations for interactive recommendations. Artif. Intell.
  \textbf{296},  103506 (2021)

\bibitem{toni-kr20}
Rago, A., Cocarascu, O., Bechlivanidis, C., Toni, F.: Argumentation as a
  framework for interactive explanations for recommendations. In: KR. pp.
  805--815 (2020)

\bibitem{doran-jair22}
Ras, G., Xie, N., van Gerven, M., Doran, D.: Explainable deep learning: {A}
  field guide for the uninitiated. J. Artif. Intell. Res.  \textbf{73},
  329--396 (2022), \url{https://doi.org/10.1613/jair.1.13200}

\bibitem{reiter-aij87}
Reiter, R.: A theory of diagnosis from first principles. Artif. Intell.
  \textbf{32}(1),  57--95 (1987),
  \url{https://doi.org/10.1016/0004-3702(87)90062-2}

\bibitem{guestrin-kdd16}
Ribeiro, M.T., Singh, S., Guestrin, C.: "why should {I} trust you?": Explaining
  the predictions of any classifier. In: KDD. pp. 1135--1144 (2016)

\bibitem{guestrin-aaai18}
Ribeiro, M.T., Singh, S., Guestrin, C.: Anchors: High-precision model-agnostic
  explanations. In: AAAI. pp. 1527--1535 (2018)

\bibitem{rivest-ml87}
Rivest, R.L.: Learning decision lists. Mach. Learn.  \textbf{2}(3),  229--246
  (1987)

\bibitem{rudin-naturemi19}
Rudin, C.: Stop explaining black box machine learning models for high stakes
  decisions and use interpretable models instead. Nature Machine Intelligence
  \textbf{1}(5),  206--215 (2019)

\bibitem{rudin-ss22}
Rudin, C., Chen, C., Chen, Z., Huang, H., Semenova, L., Zhong, C.:
  Interpretable machine learning: Fundamental principles and 10 grand
  challenges. Statistics Surveys  \textbf{16},  1--85 (2022)

\bibitem{rudin-mpc18}
Rudin, C., Ertekin, S.: Learning customized and optimized lists of rules with
  mathematical programming. Math. Program. Comput.  \textbf{10}(4),  659--702
  (2018), \url{https://doi.org/10.1007/s12532-018-0143-8}

\bibitem{rymon-amai94}
Rymon, R.: An {SE}-tree-based prime implicant generation algorithm. Ann. Math.
  Artif. Intell.  \textbf{11}(1-4),  351--366 (1994),
  \url{http://dx.doi.org/10.1007/BF01530750}

\bibitem{jarvisalo-kr16}
Saikko, P., Wallner, J.P., J{\"{a}}rvisalo, M.: Implicit hitting set algorithms
  for reasoning beyond {NP}. In: KR. pp. 104--113 (2016)

\bibitem{muller-ieee-proc21}
Samek, W., Montavon, G., Lapuschkin, S., Anders, C.J., M{\"{u}}ller, K.:
  Explaining deep neural networks and beyond: {A} review of methods and
  applications. Proc. {IEEE}  \textbf{109}(3),  247--278 (2021),
  \url{https://doi.org/10.1109/JPROC.2021.3060483}

\bibitem{muller-bk19}
Samek, W., Montavon, G., Vedaldi, A., Hansen, L.K., M{\"{u}}ller, K. (eds.):
  Explainable {AI:} Interpreting, Explaining and Visualizing Deep Learning.
  Springer (2019)

\bibitem{medium19}
Savoca, K.: When software kills.
  \url{https://medium.com/swlh/when-software-kills-ab6f48a15825} (2019)

\bibitem{szeider-aaai21a}
Schidler, A., Szeider, S.: {SAT}-based decision tree learning for large data
  sets. In: AAAI. pp. 3904--3912 (2021)

\bibitem{selman-aaai90}
Selman, B., Levesque, H.J.: Abductive and default reasoning: {A} computational
  core. In: AAAI. pp. 343--348 (1990)

\bibitem{rudin-facct22}
Semenova, L., Rudin, C., Parr, R.: On the existence of simpler machine learning
  models. In: FAccT. pp. 1827--1858 (2022)

\bibitem{seshia-cacm22}
Seshia, S.A., Sadigh, D., Sastry, S.S.: Toward verified artificial
  intelligence. Commun. {ACM}  \textbf{65}(7),  46--55 (2022),
  \url{https://doi.org/10.1145/3503914}

\bibitem{shalev-shwartz-bk14}
Shalev{-}Shwartz, S., Ben{-}David, S.: Understanding Machine Learning - From
  Theory to Algorithms. Cambridge University Press (2014),
  \url{https://bit.ly/3LBlznb}

\bibitem{shanahan-ijcai89}
Shanahan, M.: Prediction is deduction but explanation is abduction. In: IJCAI.
  pp. 1055--1060 (1989)

\bibitem{mcilraith-cp21}
Shati, P., Cohen, E., McIlraith, S.A.: {SAT}-based approach for learning
  optimal decision trees with non-binary features. In: CP. pp. 50:1--50:16
  (2021)

\bibitem{darwiche-kr20}
Shi, W., Shih, A., Darwiche, A., Choi, A.: On tractable representations of
  binary neural networks. In: KR. pp. 882--892 (2020)

\bibitem{darwiche-ijcai18}
Shih, A., Choi, A., Darwiche, A.: A symbolic approach to explaining bayesian
  network classifiers. In: IJCAI. pp. 5103--5111 (2018)

\bibitem{darwiche-aaai19}
Shih, A., Choi, A., Darwiche, A.: Compiling bayesian network classifiers into
  decision graphs. In: AAAI. pp. 7966--7974 (2019)

\bibitem{vedaldi-iclr14}
Simonyan, K., Vedaldi, A., Zisserman, A.: Deep inside convolutional networks:
  Visualising image classification models and saliency maps. In: ICLR (2014)

\bibitem{puget-ecai88}
de~Siqueira~N., J.L., Puget, J.: Explanation-based generalisation of failures.
  In: ECAI. pp. 339--344 (1988)

\bibitem{vandenbroeck-nips20}
Sivaraman, A., Farnadi, G., Millstein, T.D., den Broeck, G.V.:
  Counterexample-guided learning of monotonic neural networks. In: NeurIPS
  (2020)

\bibitem{landgraf-icml20}
Sixt, L., Granz, M., Landgraf, T.: When explanations lie: Why many modified
  {BP} attributions fail. In: ICML. pp. 9046--9057 (2020)

\bibitem{lakkaraju-aies20a}
Slack, D., Hilgard, S., Jia, E., Singh, S., Lakkaraju, H.: Fooling {LIME} and
  {SHAP:} adversarial attacks on post hoc explanation methods. In: AIES. pp.
  180--186 (2020)

\bibitem{kambhampati-aij21a}
Sreedharan, S., Chakraborti, T., Kambhampati, S.: Foundations of explanations
  as model reconciliation. Artif. Intell.  \textbf{301},  103558 (2021),
  \url{https://doi.org/10.1016/j.artint.2021.103558}

\bibitem{kambhampati-iclr22}
Sreedharan, S., Soni, U., Verma, M., Srivastava, S., Kambhampati, S.: Bridging
  the gap: Providing post-hoc symbolic explanations for sequential
  decision-making problems with inscrutable representations. In: ICLR (2022)

\bibitem{kambhampati-aij21b}
Sreedharan, S., Srivastava, S., Kambhampati, S.: Using state abstractions to
  compute personalized contrastive explanations for {AI} agent behavior. Artif.
  Intell.  \textbf{301},  103570 (2021),
  \url{https://doi.org/10.1016/j.artint.2021.103570}

\bibitem{kambhampati-ijcai19}
Sreedharan, S., Srivastava, S., Smith, D.E., Kambhampati, S.: Why can't you do
  that hal? explaining unsolvability of planning tasks. In: IJCAI. pp.
  1422--1430 (2019)

\bibitem{swartout-ijcai77}
Swartout, W.R.: A digitalis therapy advisor with explanations. In: IJCAI. pp.
  819--825 (1977)

\bibitem{swartout-aij83}
Swartout, W.R.: {XPLAIN:} {A} system for creating and explaining expert
  consulting programs. Artif. Intell.  \textbf{21}(3),  285--325 (1983)

\bibitem{tanner-plosntd08}
Tanner, L., and Jenny G~H~Low, M.S., Ong, A., Tolfvenstam, T., Lai, Y.L., Ng,
  L.C., Leo, Y.S., Puong, L.T., Vasudevan, S.G., Simmons, C.P., Hibberd, M.L.,
  Ooi, E.E.: Decision tree algorithms predict the diagnosis and outcome of
  dengue fever in the early phase of illness. PLoS neglected tropical diseases
  \textbf{2}(3), ~e196 (2008)

\bibitem{theverge20}
{The Verge}: {UK} ditches exam results generated by biased algorithm after
  student protests. \url{https://bit.ly/3kevIsB} (2020)

\bibitem{guan-ieee-tnnls21}
Tjoa, E., Guan, C.: A survey on explainable artificial intelligence {(XAI):}
  toward medical {XAI}. {IEEE} Trans. Neural Networks Learn. Syst.
  \textbf{32}(11),  4793--4813 (2021),
  \url{https://doi.org/10.1109/TNNLS.2020.3027314}

\bibitem{tseitin-scmml68}
TSEITIN, G.: On the complexity of derivation in propositional calculus. Studies
  in Constructive Mathematics and Mathematical Logic pp. 115--125 (1968)

\bibitem{umans-tcad06}
Umans, C., Villa, T., Sangiovanni{-}Vincentelli, A.L.: Complexity of two-level
  logic minimization. {IEEE} Trans. Comput. Aided Des. Integr. Circuits Syst.
  \textbf{25}(7),  1230--1246 (2006),
  \url{https://doi.org/10.1109/TCAD.2005.855944}

\bibitem{unesco-reai21}
UNESCO: Draft recommendation on the ethics of artificial intelligence.
  \url{https://unesdoc.unesco.org/ark:/48223/pf0000374266} (June 2021),
  accessed: 2021-12-01

\bibitem{liu-fat19}
Ustun, B., Spangher, A., Liu, Y.: Actionable recourse in linear classification.
  In: FAT. pp. 10--19 (2019)

\bibitem{valdes-naturesr16}
Valdes, G., Luna, J.M., Eaton, E., Simone, C.B., Ungar, L.H., Solberg, T.D.:
  {MediBoost}: a patient stratification tool for interpretable decision making
  in the era of precision medicine. Scientific reports  \textbf{6}(1), ~1--8
  (2016)

\bibitem{valiant-cacm84}
Valiant, L.G.: A theory of the learnable. Commun. {ACM}  \textbf{27}(11),
  1134--1142 (1984), \url{https://doi.org/10.1145/1968.1972}

\bibitem{vardi-cacm10a}
Vardi, M.Y.: On {P}, {NP}, and computational complexity. Commun. {ACM}
  \textbf{53}(11), ~5 (2010), \url{https://doi.org/10.1145/1839676.1839677}

\bibitem{vardi-cacm14a}
Vardi, M.Y.: Boolean satisfiability: theory and engineering. Commun. {ACM}
  \textbf{57}(3), ~5 (2014), \url{https://doi.org/10.1145/2578043}

\bibitem{magazzeni-jair22}
Vasileiou, S.L., Yeoh, W., Son, T.C., Kumar, A., Cashmore, M., Magazzeni, D.: A
  logic-based explanation generation framework for classical and hybrid
  planning problems. J. Artif. Intell. Res.  \textbf{73},  1473--1534 (2022),
  \url{https://doi.org/10.1613/jair.1.13431}

\bibitem{alfano-fat20}
Venkatasubramanian, S., Alfano, M.: The philosophical basis of algorithmic
  recourse. In: FAT. pp. 284--293 (2020)

\bibitem{schaus-cj20}
Verhaeghe, H., Nijssen, S., Pesant, G., Quimper, C., Schaus, P.: Learning
  optimal decision trees using constraint programming. Constraints An Int. J.
  \textbf{25}(3-4),  226--250 (2020),
  \url{https://doi.org/10.1007/s10601-020-09312-3}

\bibitem{schaus-ijcai20a}
Verhaeghe, H., Nijssen, S., Pesant, G., Quimper, C., Schaus, P.: Learning
  optimal decision trees using constraint programming (extended abstract). In:
  IJCAI. pp. 4765--4769 (2020)

\bibitem{verwer-cpaior17}
Verwer, S., Zhang, Y.: Learning decision trees with flexible constraints and
  objectives using integer optimization. In: CPAIOR. pp. 94--103 (2017)

\bibitem{verwer-aaai19}
Verwer, S., Zhang, Y.: Learning optimal classification trees using a binary
  linear program formulation. In: AAAI. pp. 1625--1632 (2019)

\bibitem{eisemann-corr19}
Viering, T.J., Wang, Z., Loog, M., Eisemann, E.: How to manipulate cnns to make
  them lie: the {GradCAM} case. CoRR  \textbf{abs/1907.10901} (2019),
  \url{http://arxiv.org/abs/1907.10901}

\bibitem{voronkov-cav14}
Voronkov, A.: {AVATAR:} the architecture for first-order theorem provers. In:
  CAV. pp. 696--710 (2014)

\bibitem{verwer-icml21}
Vos, D., Verwer, S.: Efficient training of robust decision trees against
  adversarial examples. In: ICML. pp. 10586--10595 (2021)

\bibitem{verwer-aaai22}
Vos, D., Verwer, S.: Robust optimal classification trees against adversarial
  examples. In: AAAI. pp. 8520--8528 (2022)

\bibitem{waldchen-phd22}
W\"{a}ldchen, S.: Towards Explainable Artificial Intelligence -- Interpreting
  Neural Network Classifiers with Probabilistic Prime Implicants. Ph.D. thesis,
  Technischen Universit\"{a}t Berlin (3 2022)

\bibitem{kutyniok-jair21}
W{\"{a}}ldchen, S., MacDonald, J., Hauch, S., Kutyniok, G.: The computational
  complexity of understanding binary classifier decisions. J. Artif. Intell.
  Res.  \textbf{70},  351--387 (2021),
  \url{https://doi.org/10.1613/jair.1.12359}

\bibitem{rudin-aistats15}
Wang, F., Rudin, C.: Falling rule lists. In: AISTATS (2015)

\bibitem{gupta-tplp22}
Wang, H., Shakerin, F., Gupta, G.: {FOLD-RM:} {A} scalable, efficient, and
  explainable inductive learning algorithm for multi-category classification of
  mixed data. Theory Pract. Log. Program.  \textbf{22}(5),  658--677 (2022),
  \url{https://doi.org/10.1017/S1471068422000205}

\bibitem{rudin-jmlr17b}
Wang, T., Rudin, C., Doshi{-}Velez, F., Liu, Y., Klampfl, E., MacNeille, P.: A
  bayesian framework for learning rule sets for interpretable classification.
  J. Mach. Learn. Res.  \textbf{18},  70:1--70:37 (2017)

\bibitem{weber-jal09}
Weber, T., Amjad, H.: Efficiently checking propositional refutations in {HOL}
  theorem provers. J. Applied Logic  \textbf{7}(1),  26--40 (2009),
  \url{https://doi.org/10.1016/j.jal.2007.07.003}

\bibitem{zhou-corr22}
Wei, M., Zhou, Z.: {AI} ethics issues in real world: Evidence from {AI}
  incident database. CoRR  \textbf{abs/2206.07635} (2022),
  \url{https://doi.org/10.48550/arXiv.2206.07635}

\bibitem{yampolskiy-corr21}
Williams, R.M., Yampolskiy, R.V.: Understanding and avoiding {AI} failures: {A}
  practical guide. CoRR  \textbf{abs/2104.12582} (2021),
  \url{https://arxiv.org/abs/2104.12582}

\bibitem{hazan-aies19}
Wolf, L., Galanti, T., Hazan, T.: A formal approach to explainability. In:
  AIES. pp. 255--261 (2019)

\bibitem{kumar-bk09}
Wu, X., Kumar, V. (eds.): The top ten algorithms in data mining. CRC press
  (2009)

\bibitem{gupta-nips17}
You, S., Ding, D., Canini, K.R., Pfeifer, J., Gupta, M.R.: Deep lattice
  networks and partial monotonic functions. In: NeurIPS. pp. 2981--2989 (2017)

\bibitem{ignatiev-cp20}
Yu, J., Ignatiev, A., Stuckey, P.J., Bodic, P.L.: Computing optimal decision
  sets with {SAT}. In: CP. pp. 952--970 (2020)

\bibitem{ignatiev-jair21}
Yu, J., Ignatiev, A., Stuckey, P.J., Bodic, P.L.: Learning optimal decision
  sets and lists with {SAT}. J. Artif. Intell. Res.  \textbf{72},  1251--1279
  (2021), \url{https://doi.org/10.1613/jair.1.12719}

\bibitem{yisnms-corr22}
Yu, J., Ignatiev, A., Stuckey, P.J., Narodytska, N., Marques{-}Silva, J.:
  Eliminating the impossible, whatever remains must be true. CoRR
  \textbf{abs/2206.09551} (2022),
  \url{https://doi.org/10.48550/arXiv.2206.09551}

\bibitem{yisnms-aaai23}
Yu, J., Ignatiev, A., Stuckey, P.J., Narodytska, N., Marques-Silva, J.:
  Eliminating the impossible, whatever remains must be true: On extracting and
  applying background knowledge in the context of formal explanations. In:
  {AAAI} (2023)

\end{thebibliography}
}{
  % Import bibl (original .bbl) file

}
\label{lastpage}

\end{document}